\documentclass[11pt,a4paper]{article}

\usepackage[utf8]{inputenc}
\usepackage[T1]{fontenc}
\usepackage{mathpazo} 
\usepackage{microtype} 

\usepackage{amsmath,amssymb,amsthm}
\usepackage{bm}

\usepackage{geometry}
\usepackage{graphicx}
\usepackage{xcolor}
\usepackage{float}
\usepackage{subcaption}

\usepackage{tikz}
\usetikzlibrary{shapes, shapes.geometric, arrows.meta, positioning, calc, fit, backgrounds, shadows}
\definecolor{myyellow}{RGB}{252, 190, 0} 

\usepackage{booktabs}
\usepackage{enumitem}
\usepackage{tabularx}
\usepackage{array}

\usepackage{titlesec}
\titleformat{\section}{\Large\bfseries\color{blue!80!black}}{\thesection}{1em}{}[\titlerule]
\titleformat{\subsection}{\large\bfseries\color{blue!70!black}}{\thesubsection}{1em}{}
\titleformat{\subsubsection}{\normalsize\bfseries\color{blue!60!black}}{\thesubsubsection}{1em}{}

\usepackage{fancyhdr}
\usepackage{tcolorbox}
\tcbuselibrary{theorems, skins, breakable}
\numberwithin{equation}{section}

\newtcbtheorem[number within=subsection]{theorembox}{Definition}{
  colback=blue!2!white,
  colframe=blue!80!black,
  fonttitle=\bfseries,
  separator sign dash,
  breakable,
  enhanced,
  drop fuzzy shadow,
  boxrule=0.8pt
}{thm}

\newtcbtheorem[number within=subsection]{examplebox}{Example}{
  colback=green!2!white,
  colframe=green!60!black,
  fonttitle=\bfseries,
  separator sign dash,
  breakable,
  enhanced,
  drop fuzzy shadow,
  boxrule=0.8pt
}{ex}

\newtcbtheorem[number within=subsection]{highlightbox}{Problem Reformulation}{
  colback=orange!2!white,
  colframe=orange!80!black,
  fonttitle=\bfseries,
  separator sign dash,
  breakable,
  enhanced,
  drop fuzzy shadow,
  boxrule=0.8pt
}{hl}

\newtcolorbox{abstractbox}{
  enhanced,
  colback=gray!5!white,    
  colframe=blue!80!black,  
  arc=0pt,                 
  outer arc=0pt,
  leftrule=3pt,            
  rightrule=0pt,           
  toprule=0pt,
  bottomrule=0pt,
  fontupper=\small,        
  boxsep=5pt,
  breakable
}

\usepackage{cite}
\usepackage{hyperref}
\hypersetup{
  colorlinks=true, 
  linkcolor=blue!80!black, 
  citecolor=red!80!black,
  urlcolor=blue
}

\usepackage[affil-it]{authblk} 
\begin{document}

\title{\vspace{-2cm}\textbf{\huge Stabilization Learning: \\ \Large A Paradigm Transition Bridging Control Theory and Machine Learning}}

\author[1,2,3]{Quan Quan\thanks{Email: qq\_buaa@buaa.edu.cn}}
\affil[1]{School of Automation Science and Electrical Engineering, Beihang University, Beijing 100191, China}
\affil[2]{Tianmushan Laboratory, Beihang University, Hangzhou 311115, China}
\affil[3]{Zhongguancun Academy, Beijing 100094, China}

\date{}
\maketitle

\begin{abstractbox}
  \noindent\textbf{\large Abstract:} 
  \vspace{0.5em}

  \noindent {Stabilization learning is an interdisciplinary paradigm that bridges control theory and machine learning. Its core idea is to enable systems to adjust their policies under perturbations or environmental changes through real-time feedback and adaptive mechanisms. It takes stability as its primary goal, distinguishing itself from certificate learning, which focuses on formal proofs, and reinforcement learning, which pursues optimality. It encompasses a range of methods, including Lyapunov-based analysis and design, deep feature extraction, and data-driven feedback synthesis, and is applicable to complex high-dimensional, nonlinear systems. This paper elaborates on the two major categories of stability in stabilization learning, as well as three typical application scenarios: control, observation, and recognition. It constructs a unified mathematical framework based on a six-tuple (including core elements like state space and action space), and expands into two types of seven-tuple models: constrained learning with barrier spaces and tracking problems with targets. It also analyzes the roles, meanings, and implementation choices of key elements such as state space, controlled system, metrics, and policy. Through the formal reformulation of 11 types of problems, including multi-agent cooperative tracking, visual servo robot position stabilization, chess games, and Push-T tasks, this paper illustrates the potential applicability of the framework across multiple domains.} Finally, it points out that future stabilization learning will focus on two major directions: constructing a unified problem framework (covering problem transformation in control and learning domains) and achieving efficient and robust learning (improving data and parameter efficiency and enhancing disturbance-rejection robustness), providing solutions for complex system control that combine theoretical rigor with engineering practicality.

  \vspace{1em}

\end{abstractbox}

\vspace{1em}

\section{Basic Concepts of Stabilization Learning}

\subsection{What is Stabilization Learning?}

\begin{figure}[H]
  \centering
  \begin{subfigure}[b]{0.48\textwidth}
    \centering
    \includegraphics[width=\textwidth]{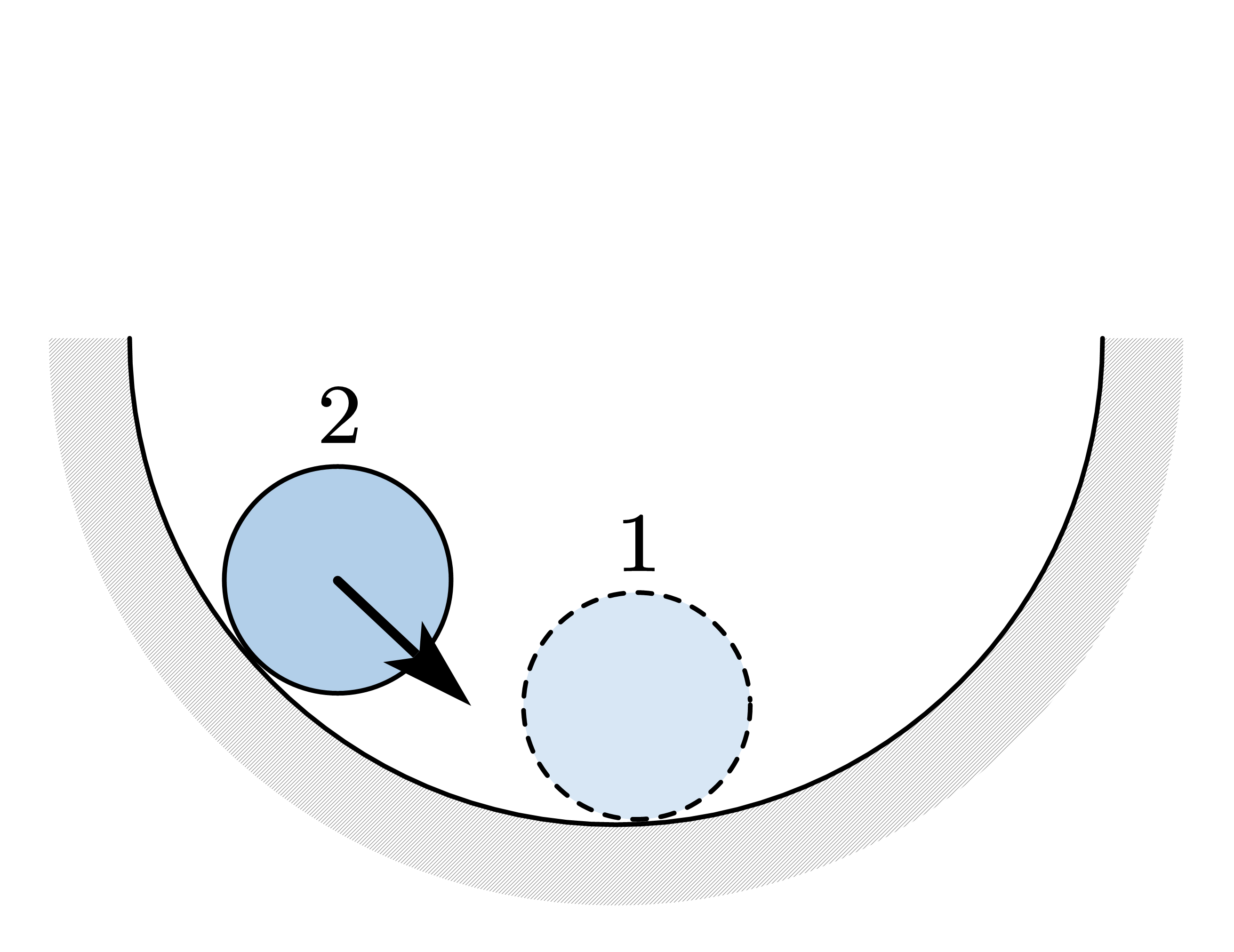}
    \caption{Stable Equilibrium}
    \label{fig:stable_fig_1}
  \end{subfigure}
  \hfill
  \begin{subfigure}[b]{0.48\textwidth}
    \centering
    \includegraphics[width=\textwidth]{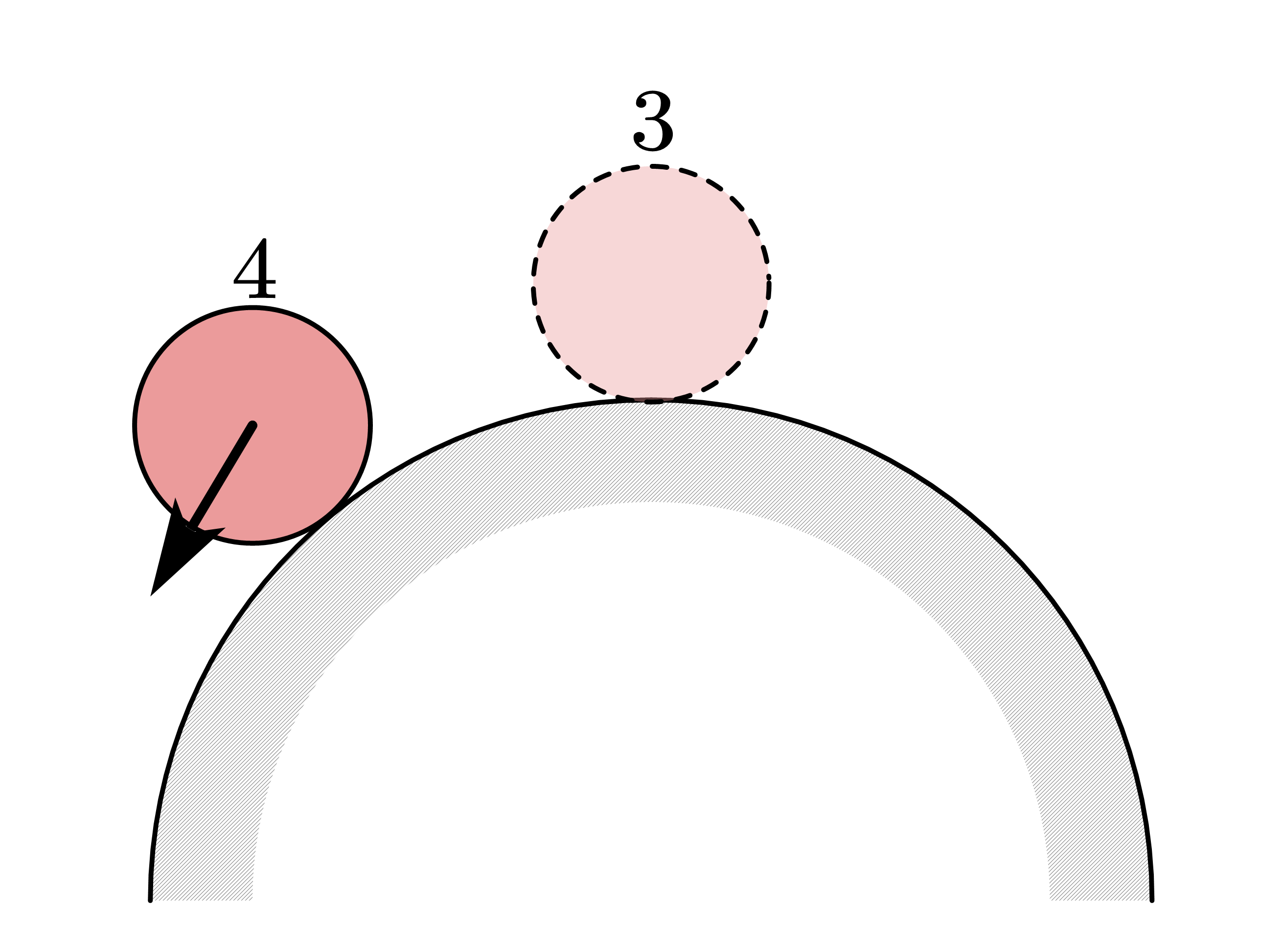}
    \caption{Unstable Equilibrium}
    \label{fig:unstable_fig_2}
  \end{subfigure}
  \caption{Two classic equilibrium states}
  \label{fig:stable_figs}
\end{figure}

Stability is a cornerstone concept in dynamical systems theory, rigorously describing a system's ability to maintain its internal structure and order when subjected to external disturbances. Its core lies in studying the sensitivity and robustness of a system's steady-state solutions (such as equilibrium points and periodic orbits) to initial conditions or parameter perturbations \cite{Wiener1961Cybernetics2nd}. A stable system can constrain the effects of perturbations within a finite range or cause them to gradually decay, thereby maintaining its inherent operating mode or converging to a desired state; conversely, an unstable system will diverge or undergo structural changes due to minor deviations. Intuitively, a ball at the bottom of a bowl is stable. If the ball is not at the bottom, it will roll back and forth until it stabilizes at the bottom, as shown in Figure \ref{fig:stable_fig_1}; a ball at the top of a hill is unstable because a slight perturbation will typically cause it to move away from the equilibrium point, as shown in Figure \ref{fig:unstable_fig_2}. Therefore, stability is defined as the system's ability to return to an equilibrium state after a perturbation, while control achieves system stability through the feedback loop of actuators and controllers. For example: aircraft rely on ``control surface + torque'' control to achieve attitude stability, robots use ``joint + torque'' feedback to maintain motion trajectories, and social systems rely on ``government + policy'' to maintain order; their core is to ensure the system state converges to the target state set through designed feedback mechanisms.

Stability can be primarily divided into two categories based on the system's dependence on initial conditions and external inputs:
\begin{itemize}[leftmargin=*, itemsep=2pt]
  \item \textbf{Stability dependent only on initial conditions}: Refers to the system's response to minor changes in initial conditions. If initial value perturbations do not cause system instability, the system is stable with respect to initial conditions. Common types of such stability include Lyapunov stability (the system state remains within a finite range under perturbations), asymptotic stability (the system eventually returns to the equilibrium point), and exponential stability \cite{khalil2002nonlinear}.
  \item \textbf{Stability dependent on both initial conditions and external inputs}: The system's stability depends on the combined effect of initial conditions and external inputs. Common types include Input-to-State Stability (ISS) and $\mathcal{L}_2$ stability, ensuring the system's robustness against perturbations in both initial conditions and external inputs \cite{khalil2002nonlinear}.
\end{itemize}

\textbf{Stabilization Learning} is a learning-and-control paradigm in which a feedback policy is learned so that the closed-loop system maintains or recovers stability under perturbations and environmental changes. Unlike traditional \textbf{Stabilizing Control} methods, stabilization learning is data-driven, dynamically optimizing decisions based on system states, and uses data and feedback to adapt policies, while reducing dependence on hand-crafted rules and problem-specific parameter tuning. This perspective aims to reduce dependence on task-specific handcrafted rules by emphasizing feedback, metrics, and stability-oriented representations, allowing the system to operate stably and continuously in complex, uncertain environments, adapt to new situations, and ensure that system performance does not experience severe fluctuations in changing environments.

Stabilization learning takes ``stability'' as its primary task, and treats optimality as a secondary objective, whereas traditional reinforcement learning primarily pursues optimality. Stability can be regarded as an important foundation for robustness and generalization, while optimality further concerns performance improvement under a specified metric, under certain inverse optimality conditions or with properly constructed cost functions, a stable strategy can be interpreted as optimal with respect to some performance metric. The design process of a Linear Quadratic Regulator (LQR) \cite{anderson1990optimal} is a typical example of the intertwined nature of stability and optimality. Designing an optimal metric can yield a certain control feedback gain, and given a stable control feedback gain, a corresponding optimal metric can also be found { under suitable inverse optimality conditions}.

Stabilization learning and certificate learning \cite{dawson2023safe} both focus on system stability, but differ in methodology and application. Certificate learning focuses on designing controllers by constructing Lyapunov functions, barrier functions, and related certificates, and on formally proving system stability, especially in control systems where stability under specific control strategies must be verified. Stabilization learning, as a broader concept, encompasses other stability criteria besides Lyapunov methods. Furthermore, through techniques like deep feature extraction, dynamical system establishment, stable system construction, stabilization learning problem transformation, and generalization to other learning problems, stabilization learning ensures stability during complex learning processes (especially in high-dimensional, nonlinear systems). Stabilization learning not only focuses on stability in known environments but also emphasizes how systems stably learn and optimize in changing environments. Overall, it employs more diverse methods to ensure systems stably adapt and continuously optimize their behavior during the learning process.

\subsection{Typical Applications}

\subsubsection{Control Problems}

Control problems are directly related to stabilization learning. For the following dynamical system:
\begin{equation}
\dot{s} = f(s, a),
\end{equation}
where $s$ is the state and $a$ is the control input. Assuming {$s_{\mathrm{d}}$} is the desired state, we design the controller $a=\pi(s,{ s_{\mathrm{d}}})$ such that:
\begin{equation}
\dot{s} = f(s, \pi(s,{s_{\mathrm{d}}})). \label{eq:control}
\end{equation}
The goal of system design is to establish stability with respect to the equilibrium point $s_{\mathrm{d}}$ at the desired state. Taking asymptotic stability as an example, the system is expected to satisfy such condition that when $t\to \infty$, $s(t)-{s_{\mathrm{d}}}(t)\to 0$.

\begin{examplebox}{Hovering Control of Multi-Rotor UAVs}{drone-hover}
  The hovering control of multi-rotor UAVs is a typical stabilization learning problem. In indoor environments, hover positioning that relies on satellite navigation is infeasible, and methods based solely on visual relative-pose feedback may be sensitive to illumination changes, occlusion, and viewpoint variation. Therefore, using global features extracted from images captured by the multi-rotor UAV for control has become an effective stabilization learning solution. Through global feature extraction, the UAV's pose and environmental information $s$ can be obtained, and a controller $a=\pi(s,{s_{\mathrm{d}}})$ can be reasonably designed so that the UAV stably hovers at the target position ${s_{\mathrm{d}}}$. Further details can be found in Section \ref{sec:drone_hover}.
\end{examplebox}

\subsubsection{Observation Problems}

Observation problems \cite{kailath1980linear} can be transformed into stabilization problems. For the following dynamical system:
\begin{align}
\dot{s} &= f(s), \\
y &= h(s).
\end{align}
We expect to observe the system state $s$ through the observable variable $y$. We can construct the following system:
\begin{equation}
\dot{\hat{s}} = \hat{f}(\hat{s})+l(y,\hat{y}). \label{eq:observer}
\end{equation}
Assuming $\tilde{s}=s-\hat{s}$, we have:
\begin{align}
\dot{\tilde{s}} &=f(s)- \hat{f}(\hat{s})-l(y,\hat{y}), \\
\hat{y} &= h(\hat{s}).
\end{align}
Similarly, we hope to design $l(y,\hat{y})$ such that when $t\to \infty$, $\tilde{s}(t)\to 0$. In this case, $l(y,\hat{y})$ is referred to as the observer correction function in the control domain.

\begin{examplebox}{UAV Pose Estimation Using Image Sequences}{pose-estimation}
  To better track a UAV, we need to estimate its 3D position and attitude (i.e., 3D pose) from 2D image sequences acquired by observation equipment, allowing anticipation of the UAV's further movement based on its attitude. Methodologically, this boils down to a stabilization learning problem. Specifically, the system first extracts features from the image sequence as the observable variable $y$; then, by designing a desired observer correction function $l(y,\hat{y})$, it infers the UAV's aerial pose $s$ using the observation information. The key to stabilization learning is ensuring the robustness of the output results. Therefore, the observer correction function $l(y,\hat{y})$ must be reasonably designed to ensure that the estimation of the UAV's 3D pose maintains stability and accuracy regardless of whether the received images are subject to disturbances such as lighting changes, perspective shifts, or partial occlusions. Further details can be found in Section \ref{sec:RelativePose}.
\end{examplebox}

\subsubsection{Recognition Problems}

Recognition problems can also be converted into stabilization problems. In object recognition tasks within complex environments, the appearance and background of the target often vary significantly, leading to unstable recognition accuracy in traditional methods. When viewing this problem as a stabilization learning problem, the core lies in how to maintain high robustness and stability when facing different types of target images by learning the intrinsic features of the objects and constructing a stable system. Stabilization learning can dynamically adjust model parameters, ensuring continuous and stable object recognition and understanding regardless of changes in the target's pose, lighting, or background.

Assuming $I$ is the image to be understood and $s$ represents the classification features of the image, the following system can be represented as:
\begin{equation}
\dot{s}=f(s,I,a). \label{eq:identifier}
\end{equation} 
We hope to design a learning policy $a=\pi(s, I)$ such that in the event of unclear or incomplete images (i.e., $I+\Delta I$), the system can still utilize the feature $s$ through the classifier $h(\cdot)$ to output a result that converges to the desired correct classification $ \mathcal{Y}_{\mathrm{d}}$, meaning the metric $d$ between them approaches 0:
\begin{equation}
d(y,{\mathcal{Y}_{\mathrm{d}}})\to 0, \quad y = h(s). \label{eq:identifier_loss}
\end{equation}

Transforming static recognition into a dynamic process is a key manifestation of the stabilization learning philosophy in technical implementation. The core advantage of this transformation is that the system no longer reacts to a single, isolated ``image'', but frames the task as a continuous decision sequence with temporal and contextual dependencies, achieving smarter and more robust outputs. For instance, traditional image generation models (such as early Generative Adversarial Networks (GANs)) attempted to map random noise into target images all at once—a typical ``static'' generation process prone to mode collapse or unnatural results; diffusion models, however, break generation down into hundreds of continuous ``denoising'' steps. Starting from pure noise, each step makes a minor optimization decision based on the current state (noisy image) and data-driven prediction (denoising direction), gradually ``sculpting'' the final image \cite{song2021score}. Furthermore, when recognizing video streams, a frame-wise static recognizer may cause inter-frame flickering, jitter, or logical contradictions (e.g., sudden changes in object color or position). Dynamic recognition introduces temporal consistency constraints \cite{bonneel2015video}; when processing the current frame, the system actively ``remembers'' and ``references'' the context of preceding and succeeding frames, thereby enhancing the continuity and stability of recognition results and achieving high-level understanding.

\begin{examplebox}{Image Navigation}{image-nav}
  In embodied AI scenarios, an agent often receives videos or images as inputs to determine the next action and execute path navigation, making the correct understanding of real-world scenes depicted in images particularly crucial. $I$ represents the input image/video, $s$ represents the deep features extracted from it, and the output $y$ represents the real-world scene label categorized by the agent. The objective of stabilization learning is to design a reasonable learning policy $a = \pi(s,I)$ to ensure that when the agent receives blurry, incomplete, or otherwise disturbed signals $I+\Delta I$, it can still use the feature $s$ through the classifier $h(\cdot)$ to output a result that converges to the expected correct classification $\mathcal{Y}_{\mathrm{d}}$, thereby achieving all-weather navigation across various application scenarios.
\end{examplebox}

\section{Control-Oriented Stabilization Learning Framework}

\begin{figure}[H]
  \centering
  \includegraphics[width=0.98\textwidth]{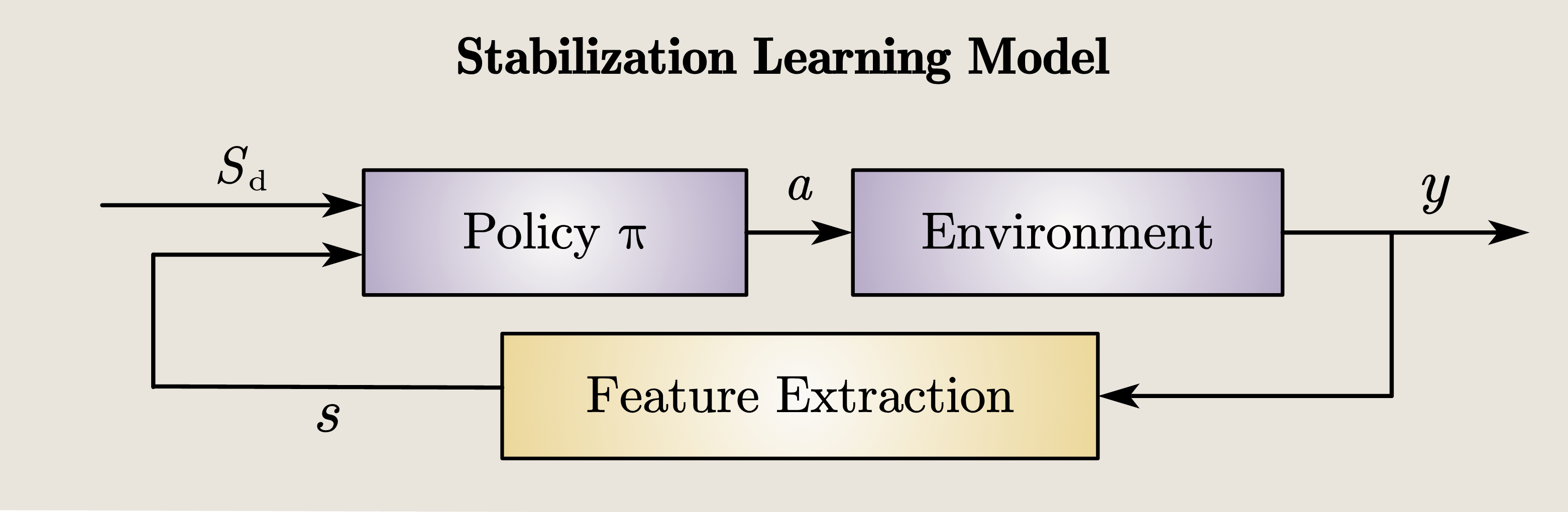}
  \caption{Stabilization learning problem framework}
  \label{fig:stabilization_learning_model}
\end{figure}

From a unified perspective, the three classical tasks of control, observation, and recognition can all be reformulated as ``closed-loop system stability" problems, thereby being integrated into the theoretical framework of stabilization learning, as shown in Figure \ref{fig:stabilization_learning_model}. Specifically, control problems aim to design policies that make state trajectories asymptotically converge to a given equilibrium point, the essence of which is verifying the stability of the closed-loop dynamics with respect to a desired state set. Thus, control, observation, and recognition no longer belong to disparate disciplines but are unified as ``finding a feedback mechanism that stabilizes a certain class of error dynamics to the origin (or invariant set)." This perspective provides a common mathematical interface for subsequent data-driven methods to simultaneously optimize stability and performance. Therefore, we can draw upon the state tracking control concepts in control theory to transform the stabilization learning problem into a state stabilization control problem. Specifically, the target states for control, observation, and recognition are uniformly abstracted into a desired state set $\mathcal{S}_{\mathrm{d}}$, and a feedback policy $a$ is designed so that the system state trajectory converges to $\mathcal{S}_{\mathrm{d}}$. Meanwhile, after each operation step, the current state is fed back to the decision-maker and actuator through a state feedback mechanism (where mathematical features of the state are extracted), ensuring continuous execution of operations according to the policy. {This guarantees that the state $s$ asymptotically converges to the domain of the desired state set $\mathcal{S}_{\mathrm{d}}$.}

\subsection{Mathematical Description of Stabilization Learning Problems}

\subsubsection{Basic Definitions}

The stabilization learning problem can be uniformly described using a mathematical six-tuple\( (\mathcal{S},\mathcal{A},\mathcal{P},\pi,\mathcal{S}_{\mathrm{d}},d)\).

\begin{theorembox}{General Mathematical Description of Stabilization Learning Problems}{stability-control}
  Real-world stabilization learning problems can basically be reduced to the form of a six-tuple\( (\mathcal{S},\mathcal{A},\mathcal{P},\pi,\mathcal{S}_{\mathrm{d}},d)\), defined as follows:
  \begin{itemize}
    \item $\mathcal{S}$: \textbf{State Space}, representing the set of all possible states of the system, which can be divided into finite or infinite state spaces;
    \item $\mathcal{A}$: \textbf{Action Space}, representing the set of all possible actions the system can execute, which can be divided into discrete or continuous action spaces;
    \item $\mathcal{P}$: \textbf{Plant}, representing the mathematical model describing how the system state evolves after an operation step. It can be divided into deterministic and stochastic models:
      \begin{itemize}
        \item \textbf{Deterministic Model}: Can be further divided into continuous and discrete models:
        \[ 
        \begin{aligned}
        \dot{s} &= f(s, a), \quad s\in\mathcal{S}, a\in\mathcal{A} \quad \text{(Continuous Model)} \\
        s_{t+1} &= f(s_t, a_t), \quad s_t,s_{t+1}\in\mathcal{S}, a_t\in\mathcal{A} \quad \text{(Discrete Model)} 
        \end{aligned}
        \]
        Given the current state and the action to be taken, the system's state change is entirely given by the deterministic mapping $f$;
        \item \textbf{Stochastic Model}: Can also be divided into continuous and discrete models:
        \[ 
        \begin{aligned}
        \mathbb{P}(\dot{s}|s,a), \quad & s\in\mathcal{S}, a\in\mathcal{A} \quad \text{(Continuous Model)} \\
        \mathbb{P}(s_{t+1}|s_t,a_t), \quad & s_t,s_{t+1}\in\mathcal{S}, a_t\in\mathcal{A} \quad \text{(Discrete Model)} 
        \end{aligned}
        \]
        The evolution of the system state is described by a state transition probability distribution;
      \end{itemize}
    \item $\pi$: \textbf{Policy}, a mapping function from states to actions, representing the action to be taken in a certain state:
      \begin{itemize}
        \item \textbf{Deterministic Policy}: $a=\pi(s)$, outputting a unique action directly;
        \item \textbf{Stochastic Policy}: $\pi(a|s) = \mathbb{P}(a |s ), a\in \mathcal{A}, s\in\mathcal{S}$, outputting a probability distribution of actions.
      \end{itemize}
    \item  $\mathcal{S}_{\mathrm{d}}$: \textbf{Desired State Space}, a subset of the state space $\mathcal{S}_{\mathrm{d}}\subseteq \mathcal{S}$;
    \item $d$: \textbf{Metric Function}, {$\mathcal{S}\times\mathcal{P}(\mathcal{S}) \to \mathbb{R}_{\geq 0}$. $d(s,\mathcal{S}_{\mathrm{d}}),\, s \in \mathcal{S}, \, \mathcal{S}_{\mathrm{d}} \in \mathcal{P}(\mathcal{S})$ represents the distance from the current state to the desired state set, which depends on the chosen metric.}
  \end{itemize}
  Taking asymptotic stability as an example, we need to find a specific policy $\pi$ such that when $t\rightarrow \infty$, \( d\left(s\left(t\right),\mathcal{S}_{\mathrm{d}}\left(t\right)\right)\rightarrow 0\).
\end{theorembox}

\subsubsection{Constrained Learning under Barrier Space $\mathcal{B}$}

In real-world systems' stabilization learning problems, the state space is often restricted by physical conditions or inherent characteristics of the problem, rendering some policies unfeasible in practice (i.e., ``prohibited") despite being feasible under unconstrained assumptions. Examples include physical limits of robotic joints, safety constraints of traffic laws in autonomous driving, and prior constraints like power limits on industrial devices. Therefore, it is necessary to introduce a key element into the problem definition—the Barrier Space $\mathcal{B}$\footnote{ The concept of a barrier space is closely related to Control Barrier Functions (CBFs) in control theory. A CBF defines a safe set by constructing a function $b(\cdot)$, such that the system state $s$ remains within the safe region when $b(s) \ge 0$. In control law design, the derivative of $b(s)$ along the system trajectory is constrained (e.g., satisfying $\dot{b}(s) \ge -\alpha(b(s))$) to guarantee the forward invariance of the safe set, thereby strictly ensuring that the system avoids hazardous regions. For further studies on CBFs, readers are referred to \cite{dawson2023safe, dawson2021safe}.}—thereby expanding the original problem into a new seven-tuple form for constrained learning: $(\mathcal{S},\mathcal{A},\mathcal{P},\pi,\mathcal{S}_{\mathrm{d}},\mathcal{B},d)$. The formal definition of $\mathcal{B}$ is provided in Definition \ref{constrained-learning}.

\begin{theorembox}{Stabilization Learning Problems with Constraints}{constrained-learning} \label{constrained-learning}
  When a constrained learning problem is transformed into a stabilization learning problem, the six elements $(\mathcal{S},\mathcal{A},\mathcal{P},\pi,\mathcal{S}_{\mathrm{d}},d)$ retain their original definitions, but an additional Barrier Space $\mathcal{B}$ must be introduced:
  \begin{itemize}
    \item $\mathcal{B}$: \textbf{Barrier Space}, representing the subset of all unreachable states in the state space $\mathcal{S}$, where $\mathcal{B}\subset\mathcal{S}$.
  \end{itemize}
  The stabilization learning problem under constraints requires finding a specific policy $\pi$ such that when $t\rightarrow \infty$, \( d\left(s\left(t\right),\mathcal{S}_{\mathrm{d}}\left(t\right)\right)\rightarrow 0\), and \(\forall t, s\left(t\right)\notin\mathcal{B}\).
\end{theorembox}

Moreover, in optimization problems, some actions can be judged as non-contributing or even harmful to stability based on prior knowledge. Thus, a barrier space can be artificially set to exclude these actions, guiding the learning process to converge more efficiently and safely. {For instance, in reinforcement learning training, known unsafe states can be pre-included in $\mathcal{B}$}, preventing the agent from exploring these areas and improving learning efficiency and system reliability. By introducing the barrier space $\mathcal{B}$, constrained learning explicitly incorporates realistic limitations into the learning framework, allowing the agent to achieve system stability and optimization while satisfying all feasibility and safety conditions.

\subsubsection{Tracking Problems}

{\paragraph{(1) Definition of Tracking Problems}}

Assume we have a dynamical system with intrinsic features denoted by $s$. The output is $h(s)$, where $h(\cdot)$ is the output conversion function (in tracking problems, the system state $s$ is often bounded), and the tracking target is $\mathcal{Y}_{\mathrm{d}}$, which lies in the output space $\mathcal{Y}$ and is dimensionally compatible with $h(s)$. Therefore, tracking problems can be converted into stabilization learning problems, but the original six-tuple must be expanded into a new seven-tuple for tracking problems: $(\mathcal{S},\mathcal{A},\mathcal{P},\pi,h,\mathcal{Y}_{\mathrm{d}},d_\mathcal{Y})$. Here, the desired state set $\mathcal{S}_{\mathrm{d}}$ is replaced by the tracking target $\mathcal{Y}_{\mathrm{d}}$, and an output conversion function $h$ is added. The definitions are provided in Definition \ref{tracking-control}.

\begin{theorembox}{Transforming Tracking Problems into Stabilization Learning Problems}{tracking-control}\label{tracking-control}
When transforming a tracking problem into a stabilization learning problem, the elements $(\mathcal{S},\mathcal{A},\mathcal{P},\pi)$ retain their original definitions. The changed elements are explained as follows:
\begin{itemize}
  \item $\mathcal{Y}_{\mathrm{d}}$: \textbf{Desired Output Space}, representing the desired output of the system, i.e., the tracking target, where $\mathcal{Y}_{\mathrm{d}} \subseteq \mathcal{Y}$ and $\mathcal{Y}$ is the system output space;
  \item $h(\cdot)$: \textbf{Output Function}, $h:\mathcal{S}\to\mathcal{Y}$, mapping the system state to the output space;
  \item $d_\mathcal{Y}$: \textbf{Metric Function}, {$d_\mathcal{Y}:\mathcal{Y}\times\mathcal{P}(\mathcal{Y})\to \mathbb{R}_{\geq 0}$, where $ d_\mathcal{Y}=d(h(s),\mathcal{Y}_{\mathrm{d}}), \mathcal{Y}_{\mathrm{d}} \in \mathcal{P}(\mathcal{Y})$ expresses the distance between the system state output and the tracking target.}
\end{itemize}
Taking asymptotic tracking as an example, we need to find a specific policy $\pi$ such that when $t \rightarrow \infty$, $d\left(h\left(s\left(t\right)\right),\mathcal{Y}_{\mathrm{d}} \left(t\right)\right)\rightarrow 0$, meaning the system output approaches the target tracking trajectory while the system state $s$ remains bounded, i.e., $s \in \mathcal{L}_{\infty}$.\\ 

\qquad Tracking problems may also contain a barrier space $\mathcal{B}$, representing regions the system state cannot enter. In this case, the problem is comprehensively represented as an eight-tuple: $ (\mathcal{S},\mathcal{A},\mathcal{P},\pi,h,\mathcal{Y}_{\mathrm{d}},\mathcal{B},d_\mathcal{Y})$.
\end{theorembox}

In discussing whether an output tracking problem can be transformed into a stabilization problem, { a key concept} is the \textbf{relative degree}, here referring specifically to the input--output relative degree. Intuitively, the relative degree characterizes the number of differentiations required for the control input to propagate through the system dynamics and appear in the output. For a smooth control-affine system under a fixed operating mode,
\begin{equation}
\dot{s}=f(s)+g(s)a,\qquad y=h(s),
\end{equation}
where \(s\) is the system state, \(a\) is the control input, and \(y\) is the system output. The goal of a tracking problem is to design a control law or policy such that the system output \(y(t)\) approaches the desired output \(y_d(t)\) or the desired output set \( \mathcal{Y}_{\mathrm{d}}(t)\). If the control input \(a\) first appears explicitly after differentiating \(y\) continuously \(r\) times, then the output is said to have relative degree \(r\) near the corresponding state point \cite{khalil2002nonlinear}. More rigorously, if there exists a neighborhood \(\mathcal{U}(s_0)\) of a state point \(s_0\) and an integer \(r\) such that for any \(s\in\mathcal{U}(s_0)\),
\begin{equation}
L_gL_f^k h(s)=0,\quad k=0,\ldots,r-2,
\quad
L_gL_f^{r-1}h(s)\neq0,
\end{equation}
then the output \(y=h(s)\) has relative degree \(r\) near \(s_0\). Here, \(L_fh\) and \(L_gh\) denote the Lie derivatives of the function \(h\) along the vector fields \(f\) and \(g\), respectively.

When the nonzero property of the coefficient \(L_gL_f^{r-1}h(s)\), which determines whether the input appears explicitly, depends on the state \(s\), the system may exhibit situations in which the relative degree is \textbf{ not well-defined} at certain state points. Relative degree is not a global label independent of output selection and operating mode. When the input-to-selected-output relation does not satisfy a fixed-relative-degree condition, or when the operating mode switches, a finite relative degree in the classical sense may not exist, or may vary with the state. Physically, this means that the input no longer affects the output, or that the effect exerted by the input cannot be inferred through the output, so feedback cannot be used to correct the control action. Example \ref{exmaple:disk-pushing} below will illustrate this class of problems in which the relative degree varies with the state or task phase.

{\paragraph{(2) Transforming Tracking Problems into Stabilization Problems}}

The transformation of tracking problems into stabilization problems is the core logic of control system theory and engineering practice. Its essence is transforming the dynamic goal of ``output approaching the desired trajectory" into a stabilization problem of ``state or error signal converging to the origin, an equilibrium, or an invariant target set while remaining bounded," providing a quantifiable theoretical framework for solutions. The following three approaches are built upon different system characteristics and scenarios, forming a complementary and complete system with clear theoretical foundations and applicability boundaries.

\textbf{The first approach transforms tracking problems into stabilization problems through the ``direct mapping of expected output and system state".} Specifically, this method establishes a causal relationship between system states and outputs to derive an ideal state trajectory, thereby achieving stable tracking of the state. Its advantages are intuitive interpretation and the absence of additional model reconstruction, making it suitable for systems with low nonlinearity and relatively accurate modeling. Under this approach, the system output will track the ideal trajectory and ultimately reach a stable state.

\begin{examplebox}{Trimming Problem in UAV Flight Control}{trim-control}
The nonlinear dynamic model of small UAVs only possesses engineering solvability near the Trim Condition. Through linearization, the complex trajectory tracking problem can be transformed into a stabilization problem of state deviation. Specifically, the state of a UAV in flight can be represented as $[z\quad v\quad \alpha\quad \theta]^\top$ ($z$ for altitude, $v$ for velocity, $\alpha$ for angle of attack, and $\theta$ for pitch angle), while the desired equilibrium flight state is $s_\text{eq} = [z_\text{eq}\quad v_\text{eq}\quad \alpha_\text{eq}\quad \theta_\text{eq}]^\top$. By designing a controller to make the state deviation $\tilde{s} = s - s_\text{eq}$ stabilize to 0 after perturbations, robust tracking of the desired state is achieved. This method effectively simplifies complex trajectory tracking into a state deviation stabilization problem, ensuring the UAV stably maintains its predetermined attitude and trajectory \cite{beard2012small}.
\end{examplebox}

\textbf{The second approach relies on output linearization{ (more precisely, input--output linearization, Input--Output Linearization)} to transform the tracking problem into a stabilization problem.} Grounded in the theory of Lyapunov derivatives and relative degree, this method successively differentiates the output of the nonlinear system until the control input appears explicitly, employs nonlinear state feedback to cancel the system nonlinearities, and renders the input--output dynamics into a standard linear integrator chain. For minimum-phase systems, stable tracking can be achieved directly; non-minimum phase systems\footnote{Non-minimum phase systems \cite{khalil2002nonlinear} describe a class of systems with more complex dynamic behaviors. For Linear Time-Invariant (LTI) systems, they possess at least one transmission zero in the right half of the complex plane. This structural feature causes their dynamic response to exhibit an ``inherent contradiction": initially, the system output might move in the opposite direction of the desired final direction, known as ``inverse response" or ``non-minimum phase behavior." Furthermore, such systems usually suffer from larger phase lags, requiring cautious feedback control design to avoid stability deterioration.} suffer from unstable zero dynamics and thus require first constructing a minimum-phase virtual output and establishing a steady-state mapping, then linearizing the virtual output, thereby indirectly realizing tracking of the original output while guaranteeing global boundedness of the system states.

\begin{examplebox}{End-effector Trajectory Tracking Control of Flexible Manipulators}{flexible-arm}
Flexible manipulators are typical non-minimum phase systems. Dynamically, their transfer functions exhibit right-half-plane zeros (the essential feature of non-minimum phase behavior). If controllers are directly designed to track the desired trajectory of the end-effector, an ``inverse overshoot" phenomenon occurs: the torque input of the joint motor initially causes the end-effector to move opposite to the desired trajectory before adjusting toward the target, ultimately may lead to inverse response, large transient errors, oscillations, or even instability if the controller is not properly designed. To enhance output stability, practice usually avoids taking the ``desired end-effector trajectory" directly as the tracking target. Instead, a reconstructed reference signal coupling joint angles and elastic deformation compensation is built. The controller's goal shifts from ``tracking the non-minimum phase output of the end-effector" to stabilizing the reconstructed minimum phase output \cite{benosman2003stable}.
\end{examplebox}

\textbf{The third approach relies on the Internal Model Principle{ (IMP)}\footnote{The Internal Model Principle \cite{franciswonham1976imp,byrnes2004nonlinear,bin2022internal} is a core principle of robust tracking and disturbance rejection in modern control theory. It states that for a closed-loop control system to achieve asymptotic zero-error tracking of an external reference signal or complete asymptotic rejection of a sustained disturbance, the controller must contain an internal model capable of precisely reproducing the dynamic characteristics of that external signal. This idea reduces the servo mechanism to an embedded replication of the signal generation structure, theoretically guaranteeing zero steady-state error for specific signal patterns. It provides a fundamental theoretical basis for designing servo systems and modern robust controllers.}, which embeds the desired signal generation mechanism into the closed-loop system, ensuring stable tracking of the target output without external signal disturbances.} Ultimately, the system tracking problem is transformed into a system stabilization problem by embedding a model of the exogenous reference or disturbance signal into the controller. The
closed-loop system can achieve robust tracking or disturbance rejection. After system stabilization, it can automatically adjust and maintain a stable state regardless of environmental changes while performing tracking. The most common example is PID control{ (Proportional--Integral--Derivative Control)}, where the integral term is a constant signal embedded into the closed-loop to track and compensate for external constant signals. {It should be noted that the internal model principle cannot replace the physical interaction structure of the system itself. If the fixed input--output relative degree itself is no longer well-defined, that is, if a stable causal relation between the input and the output has not yet been established and the control input cannot affect the target output, then even adding an integral term to the controller cannot achieve stable tracking of that output. The disk-pushing task in Example \ref{exmaple:disk-pushing} below is a typical example of this type of problem: before the passive disk is contacted, no amount of integral action can move it.}

{
\paragraph{(3) Transforming a Class of Problems with State- or Mode-Dependent Relative Degree}

The aforementioned three approaches typically assume that the input--output relationship is well-defined within the considered operating region. When the relative degree varies with the state or operational mode, the transformation of the tracking problem becomes significantly more complex. Robotic manipulation control tasks serve as a typical example of such problems: the target {controlled} object is often not the robot itself, and only when the robot and the target {controlled} object establish an interaction relationship capable of influencing the target object's motion will the control input affect the target object's output. Consequently, such tasks cannot be simply regarded as standard tracking problems with a fixed relative degree, but should instead be understood as coupled {``motion--manipulation''} problems, wherein the effect of the input on the target output varies according to the state or mode.

\begin{examplebox}{{State- or Mode-Dependent Relative Degree in a Disk-Pushing Task}}{disk-pushing}
\label{exmaple:disk-pushing}
Consider a two-disk pushing task on a plane. Disk 1 is an active disk and can be driven by the robot end-effector; Disk 2 is a passive disk and has no actuation capability of its own. The task requires the active disk to start from its initial position, approach the passive disk, and then push the passive disk so that it eventually reaches the target position. Let \(p_1\) and \(p_2\) denote the positions of the two disks, respectively. The control input can be understood as the motion command of the active disk, while the final task objective is
\begin{equation}
p_2(t)\rightarrow p_2^\ast .
\end{equation}

The key difficulty of this task is that the input--output relative degree changes across phases. In the motion-control phase of Disk 1, the control input directly changes the position of the active disk. However, at this stage, the passive disk has not yet formed an interaction relation with the active disk that can influence its motion. If the position of the passive disk is selected as the output, then this phase is not suitable for describing the relation between the control input and the passive-disk position by a fixed relative degree. In the manipulation-control phase of Disk 1, after the active disk pushes the passive disk, the control input affects the passive-disk position through the interaction relation between the disks. Therefore, the relative degree of the complete task is not a fixed quantity, but depends on the system state and the task phase.
\end{examplebox}

\begin{figure}[H]
  
  \centering
  \begin{tikzpicture}[
    diskA/.style={circle, draw=blue!70!black, fill=blue!15, thick, minimum size=1.1cm},
    diskB/.style={circle, draw=orange!80!black, fill=orange!20, thick, minimum size=1.1cm},
    target/.style={circle, draw=green!50!black, dashed, thick, minimum size=1.25cm},
    arr/.style={-{Stealth[length=3mm,width=2mm]}, thick, draw=black!65},
    label/.style={font=\small}
  ]
    \node[label] at (-1.0,2.05) {\textbf{Motion-Control Phase}};
    \node[diskA] (a1) at (-2.7,0.7) {1};
    \node[diskB] (b1) at (0.7,0.7) {2};
    \draw[arr] (a1) -- node[above,label] {approach} (b1);
    \node[label, blue!70!black] at (-2.7,1.55) {$p_1$};
    \node[label, orange!80!black] at (0.7,1.55) {$p_2$};
    \node[label] at (-2.7,-0.05) {active disk};
    \node[label] at (0.7,-0.05) {passive disk};

    \node[label] at (5.0,2.05) {\textbf{Manipulation-Control Phase}};
    \node[diskA] (a2) at (3.3,0.7) {1};
    \node[diskB] (b2) at (4.55,0.7) {2};
    \node[target] (goal) at (6.55,0.7) {};
    \fill[black!75] (3.925,0.7) circle (1.7pt);
    \node[label] at (3.925,-0.05) {contact};
    \node[label, blue!70!black] at (3.3,1.55) {$p_1$};
    \node[label, orange!80!black] at (4.55,1.55) {$p_2$};
    \draw[arr] (b2) -- node[above,label] {push} (goal);
    \node[label] at (6.55,-0.05) {target position};
  \end{tikzpicture}
  \caption{Two-phase illustration of the disk-pushing task}
  \label{fig:disk_pushing_relative_degree}
\end{figure}
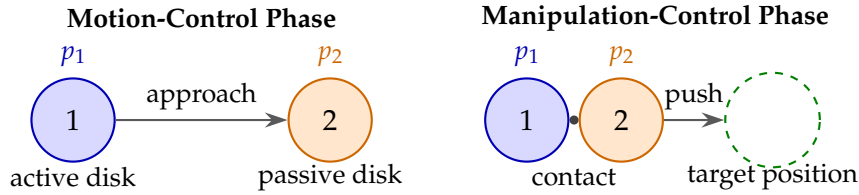

For this class of problems in which the relative degree varies with the state or mode, one possible treatment is \textbf{state augmentation}. The idea is to consider, within a finite prediction horizon, the control-input sequence, predicted state sequence, and output sequence simultaneously, rather than checking only whether the current input can directly change the current output. The augmented predicted state, control input, and output sequences can be constructed as
\begin{equation}
\begin{aligned}
{S_{t:t+N-1}} &= {(s_t,s_{t+1},\ldots,s_{t+N-1})},\\
{A_{t:t+N-1}} &= {(a_t,a_{t+1},\ldots,a_{t+N-1})},\\
Y_{t:t+N-1} &= {(y_t,y_{t+1},\ldots,y_{t+N-1})}.
\end{aligned}
\end{equation}
Correspondingly, the evolution relation within the finite prediction horizon can be summarized as
\begin{equation}
S_{t+1:t+N}=\mathcal{F}(S_{t:t+N-1},A_{t:t+N-1}),\qquad
Y_{t:t+N-1}=\mathcal{H}(S_{t:t+N-1}).
\end{equation}
Here, \(\mathcal{F}\) denotes the augmented state-transition mapping that describes the recursive relation between adjacent states within the finite prediction horizon, given the predicted state sequence \(S_{t:t+N-1}\) and the control-input sequence \(A_{t:t+N-1}\). The mapping \(\mathcal{H}\) denotes the augmented output mapping, which applies the single-step output function \(h\) pointwise to the predicted states corresponding to the control-input sequence. In this way, the policy or optimizer no longer focuses only on the one-step input--output relation at the current time, but can arrange the continuous process of ``approaching the target object" and ``pushing the target object" within the same planning--control problem, thereby alleviating the difficulty caused by phase-dependent relative degree. It should be emphasized that although the control-input sequence \(A_{t:t+N-1}\) is introduced for prediction and planning, the actual closed-loop execution usually applies only the first control input \(a_t\) at the current time, and reconstructs the prediction sequence and replans at the next time according to the updated state feedback. This is consistent with the closed-loop execution mechanism of Receding Horizon Control. Furthermore, output linearization can be used to transform the tracking problem into a stabilization problem.

{It is thus evident that there exist both connections and distinctions between motion control problems and manipulation control problems. In motion control problems, the control input typically acts directly on the robot's own state, resulting in a well-defined input--output relationship. In contrast, for manipulation control problems, the target object is not the robot itself, and the control input can only influence the target output through the interaction relationship between the robot and the target object; consequently, the relative degree is more susceptible to variations with the state or mode. Both can be formulated as tracking problems where the ``output approaches a target trajectory or a target set'', and can be further extended to more complex manipulation control tasks, such as the Push-T task, as discussed in detail in Section \ref{sec:push-t} of this paper.}
}

\subsection{Analysis of Key Elements in Stabilization Learning Problems}

\subsubsection{State Space and Desired State Set}

The state is a mathematical expression sufficient to describe the system's evolution and to
predict future behavior under given inputs, possibly formulated as vectors, sets, or probability distributions. In stabilization learning problems, $s$ describes the system state initially and after taking a step during evolution, and $\mathcal{S}_{\mathrm{d}}$ refers to the set of all possible optimization target points. Specifically, during the transformation of concrete problems into stabilization learning problems, two typical scenarios are often encountered: structured feedback data scenarios and unstructured feedback data scenarios \cite{siddiqa2017unstructured}.

\paragraph{(1) Structured Feedback Data Scenarios}

Controlled variables can typically be characterized through well-defined mathematical models or explicit state indicators, such as the stability of an aircraft's flight altitude, the convergence behavior of iterative algorithms, and the consistency of product performance. The target states of such systems are often directly represented by specific numerical quantities or statistical measures, including velocity vectors, minima of loss functions, and performance variances. In this context, the stability of the control problem can be formulated as the bounded convergence and asymptotic stability of system state variables under prescribed constraints. The central objective is to design appropriate control strategies and stabilization-oriented learning mechanisms that ensure the system states evolve within a predefined feasible region while progressively approaching the desired equilibrium state. In this way, performance optimization can be achieved while maintaining the dynamical stability of the system and preventing undesirable behaviors such as instability, oscillation, or divergence.

\paragraph{(2) Unstructured Feedback Scenarios}

Systems often lack explicit mathematical models, and their state features are difficult to extract directly. For example, temporal coherence of video output in generative models, multi-view object recognition consistency in detection models, and context logic coherence in dialogue systems have system states which are often hidden within high-dimensional feature spaces. To characterize the states of such systems, deep feature extraction techniques are often required, such as using hidden layer activations of neural networks or manifold projections of generated results in semantic space. In such scenarios, achieving stability relies not only on foundational model training but also requires auxiliary techniques like data augmentation \cite{shorten2019survey} and metric learning \cite{hadsell2006dimensionality}, or adopting network structures with inherent ``invariance" (like Convolutional Neural Networks (CNNs)) to enhance robustness. These methods jointly support stabilization learning in complex unstructured environments, allowing systems to maintain consistent and reliable output behaviors despite input perturbations and condition changes.

Thus, regardless of the data scenario, the key to building a stabilization learning network and guiding the system to produce stable feedback lies in the \textbf{identification and extraction of the system's deep essential features}. These essential features are embedded in the internal structure and operational mechanisms of the system, acting as fundamental factors dictating its dynamic behavior. Especially in unstructured system scenarios, due to the lack of explicit mappings between system output and external regulation, surface interventions cannot directly adjust external behaviors. Therefore, one must first identify and extract intrinsic features reflecting the system's eigenproperties from high-dimensional, chaotic raw data to accurately predict and effectively regulate the system's feedback behaviors. Feature extraction is arguably the logical starting point and foundational prerequisite for achieving data-driven stabilization learning.

\paragraph{(3) Feature Extraction}

Regarding feature extraction methods, alongside classical dimensionality reduction and decoupling techniques like Principal Component Analysis (PCA) and Independent Component Analysis (ICA) \cite{vandermaaten2009dimensionality}, a feature selection paradigm based on \textbf{invariance learning} \cite{zhao2022fundamental} has developed in recent years. Classical methods aim to map original high-dimensional inputs to lower-dimensional feature spaces through orthogonal constraints to eliminate redundancy and coupling, {thereby ensuring statistical independence, or approximate independence, among feature components at the representation level}. { Constructing orthogonal features not only improves model generalization but also provides a foundation for identifying endogenous independent mechanisms under the causal inference framework, thereby enhancing the learning system's stability against data-distribution shifts and external disturbances}. However, during complex data-driven modeling, models easily learn ``spurious correlations," such as backgrounds in image classification or specific word frequencies in text. While statistically significant in training sets, these features have no causal link to outputs and severely harm model generalization on out-of-distribution data. To solve this problem, invariant feature learning mechanisms can be introduced to actively identify and suppress spurious correlations while preserving and aligning causal essential features across environments and data distributions, constructing highly robust and interpretable stabilization learning models.

In complex system analysis, the \textbf{Latent Space} \cite{bengio2013representation} acts as a core intermediate representation. By mapping high-dimensional data into low-dimensional embeddings, it achieves an efficient and structured feature extraction mechanism. This method usually leverages deep learning models (like Autoencoders or Variational Autoencoders \cite{kingma2014auto}) for nonlinear dimensionality reduction and information condensation, capturing potential essential features and structures within the data. This feature not only strips noise and redundancy from observational data but also reveals abstract properties, dynamic patterns, and key variables invisible on the surface, providing a low-dimensional, expressive feature foundation for subsequent modeling, state identification, and interpretability analysis.

\subsubsection{Plant}

\paragraph{(1) Dynamical System} 

A dynamical system refers to a mathematical model in which the system state evolves over time in a time series. It usually consists of state variables and dynamic equations (such as differential equations or state transition matrices) that describe state evolution, including both deterministic and stochastic models. The models we construct in stabilization learning problems such as control, observation, and recognition are essentially modeling dynamical systems.

Based on the modeling method, dynamical systems can be primarily divided into two major paradigms: mechanism modeling based on first principles and non-first-principle modeling based on data learning. The former, first-principle modeling, is rooted in physical laws and conservation equations that describe the fundamental laws of nature. For example, in the realm of classical mechanics, whether through vector analysis in Newtonian mechanics (force-momentum relationship) or scalar analysis in Lagrangian mechanics (energy-work relationship), the core lies in utilizing prior physical knowledge to derive ordinary or partial differential equations that govern system behaviors. Such models possess clear physical interpretability and are suitable for theoretical scenarios with clear mechanisms and measurable variables. The latter, non-first-principle modeling, relies on external observation data generated during system operation when the internal mechanisms of the system are too complex or not fully understood. This method does not presuppose specific physical laws but treats the system as a ``black box" or ``gray box," leveraging data-driven technologies such as machine learning and system identification to directly learn and reconstruct the dynamic evolutionary laws of the system from massive data. Such models excel at handling high-dimensional, nonlinear problems; their efficacy highly depends on the quality and scale of the data, and they exhibit significant advantages in complex systems where mechanism models are difficult to establish.

From the perspective of the mathematical structure of system dynamic characteristics, dynamical systems can be classified into linear systems, nonlinear systems, and hybrid systems. \textbf{Linear Systems} are described by linear differential or difference equations, and their core characteristic is satisfying the principle of superposition, meaning the system's response to the sum of multiple inputs equals the sum of its responses to individual inputs. This characteristic endows linear systems with a mature and complete theoretical framework for analysis and synthesis (such as transfer functions, state-space methods, and controllability/observability criteria), and their solutions can usually be expressed analytically. In contrast, \textbf{Nonlinear Systems} are ubiquitous in the real world; their dynamic equations contain nonlinear terms (such as trigonometric functions, product terms, etc.) of state or input variables, thus failing to satisfy the superposition principle. Nonlinear systems exhibit dynamic behaviors far richer and more complex than linear systems, such as multi-stability, limit cycles, bifurcation, and chaos, and their analysis typically relies on phase plane methods, Lyapunov stability theory, or numerical simulations. In practical applications, certain nonlinear systems (such as piecewise smooth systems or nonlinear systems exhibiting local linear characteristics) can be alternatively modeled as Linear Gaussian Mixture Systems \cite{roweis1999unifying,ghahramani2000mixture}. Its core mechanism is to approximate or characterize the global nonlinear dynamic characteristics of the system through the combination of several linear sub-systems, where switching between sub-systems follows specific probability rules usually based on Gaussian distribution assumptions. This model elegantly combines the analytical convenience of linear systems with the expressive power of probabilistic systems in theory, making it particularly suitable for describing multi-modal stochastic processes or performing estimation (like target tracking) and control on complex systems with uncertain jump characteristics. Distinct from the simple division of linear and nonlinear systems, \textbf{Hybrid Systems} \cite{goebel2009hybrid} are more complex, containing both continuous and discrete dynamical systems that interact with each other, potentially operating across multiple temporal and spatial scales. Analyzing hybrid systems often requires the use of multiple Lyapunov functions or switched Lyapunov functions, constructing energy functions for each mode separately and providing global stability criteria in conjunction with jump conditions.

\paragraph{(2) State Transition Probability}

In stabilization learning problems, scenarios involving unstructured data are often modeled as dynamical systems of sequential decision processes, whose core feature lies in the evolutionary behavior of the system state over time. Such systems are widely present in scenarios like robot navigation, autonomous driving, and large language models, where system dynamics can be precisely characterized by \textbf{State Transition Probabilities}. This probability model describes the conditional distribution of the system's next state given the current state and the agent's action, namely $\mathbb{P}(s_{t+1} | s_t, a_t)$. By learning and estimating the state transition probabilities of Markov Decision Processes through Bayesian inference and Monte Carlo methods, the agent can effectively predict uncertain environments and optimize decisions, thereby achieving the goal of maximizing long-term cumulative rewards.

The process of a system evolving between different states via state transition probabilities can be described by Markov process theory, whose core lies in the mathematical description of cumulative state transition probabilities. This theory assumes that the future state of the system at any given moment depends only on the current state and is independent of historical states, a property known as the Markov property. Specifically, the state transition probability is defined as the conditional probability of the system transitioning from the current state $i$ to the next state $j$, denoted as $P_{ij}$. These probabilities are usually systematically represented by a transition probability matrix $\mathbf{P}=[P_{ij}]$, where the matrix elements satisfy non-negativity and the constraint that row sums equal 1, i.e., {$P_{ij}\geq 0, \sum_j P_{ij}=1$}. Long-term multi-step state transition probabilities are described by the Chapman-Kolmogorov equation:
\begin{equation}
P_{ij}^{(m+n)} = \sum_k P_{ik}^{(m)}P_{kj}^{(n)}.
\end{equation}
State transition probabilities and Markov processes are particularly important in dynamical systems such as game problems. The reason is that this type of system can characterize the uncertainty and dynamic evolutionary features faced by participants during sequential decision-making processes through a discrete state space and corresponding probabilistic transition rules. Taking board games as an example, all possible situations on the board constitute a finite state set, and each legal move can be viewed as a random event inducing a transition between states. Its state transition probability is jointly determined by the distribution of pieces (system state) and the opponent's operations (trend of system state changes influenced by external factors). In complete-information two-player zero-sum games (like Go or Chess), if the opponent's coping strategy is modeled as a stochastic process, the entire game can be represented as a type of stochastic process with the Markov property—that is, the next state only depends on the current state and the chosen action, regardless of the historical state path. Under this framework, the goal of the game participants can be transformed into finding an optimal policy that maximizes cumulative expected returns within this Markov decision process, while the stabilization learning problem aims to enhance the stability of the final result converging to the optimum.

\paragraph{(3) World Model}

A World Model \cite{ding2024understanding} is a generative artificial intelligence model capable of simulating the real-world environment and predicting the future based on text, images, or videos, and is considered a crucial technological route for embodied AI to become reality in the future.

From a control perspective, a world model is essentially the joint identification of system dynamics and observation models, functioning similarly to an internal model in uncertain systems. A world model first needs to extract representations; based on different State Representations, they can be divided into five main forms: Pixels, Latent variables, 3D Particles, Keypoints, and Object-centric representations. Representations affect the model's expressive capacity, sample efficiency, generalization, and computational cost. These dynamic models can be combined with motion planning (like trajectory optimization) and policy learning (like reinforcement learning) to achieve complex tasks such as object repositioning and flexible object manipulation \cite{ai2025review}. Neural ODE \cite{chen2018neural} proposed generalizing traditional discrete-layer neural networks into continuous deep models—using neural networks to parameterize the time derivative of system states and obtaining the state's evolution over time through a black-box ODE solver, thereby transforming the learning problem into a task of fitting continuous dynamic fields. This method efficiently computes gradients for ODE solutions via the adjoint sensitivity method, significantly reducing the memory overhead of backpropagation and enabling control over numerical errors; it also brings advantages like adaptive computation (number of evaluations changes with input complexity) and parameter coupling (adjacent ``layer" parameters are automatically shared). Based on this framework, Continuous Normalizing Flows for invertible density modeling and Latent ODEs for modeling, interpolating, and extrapolating irregularly sampled time-series data were developed, thus providing a general and scalable toolbox for data-driven learning of system dynamics, time-series modeling, and generative modeling.

Explained from the perspective of intelligence technology, a world model is an internal computational framework constructed by an agent to understand its environment, with its core function being to predict and simulate dynamic changes in the environment. Simply put, it acts like a ``simulator" within the agent's ``brain," capable of deducing possible future state sequences and their outcomes based on the current state and the actions it might take. This model allows the agent to conduct ``thought experiments" internally without costly real-world trial and error, thereby achieving action planning, decision optimization, and reasoning about unknown situations. Therefore, world models are regarded as the cornerstone for achieving general intelligence; by compressing and abstracting sensory experiences, they transform the complex external world into a computable, operable internal representation, serving as a key cognitive component for agents to achieve purposeful behavior.

Within the theoretical framework of intelligent systems based on reinforcement learning and generative modeling, a world model is essentially an internal probabilistic abstraction of the environmental dynamics in which the agent is situated, and can be viewed as a highly comprehensive ultimate dynamic model. Its core mathematical representation is realized precisely through the state transition probability $\mathbb{P}(s_{t+1}|s_t,a_t)$, which defines the stochastic rules by which the environment transitions to the next state $s_{t+1}$ given the current state $s_t$ and the agent's action $a_t$. Therefore, a world model that has learned the true environmental dynamics is equivalent to accurately approximating the state transition probability function in this underlying Markov decision process, thereby empowering the agent to perform Monte Carlo predictions for policy evaluation and sequential decision-making in internal simulations. \textbf{In this sense, generative models do not replace control theory, but provide a learnable model foundation for it}.

\paragraph{(4) Controllability and Observability}

Controllability and observability are fundamental structural properties of dynamical systems and serve as essential prerequisites for the effective operation of closed-loop mechanisms in stabilization learning. For a general dynamical system, \textbf{controllability} refers to the existence of a finite-time control input capable of driving any initial state to an arbitrary target state within the state space; \textbf{observability} implies that the initial state of the system can be uniquely determined through finite-time output observations. These two properties respectively define the boundaries of external intervention and internal perception of the system. If a system is uncontrollable, no global stabilization strategy exists; if it is unobservable, it is impossible to construct effective state feedback.

For linear systems, the degrees of controllability and observability can be quantified by metrics such as the minimum eigenvalues and condition numbers of the controllability and observability Gramian matrices. These metrics respectively reflect the control energy required to drive the system and the impact of observation noise on state estimation. For nonlinear systems, generalized degrees of controllability and observability can be defined using tools such as controllability and observability Lyapunov functions, which characterize the extent to which the system is controllable and observable across different state regions. A low degree of controllability typically leads to control input saturation or slow convergence, while a low degree of observability amplifies observation errors. Both issues severely undermine the robustness of stabilization learning.

Within the framework of stabilization learning, \textbf{feature extraction} is deeply coupled with the controllability and observability of the controlled plant. While feature extraction in traditional machine learning primarily aims at data compression, feature extraction in stabilization learning must prioritize preserving the controllability and observability of the system. In practical feature extraction processes, the degrees of controllability and observability can serve as regularization constraints. By incorporating penalty terms related to these metrics into the loss function, neural networks can be guided to actively retain core modes that are strongly correlated with system dynamics, thereby filtering out irrelevant and redundant features such as textures and backgrounds. For instance, in the vision-based UAV position stabilization task which in Section \ref{sec:drone_hover}, incorporating observability regularization compels the network to extract geometric features relevant to the pose; in the Push-T manipulation task which in Section \ref{sec:push-t}, applying controllability regularization enhances the learning of features related to contact dynamics, ensuring that the end-effector can effectively drive the motion of the object.

\subsubsection{Metric}

A metric is a function that measures the distance between the current state and the desired state, typically defined as $d(s,\mathcal{S}_{\mathrm{d}})$, where $s\in\mathcal{S}$ is the current state and $\mathcal{S}_{\mathrm{d}} \subseteq \mathcal{S}$ is the desired state set (which may be a single point or a set). The process of solving a stabilization learning problem is precisely the process of designing a policy to { drive the metric to zero}, ultimately causing the state to fall into the desired state set. Depending on different application scenarios, the metric can adopt different definitions to better guarantee the effectiveness of the stabilization learning policy. \textbf{For the remainder of this paper, unless a specific measure is explicitly indicated, the notation \( \| \cdot\| \) will be understood generically to refer to any appropriate measure.}

\paragraph{(1) Point-to-Point Metrics}

For stability problems where data is structured and optimization goals are clear (where quantitative indicators $x=(x_1,x_2,\cdots,x_n)$ and a definite optimization target point $y=(y_1,y_2,\cdots,y_n)$ ($x,y \in\mathbb{R}^n$) can be extracted), it is only necessary to ensure the quantitative indicator $x$ evolves in the direction of $y$. The distance during this process is the distance between two points in Euclidean space. Commonly used metrics include:

\begin{itemize}[leftmargin=*, itemsep=2pt]
  \item \textbf{Euclidean Distance}: $||x-y||_2=\sqrt{\sum_{i=1}^n (x_i-y_i)^2}$
  \item \textbf{Manhattan Distance}: $||x-y||_1=\sum_{i=1}^n |x_i-y_i|$
  \item \textbf{Chebyshev Distance}: $||x-y||_\infty=\max_{i=1,\cdots,n} |x_i-y_i|$.
\end{itemize}

\paragraph{(2) Point-to-Set Metrics}

When the quantitative indicator is a vector and the desired state set is a collection of multiple possible target states, one can consider using the distance between a point and a set to measure the stability state of the problem, such as using the minimum distance:
\begin{equation}
d(s,\mathcal{S}_{\mathrm{d}})=\min_{s_{\mathrm{d}} \in \mathcal{S}_{\mathrm{d}}} ||s-s_{\mathrm{d}}||.
\end{equation}
This distance is generally used for nearest neighbor search, i.e., finding the target state closest to the current state. Alternatively, the maximum distance can be used:
\begin{equation}
d(s,\mathcal{S}_{\mathrm{d}})=\max_{s_{\mathrm{d}} \in \mathcal{S}_{\mathrm{d}}} ||s-s_{\mathrm{d}}||.
\end{equation}
This distance is generally used for anomaly detection, i.e., finding the maximum distance from normal states. Or the average distance can be adopted:
\begin{equation}
d(s,\mathcal{S}_{\mathrm{d}})=\frac{1}{|\mathcal{S}_{\mathrm{d}}|}\sum_{s_{\mathrm{d}} \in \mathcal{S}_{\mathrm{d}}} ||s-s_{\mathrm{d}}||,
\end{equation}
where $|\mathcal{S}_{\mathrm{d}}|$ is the measure of the desired state set. The definition of this metric is generally used to estimate the probability that the system's current state falls within the category of the target set.

\paragraph{(3) Set-to-Set Metrics}
When both the current state and the desired state are sets, one can consider using distances between sets to measure the stability state of the problem, such as the Hausdorff Distance \cite{deza2016encyclopedia}, which is used to measure the maximum-minimum distance between two sets and is commonly used in shape matching. At this time, the system state is specifically denoted as $S$($S \in \mathcal{S}$):
{
\begin{equation}
d_H(S,\mathcal{S}_{\mathrm{d}})=\max\left\{\sup_{s\in S}\inf_{s_{\mathrm{d}}\in\mathcal{S}_{\mathrm{d}}}\|s-s_{\mathrm{d}}\|,\sup_{s_{\mathrm{d}}\in\mathcal{S}_{\mathrm{d}}}\inf_{s\in S}\|s-s_{\mathrm{d}}\|\right\}.
\end{equation}}
{
When emphasizing the degree of similarity between sets, the Jaccard similarity coefficient \cite{deza2016encyclopedia} can be defined based on the intersection and union of the two sets:
\begin{equation} 
J(S,\mathcal{S}_{\mathrm{d}})=\frac{|S\cap \mathcal{S}_{\mathrm{d}}|}{|S\cup \mathcal{S}_{\mathrm{d}}|},
\end{equation}
where $ |S\cap \mathcal{S}_{\mathrm{d}}|$ is the number (or measure) of intersection elements, and $ |S\cup \mathcal{S}_{\mathrm{d}}|$ is the number (or measure) of union elements. The corresponding Jaccard distance is
\begin{equation}
d_J(S,\mathcal{S}_{\mathrm{d}})=1-J(S,\mathcal{S}_{\mathrm{d}}).
\end{equation}
The Jaccard distance takes values in the range $[0,1]$, with lower values indicating greater similarity between the two sets.}

Additionally, for the optimal transport problem between two sets, the Wasserstein distance \cite{deza2016encyclopedia} can be adopted to measure the minimum-cost transport distance between the two sets.

Define the two non-empty sets to be measured as the source set $\mathcal{X}$ and the target set $\mathcal{Y}$, respectively:

\begin{itemize}[leftmargin=*, itemsep=2pt]
  \item Source set: $\mathcal{X} = \{x_1, x_2, \ldots, x_n\}$, where $\sum_{i=1}^n x_i = 1 , x_i \geq 0$.
  \item Target set: $\mathcal{Y} = \{y_1, y_2, \ldots, y_m\}$, where $\sum_{j=1}^m y_j = 1 , y_j \geq 0$.
\end{itemize}

For the transport process, the transport matrix $T \in \mathbb{R}^{n \times m}_{\geq 0}$ provides the transport scheme (where $T_{ij}$ is the amount transported from the $i$-th source sample to the $j$-th target sample), and the cost matrix $C \in \mathbb{R}^{n \times m}$ characterizes the transport cost (where $C_{ij}$ is the cost from the $i$-th source sample to the $j$-th target sample). The Wasserstein distance can be defined as:
\begin{equation} 
W(\mathcal{X}, \mathcal{Y}) = \min_{T \in \mathbb{R}^{n \times m}_{\geq 0}} \sum_{i=1}^n \sum_{j=1}^m C_{ij} T_{ij}.
\end{equation}
The Wasserstein distance takes values in the range $[0,\infty)$, where smaller values indicate lower transport costs between the two sets, characterizing the metric between two sets in a specific task space.

Furthermore, for special data structures like text, images, and audio, distance metrics based on characters, pixels, or samples can be used, such as Edit distance, Cosine similarity, and Dynamic time warping.

\paragraph{(4) Probability Distribution Metrics}

The previously considered situations are all based on the system state being located in a discrete space, but in some cases, the state space is continuous and needs to be described using probability distributions and other methods. When both the current state and the desired state are probability distributions, one can consider using distances between probability distributions to measure the stability state of the problem, such as the Kullback-Leibler (KL) divergence \cite{deza2016encyclopedia}, which is used to measure the difference between two probability distributions:
\begin{equation}
D_{KL}(P||Q)=\sum_{x\in\mathcal{X}} P(x)\log\frac{P(x)}{Q(x)},
\end{equation}
where $P(x)$ is the current state probability distribution, $Q(x)$ is the desired state probability distribution, and $\mathcal{X}$ is the state space. The KL divergence takes values in the range $[0,\infty)$, with smaller values indicating greater similarity between the two probability distributions. The KL divergence is not a distance in the traditional sense, but rather a means to holistically measure the discrepancy between two probability distributions.

Note that the KL divergence is asymmetric, i.e., $D_{KL}(P \mid \mid Q) \neq D_{KL}(Q \mid \mid P)$; symmetrizing it yields the Jeffreys divergence \cite{deza2016encyclopedia}:
\begin{equation}
J(P, Q) = D_{KL}(P \mid \mid Q) + D_{KL}(Q \mid \mid P).
\end{equation}

Simultaneously symmetrizing and bounding the KL divergence yields the Jensen-Shannon (JS) divergence \cite{deza2016encyclopedia}:
\begin{equation}
J(P, Q) = \frac{1}{2} D_{KL}(P \mid \mid M) + \frac{1}{2} D_{KL}(Q \mid \mid M), \, M = \frac{1}{2}(P + Q),
\end{equation}
where $M$ is the average distribution of $P$ and $Q$. { The JS divergence takes values in $[0,\ln 2]$ under the natural logarithm, or in $[0,1]$ under the base-2 logarithm}, with smaller values indicating greater similarity between the two probability distributions. The KL divergence and its variants are often applied in scenarios such as likelihood estimation, coding, and variational methods.

In scenarios like hypothesis testing, $L_p$ and $\chi^2$ class distribution metrics \cite{deza2016encyclopedia} are frequently used. The $L_p$ metric tends to measure the absolute distance between the sample and the expected state:
\begin{equation}
L_p(P,Q) = (\int |p(x) - q(x)|^p \text{d}x)^{1/p},1\leq p < \infty.
\end{equation}
While the $\chi^2$ class distribution metric focuses more on the degree of difference between the sample and the expected state, especially for the detection of tails and outliers:
\begin{equation}
\chi^2(P||Q) = \int\frac{(p(x) - q(x))^2}{q(x)}\text{d}x = \int \frac{p^2(x)}{q(x)}\text{d}x - 1.
\end{equation}
Meanwhile, there is also an optimal transport problem between two probability distributions, from which the Wasserstein distance between probability distributions can be defined.

Define the two probability distributions to be measured as the source probability distribution $(\mathcal{X}, \mu)$ and the target probability distribution $(\mathcal{Y}, \nu)$:

\begin{itemize}[leftmargin=*, itemsep=2pt]
\item Source probability distribution $(\mathcal{X}, \mu)$, satisfying the normalization condition 
$$\int_{x\in\mathcal{X}} \mu(x) \text{d}x = 1.$$
\item Target probability distribution $(\mathcal{Y}, \nu)$, satisfying the normalization condition 
$$\int_{y\in\mathcal{Y}} \nu(y) \text{d}y = 1.$$
\end{itemize}

{
Given the push-forward condition $T_{\#}\mu=\nu$ and the cost function $c:\mathcal{X}\times\mathcal{Y} \to \mathbb{R}_{\geq 0}$, the corresponding Wasserstein distance can be determined as:
\begin{equation}
W(\mu,\nu)=\inf_{T:\,T_{\#}\mu=\nu}\int_{\mathcal{X}}c(x,T(x))\,\mathrm{d}\mu(x).
\end{equation}}

\paragraph{(5) Metrics Based on Lyapunov Functions}

Although the above metric methods fully consider the distance between the system state and the desired state set, they are mostly built upon the assumption of an unconstrained free state space. However, in real physical systems and control engineering such as robot kinematics and aerospace, the state space is often subject to various physical constraints and dynamic limitations, possessing complex topological structures and feasibility boundaries. In such constrained spaces, the Lyapunov function, as a stability analysis tool combining the physical core of the system and spatial geometric structures, can effectively characterize the dynamic migration behavior of the system state towards the desired state set.

Specifically, let $V: \mathcal{S} \to \mathbb{R}_{\geq 0}$ be a Lyapunov function used to measure the ``generalized distance" between the system state $s \in \mathcal{S}$ and the desired state set $\mathcal{S}_{\mathrm{d}}$. If a quadratic form is adopted, it can be defined as:
\begin{equation} 
V(s) = (s - s_{\mathrm{d}})^\top P(s - s_{\mathrm{d}}), \quad \forall s \in \mathcal{S},
\end{equation}
where $ s_{\mathrm{d}} \in \mathcal{S}_{\mathrm{d}}$ is the desired state, and $P$ is a symmetric positive definite matrix used to weight the importance of different state dimensions in the metric. This function satisfies $V(s) \geq 0$, and equals zero if and only if $s = s_{\mathrm{d}}$.

\begin{figure}[H]
  \centering
  \includegraphics[width=1\textwidth]{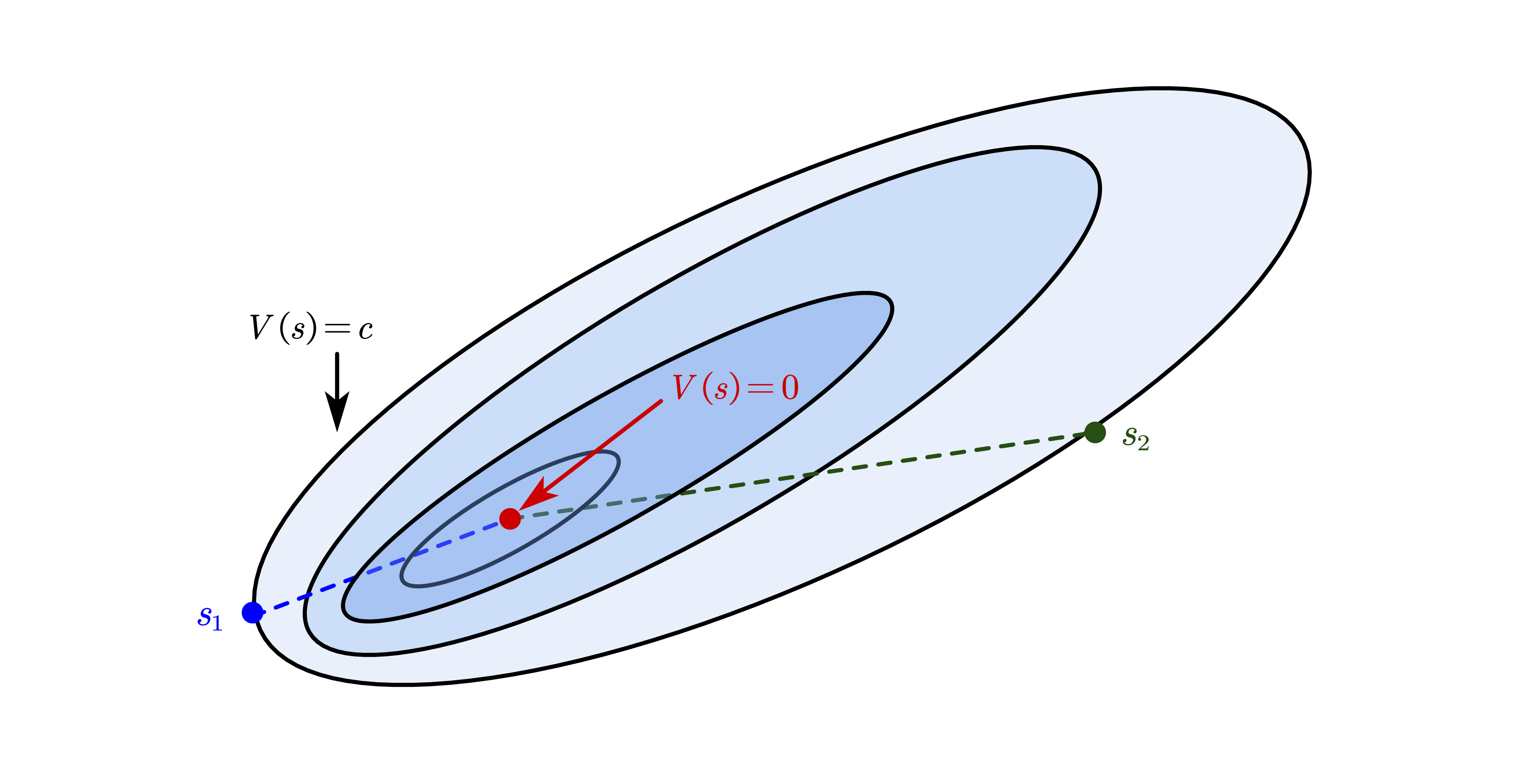}
  \caption{Metric based on Lyapunov function}
  \label{fig:Lyapunov distance}
\end{figure}

Figure \ref{fig:Lyapunov distance} reveals the difference between Euclidean distance and the distance based on the Lyapunov function. The figure contains the origin $0$ and points $s_1, s_2$. In terms of Euclidean distance, the distance from $s_2$ to the origin is greater than the distance from $s_1$ to the origin; however, under the metric based on the Lyapunov function, because $V(s_1) = V(s_2)=c$, the distances of the two points to the origin are the same. Both the Euclidean distance and the distance in the sense of the Lyapunov function belong to the concept of distance, and when the state $s$ approaches $0$, both their Euclidean distance and the distance in the sense of the Lyapunov function approach $0$, showing consistency in this metric trend.

{The distinctiveness of a Lyapunov-based distance metric lies in the fact that its negative gradient direction, \( -\nabla V(s) \), indicates the direction of steepest descent for the state evolving toward the desired set of states. If \( \dot{V}(s) = (\nabla V(s))^\top f(s, \pi(s)) < 0 \) (where $\dot{s} = f(s, \pi(s))$ represents the system's evolution equation), it signifies that the system state is converging toward the desired state set; the monotonic decrease of \( V(s) \) corresponds to the contraction of a ``generalized distance'' between the system and the target. In contrast to static metrics relying solely on Euclidean distance, the Lyapunov function incorporates the underlying system dynamics, thereby providing a directionally guided stabilization policy path that more accurately reflects the actual physical processes.}

\subsubsection{Policy}

To construct a stabilization learning policy, one must first grasp the criteria for determining system stability. There are primarily two approaches to evaluating system stability, starting from the frequency domain and the time domain of the system, respectively.

\paragraph{(1) Frequency-Domain Stability Criteria}

Drawing on control theory, the closed-loop stability of single-input single-output linear systems (whose typical structure is illustrated in Figure \ref{fig:linear_feedback_system}) can generally be determined through the following two methods:

\begin{figure}[H]
  \centering
  \includegraphics[width=1\textwidth]{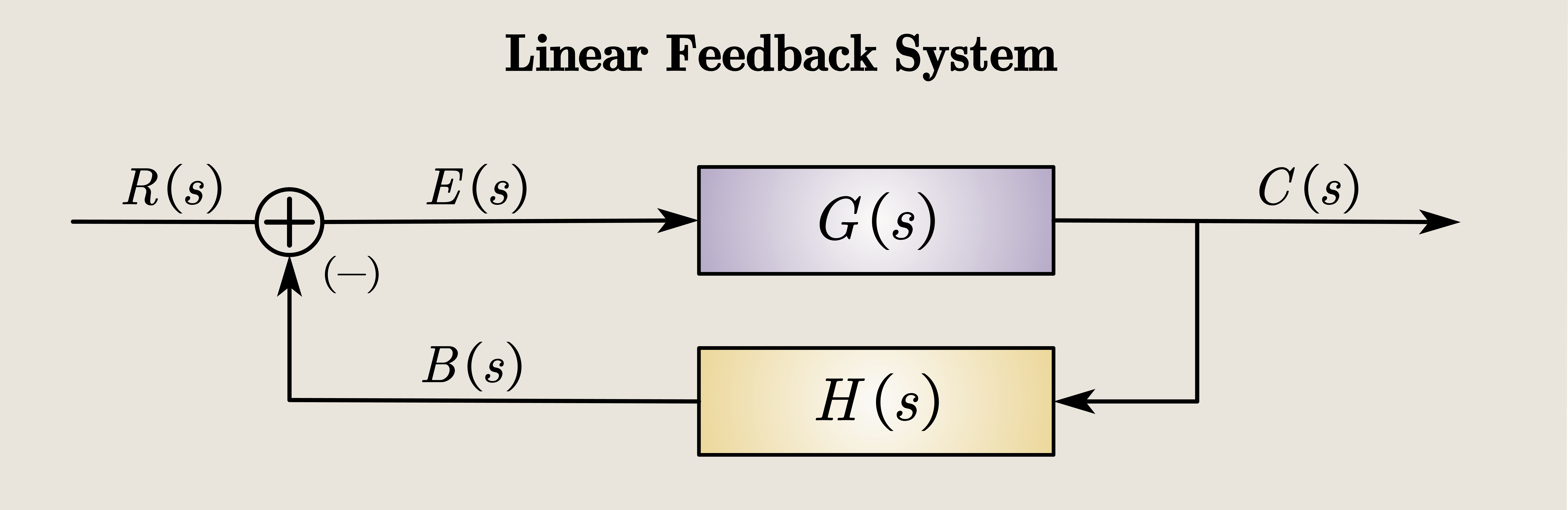}
  \caption{Linear feedback system structure}
  \label{fig:linear_feedback_system}
\end{figure}

\begin{figure}[H]
  \centering
  \begin{subfigure}[b]{0.48\textwidth}
    \centering
    \includegraphics[width=\textwidth]{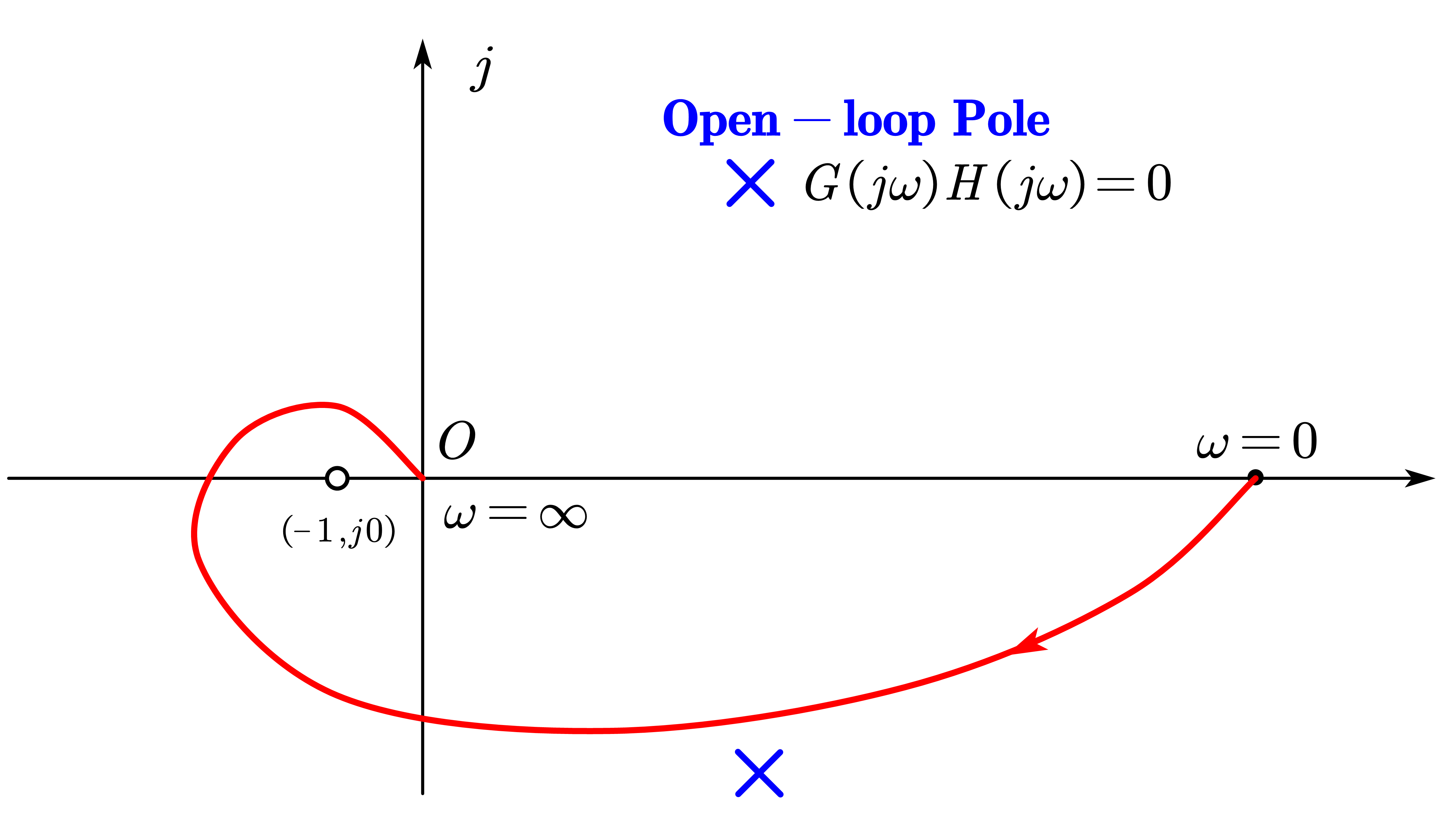}
    \caption{Nyquist Criterion}
    \label{fig:subimg1}
  \end{subfigure}
  \hfill
  \begin{subfigure}[b]{0.48\textwidth}
    \centering
    \includegraphics[width=\textwidth]{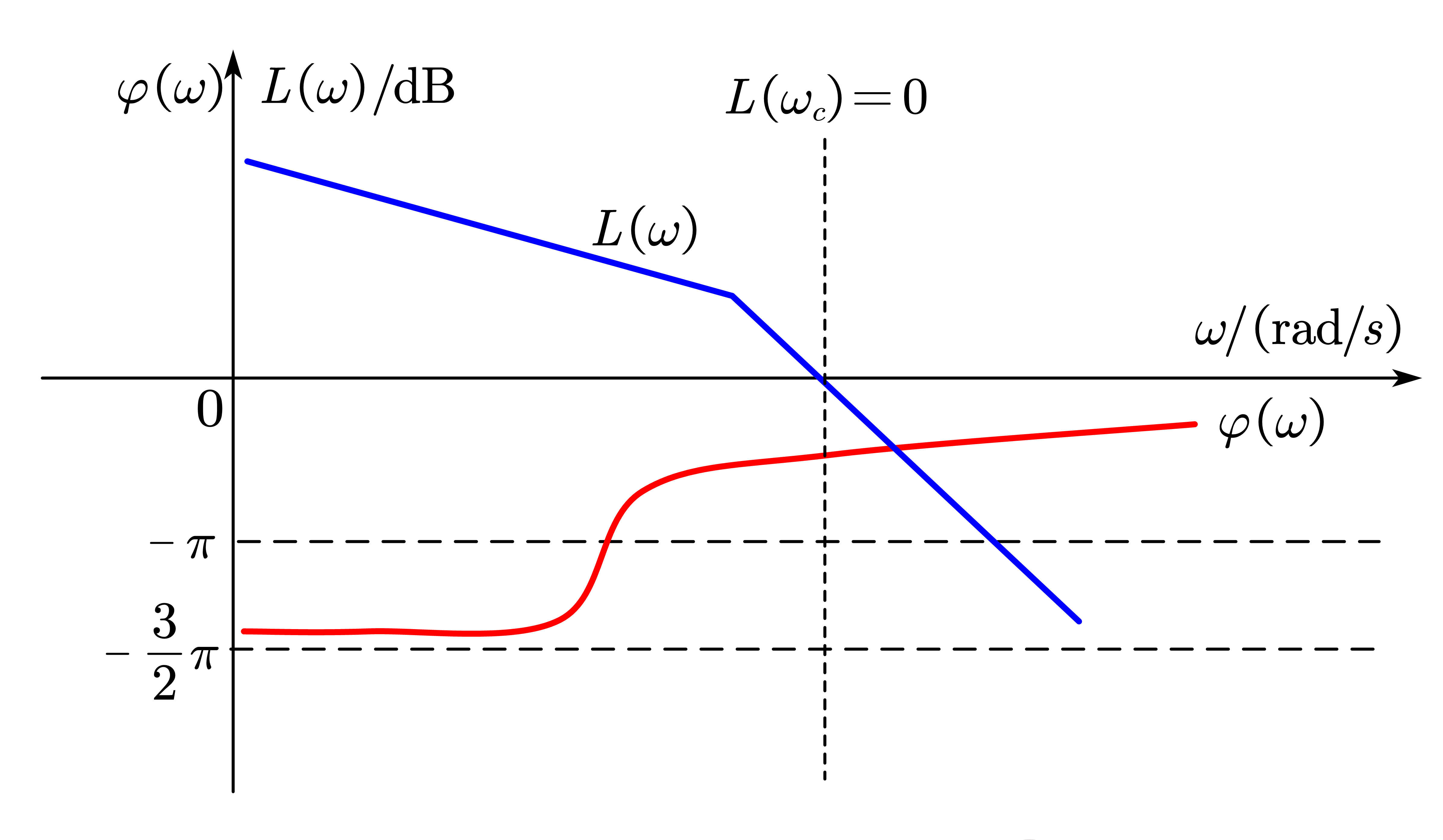}
    \caption{Bode Criterion}
    \label{fig:subimg2}
  \end{subfigure}
  \caption{Frequency-domain stability criteria for linear feedback systems}
  \label{fig:imgboth}
\end{figure}

\begin{itemize}[leftmargin=*, itemsep=2pt]
  \item \textbf{Nyquist Stability Criterion} \cite{kailath1980linear}: Based on the frequency characteristics of the open-loop transfer function $G(s)H(s)$ and the encirclement of the Nyquist curve on the complex plane relative to the critical point $(-1,j0)$, as shown in Figure \ref{fig:subimg1}, the stability of the closed-loop system is determined. The number of poles of the open-loop transfer function $G(s)H(s)$ in the right half of the complex plane is denoted as $P$, and the number of clockwise encirclements of the critical point $(-1,j0)$ by the Nyquist plot $G(j\omega)H(j\omega)$ is denoted as $N$. { Let $P$ be the number of open-loop right-half-plane poles and $Z$ be the number of closed-loop right-half-plane poles. With $N_{\mathrm{ccw}}$ denoting the net counterclockwise encirclements of $-1$ by the Nyquist plot, the Nyquist criterion gives
\[
Z = P + N_{\mathrm{ccw}}.
\]
Equivalently, if $N_{\mathrm{cw}}=-N_{\mathrm{ccw}}$ denotes the net clockwise encirclements, then
\[
Z = P- N_{\mathrm{cw}}.
\]
Closed-loop stability requires $Z=0$.}
  \item \textbf{Bode Plot Stability Criterion} \cite{kailath1980linear}: Based on the frequency characteristics of the open-loop transfer function $G(s)H(s)$, { plot the magnitude-frequency characteristic curve
$L(\omega) = 20\lg |G(j\omega)H(j\omega)|$
and the phase-frequency characteristic curve
$\varphi(\omega) = \angle G(j\omega)H(j\omega)$.
The Bode criterion can be viewed as a frequency-domain interpretation of the Nyquist criterion under appropriate assumptions. For minimum-phase open-loop systems, a positive phase margin at the gain crossover frequency is commonly used as an indicator of closed-loop stability}, as shown in Figure \ref{fig:subimg2}. In the frequency range where $L(\omega)\geq 0$, the phase-frequency characteristic curve crossing $-\pi$ from bottom to top is counted as one positive crossing; a curve starting from the $-\pi$ line and going upward is counted as a half positive crossing, denoted by $N_+$. In the frequency range where $L(\omega)\geq 0$, the phase-frequency characteristic curve crossing $-\pi$ from top to bottom is counted as one negative crossing; a curve starting from the $-\pi$ line and going downward is counted as a half negative crossing, denoted by $N_-$. The total number of crossings is defined as $N = N_+ - N_-$. Let $P$ denote the number of open-loop unstable poles; the necessary and sufficient condition for closed-loop system stability is $P = 2N$ (equivalent to the Nyquist criterion). Furthermore, for minimum phase systems, system stability can be judged by the relationship between the phase-frequency characteristic curve value at the cutoff frequency $\omega_c$ and $-\pi$: if $\varphi(\omega_c) > -\pi$, the closed-loop system is stable; otherwise, it is an unstable system.
\end{itemize}

\begin{figure}[H]
  \centering
  \includegraphics[width=1\textwidth]{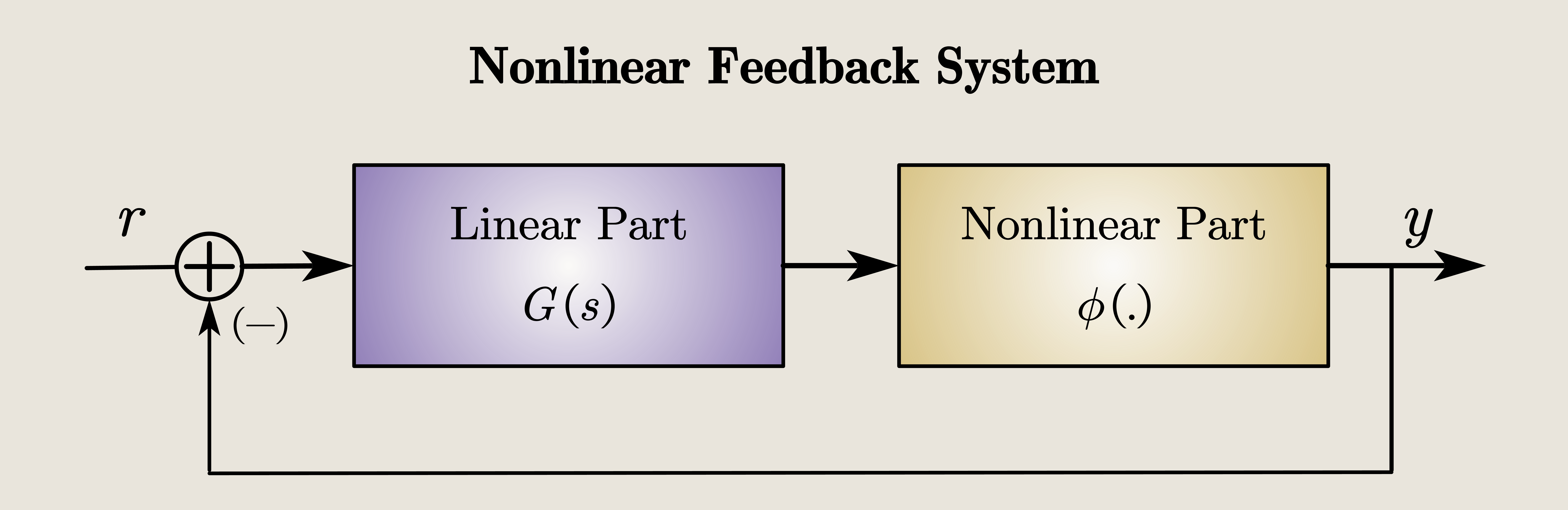}
  \caption{Nonlinear feedback system structure}
  \label{fig:popov_criterion}
\end{figure}

{For nonlinear systems (whose general structure is illustrated in Figure \ref{fig:popov_criterion}), \textbf{Popov's Criterion} \cite{khalil2002nonlinear} can be used to determine system stability in the frequency domain. The nonlinearity to which Popov's criterion applies often consists of two parts: the transfer function of the linear part $G(s)$ needs to ensure its own stability, while the nonlinear part $\phi(y,t)$ must satisfy the \textbf{Sector-Bound Condition}, i.e.:
\begin{equation}
\exists  k_1,k_2,\quad\text{s.t.}\quad k_1\le \frac{\phi(y,t)}{y}\le k_2 .
\end{equation}
In the vast majority of cases, the sector boundary condition is set as $k_1=0,k_2=K$, which is the \textbf{Regular Sector Condition}.}

Popov's theorem states that when the transfer function of the linear part satisfies the following requirements, the system is absolutely stable:
\begin{equation}
\forall \omega\ge 0,\exists q\ge 0, \quad \text{s.t.} \quad\text{Re}[(1+j\omega q)G(j\omega)]+\frac{1}{K} \ge 0.
\end{equation}

\paragraph{(2) Time-Domain Stability Criteria}

For general stability problems, which may involve various complex systems, their stability can be determined by analyzing their actual output responses using the mathematical tool of the \textbf{Lyapunov function} \cite{khalil2002nonlinear}. The Lyapunov function is a core tool for analyzing system stability in control theory and dynamical systems, evaluating the stability near the equilibrium point (usually the origin $s = 0$) by constructing a scalar function $V(s)$. Several essential stability classification definitions based on the Lyapunov function are introduced below.

\begin{itemize}[leftmargin=*, itemsep=2pt]
  \item \textbf{Lyapunov Stable}: For any $\epsilon > 0$, there exists a $\delta > 0$ such that if $\|s(0)\| < \delta$, then for all $t \geq 0$, $\|s(t)\| < \epsilon$. It requires $V(s)$ to be positive definite and $\dot{V}(s) \leq 0$ (negative semi-definite) near the equilibrium point. In this case, the system trajectory will not move far from the equilibrium point, but it does not necessarily converge.
  \item \textbf{Asymptotically Stable}: The system is Lyapunov stable, and there exists a $\delta > 0$ such that if $\|s(0)\| < \delta$, then $\lim_{t \to \infty} s(t) = 0$. It requires $V(s)$ to be positive definite and $\dot{V}(s)$ to be negative definite near the equilibrium point. In this case, the system trajectory eventually converges to the equilibrium point. If $V(s)$ is positive definite and radially unbounded (i.e., when $\|s\| \to \infty$, $V(s) \to \infty$), and $\dot{V}(s)$ is negative definite { throughout the state space}, the system converges to the equilibrium point under all initial conditions. At this point, the system is said to be \textbf{globally asymptotically stable}.
  \item \textbf{Exponentially Stable}: There exist constants $\alpha, \beta, \gamma > 0$ such that for all $t \geq 0$, $\|s(t)\| \leq \alpha \|s(0)\| e^{-\beta t}$. It requires $V(s)$ to satisfy $c_1 \|s\|^2 \leq V(s) \leq c_2 \|s\|^2$, and $\dot{V}(s) \leq -c_3 V(s)$, where $c_1, c_2, c_3 > 0$. In this case, the system converges to the equilibrium point at an exponential rate.
  \item {\textbf{Input-to-State Stable (ISS)}: ISS focuses on the response characteristics of the system under external input excitation; the analysis framework based on the Lyapunov function can be formulated as follows: consider a nonlinear system $\dot{s} = f(s, u), y = h(s)$ where $s \in \mathbb{R}^n$ is the state, $u \in \mathbb{R}^m$ is the input, and $y \in \mathbb{R}^p$ is the output. If there exists a continuously differentiable Lyapunov function $V(s)$ and class $\mathcal{K}$ functions\footnote{A continuous function \(\alpha: [0,a) \to [0,\infty)\) is said to be a class \(\mathcal{K}\) function if it is strictly monotonically increasing and satisfies \(\alpha(0) = 0\). Furthermore, if \(a = +\infty\) and \(\lim_{r\to +\infty}\alpha(r) = +\infty\), then \(\alpha\) is said to be a class \(\mathcal{K}_\infty\) function.} $\alpha_1, \alpha_2, \alpha_3, \sigma$ satisfying $\alpha_1(\|s\|) \leq V(s) \leq \alpha_2(\|s\|), \dot{V}(s) \leq -\alpha_3(\|s\|) + \sigma(\|u\|)$, then the system is input-to-state stable. This condition guarantees the boundedness of the system state and the finite gain characteristic from external input to internal state.}
  \item \textbf{$\mathcal{L}_2$ Stable}: $\mathcal{L}_2$ stability describes the response characteristics of a system under finite energy input, establishing a connection with Lyapunov theory through the dissipation inequality. A system is said to be finite-gain $\mathcal{L}_2$ stable if there exist constants $\gamma \geq 0$ and $\beta \geq 0$ such that $\|y\|_{\mathcal{L}_2} \leq \gamma \|u\|_{\mathcal{L}_2} + \beta$. The criterion based on the Lyapunov function requires the existence of a storage function $V(s) \geq 0$ satisfying the dissipation inequality: $\dot{V}(s) \leq \frac{1}{2}(\gamma^2 \|u\|^2 - \|y\|^2)$; integration yields $V(s(t)) - V(s(0)) \leq \frac{1}{2} \int_0^t (\gamma^2 \|u(\tau)\|^2 - \|y(\tau)\|^2) \text{d}\tau$. Rearranging this equation gives the $\mathcal{L}_2$ gain condition, reflecting the intrinsic connection between the Lyapunov function and the system's input-output performance.
\end{itemize}

Traditional Lyapunov stability theory requires $\dot{V}(s)$ to be negative definite to guarantee asymptotic stability, a condition that is often difficult to meet in actual systems. The \textbf{Invariance Principle} \cite{khalil2002nonlinear} significantly expands the application scope of the Lyapunov method by relaxing this requirement to merely $\dot{V}(s) \leq 0$, providing a unified framework for various stability analyses including asymptotic stability, limit cycles, and multi-equilibrium point systems.

The Invariance Principle states: suppose there exists a continuously differentiable function $V: \mathcal{S} \rightarrow \mathbb{R}$ satisfying:

\begin{itemize}[leftmargin=*, itemsep=2pt]
  \item \textbf{Boundedness Condition}: There exists a compact set $\Omega \subset \mathcal{S}$ such that $\Omega_c = \{ s \in \mathcal{S} \mid V(s) \leq c \}$ is a compact set and positively invariant.
  \item \textbf{Non-positivity Condition}: On $\Omega$, it holds that $\dot{V}(s) = (\nabla V(s))^\top f(s) \leq 0$.
\end{itemize}
Define the set:
\begin{equation}
E = \{ s \in \Omega \mid \dot{V}(s) = 0 \}.
\end{equation}
Let $M$ be the largest invariant set in $E$, then for all trajectories $s(t)$ originating from $\Omega$, as $t \rightarrow \infty$, it satisfies:
\begin{equation}
\lim_{t \to \infty} d(s(t), M) = 0.
\end{equation}

\begin{figure}[H]
  \centering
  \includegraphics[width=0.98\textwidth]{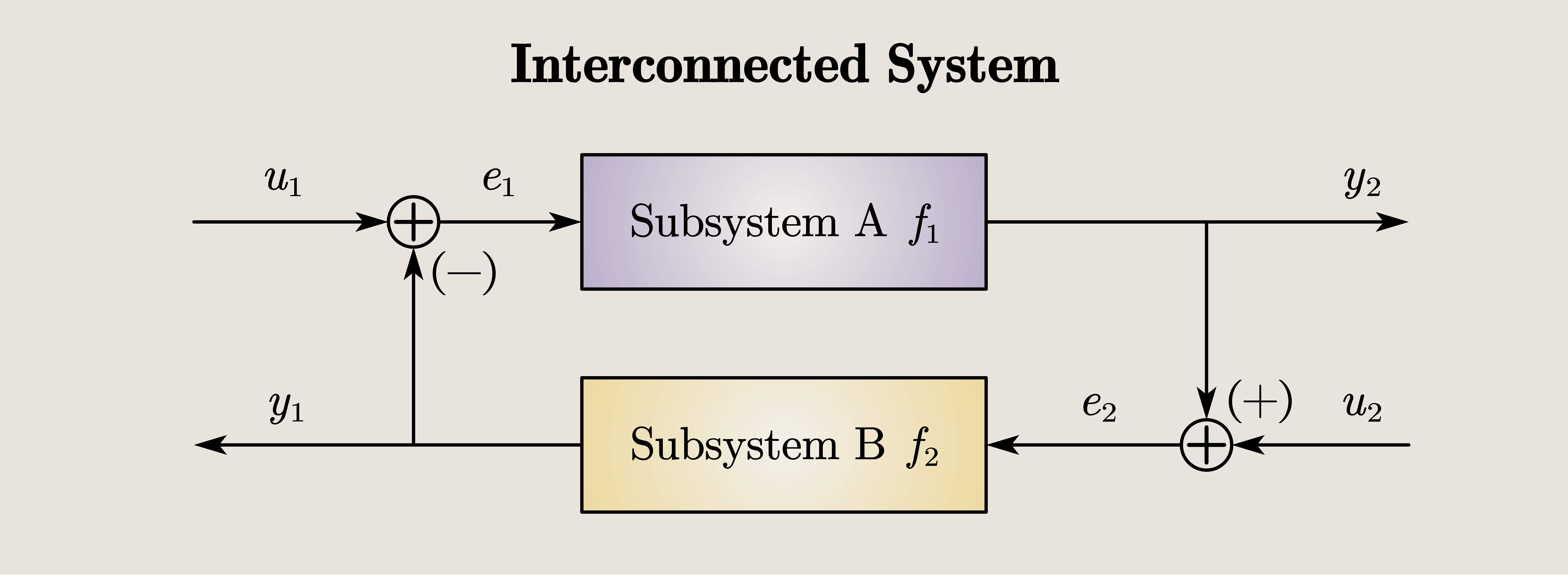}
  \caption{Structure of the interconnected system}
  \label{fig:united_system}
\end{figure}

Furthermore, for interconnected systems in engineering applications, the \textbf{Small-Gain Theorem} \cite{khalil2002nonlinear} is frequently employed to determine the overall stability of the closed-loop system based on the gain relationships among its subsystems. The small-gain theorem states that in an interconnected system (as illustrated in Figure \ref{fig:united_system}), if two stable subsystems are connected to form a feedback loop and the total loop gain is strictly less than one, then the entire closed-loop system is stable. This stability can be rigorously proven using Lyapunov functions.

Suppose that each subsystem within the interconnected system satisfies the input-to-state stability (ISS) condition. That is, there exist ISS-Lyapunov functions \( V_1(s_1) \) and \( V_2(s_2) \), obtained via smooth \(\mathcal{K}_\infty\) transformations, such that:
\begin{equation}
  \begin{aligned}
  \dot{V}_1(s_1) &\leq -V_1(s_1) + \gamma_1(V_2(s_2)), \\
  \dot{V}_2(s_2) &\leq -V_2(s_2) + \gamma_2(V_1(s_1)),
  \end{aligned}
\end{equation}
where \( \gamma_1 \) and \( \gamma_2 \) are class \(\mathcal{K}_\infty\) gain functions that satisfy the following small-gain condition:
\begin{equation}
  \gamma_1 \circ \gamma_2(s) < s, \quad \forall s > 0.
\end{equation}
Consequently, the overall closed-loop interconnected system composed of these two subsystems is input-to-state stable. For further proofs and comprehensive discussions regarding this theorem, readers are referred to \cite{khalil2002nonlinear, jiang1996lyapunov, dashkovskiy2010small}.

\paragraph{(3) Stabilization Learning Policy}

Based on system stability criteria, the core of stabilization learning problems lies in designing a feedback control law $a$ that ensures the system state converges to the desired state set $ \mathcal{S}_{\mathrm{d}}$. To achieve this goal, whether based on manually designed control laws or automated design methods leveraging optimization theory, dynamic programming, or reinforcement learning, the essence is to ensure the closed-loop system satisfies stability conditions at the equilibrium point. Specifically, it seeks an appropriate control law $a = \pi(s)$ such that the closed-loop system dynamics:
\begin{equation}
\dot{s} = f(s, \pi(s))
\end{equation}
is stable at the equilibrium point $s = 0$.

Lyapunov controller design based on traditional control theory is a classic and effective scheme in the fields of control and embodied AI. This method constructs a positive definite Lyapunov function $V(s) > 0$ ($\forall s \neq 0$) satisfying $V(0) = 0$, and designs a control input $u$ such that the Lyapunov derivative along the closed-loop system trajectory satisfies:
\begin{equation}
\dot{V}(s) = (\nabla V(s))^\top f(s, u) \leq -W(s) < 0 \quad (\forall s \neq 0),
\end{equation}
where $W(s)$ is a positive definite function. For example, for a typical system $\dot{s} = f(s) + g(s)u$, if $V(s) = \dfrac{1}{2} s^\top s$ is chosen, then:
\begin{equation}
\dot{V}(s) = s^\top (f(s) + g(s)u).
\end{equation}
Further designing the control law as $u = -k g(s)^\top s$ ($k > 0$) yields:
\begin{equation}
\dot{V}(s) = s^\top f(s) - k \|g(s)^\top s\|^2.
\end{equation}
If the gain $k$ can be appropriately selected to ensure $\dot{V}(s)$ is negative definite, it can be asserted according to the Lyapunov stability theorem that system stabilization is achieved.

\subsubsection{Stable System Construction Example: Attractor Networks}

Besides controller design based on Lyapunov functions, other classical methods exist for achieving system stability. For instance, discrete and continuous attractor networks guide the system toward stable behavior by constructing topological structures with ``attractor" properties.

\paragraph{(1) Discrete Attractor Networks}

A representative example of discrete attractor networks is the \textbf{Hopfield Network} \cite{hopfield1982emergent}. Under structural conditions such as symmetric weights, no self-feedback, and asynchronous updates, a Hopfield network admits an explicit energy function whose value decreases monotonically along the state-update process.

For example, for an $N$-neuron network, {the neuron state $s_i\in\{+1,-1\}$}, the connection weight between neurons is $W_{ij}$ (satisfying symmetry $W_{ii}=0,W_{ij}=W_{ji}$), and the neuron threshold is $\theta_i$. Its energy function can be constructed as:
\begin{equation}
E=-\frac{1}{2}\sum_{i=1}^{N}\sum_{j\neq i}^{N}\omega_{ij}s_i s_j+\sum_{i=1}^{N}\theta_i s_i.
\end{equation}
To ensure the energy function $E$ monotonically decreases (and has a lower bound) as the network dynamically evolves, the update of every neuron in the network must follow specific rules. For the local field of neuron $s_{i}$:
\begin{equation}
h_i = \sum_{j = 1}^N\omega_{ij}s_j - \theta_i.
\end{equation}
The energy change is:
\begin{equation} \label{eq:energy_change}
\Delta E_i = -\frac{1}{2}(s_i^{\text{new}} - s_i^{\text{old}})h_i.
\end{equation}
Since the network energy must maintain a decreasing trend during the updating process ($\Delta E_i \leq 0$), the right-hand side of Equation~\eqref{eq:energy_change} must always remain non-positive. Therefore, the updated neuron state $s_i^{\text{new}}$ must consistently share the same sign as $h_i$, and its assignment rule is formulated as follows:
\begin{equation}
s_i^{\text{new}}=\text{sign}\left(\sum_{j\neq i}^{N}\omega_{ij}s_j\left(t\right)-\theta_i\right) = 
\begin{cases}
+1 & \text{if } h_i > 0 \\
-1 & \text{if } h_i < 0 \\
s_i^{\text{old}} & \text{if } h_i = 0
\end{cases}.
\end{equation}
This equation fully complies with the requirements of Lyapunov stability. Therefore, the network will inevitably converge to a stable state. These stable states (local energy minima) are the preset ``equilibrium points" of the network, representing its stored memories, thus achieving the integration of stabilization learning and optimization learning.

The key to creating system stabilization lies in constructing such stable networks, which involves learning the weight parameters $[w_{ij}]$ in the network. To guide the system state to evolve towards the predicted set, it can follow the Hebbian learning rule, making specific patterns $x_i^\mu (\mu = 1, \dots, P, x_i^\mu = \pm 1)$ network attractors. In this case, the connection weights can be set as:
\begin{equation}
w_{ij}=\frac{1}{N}\sum_{\mu=1}^{P}x_i^{\left(\mu\right)}x_j^{\left(\mu\right)}\left(i\neq j\right).
\end{equation}
This rule indicates that if two neurons tend to activate or inhibit simultaneously across multiple patterns, the connection between them will be strengthened. This mechanism is equivalent to ``excavating" energy wells centered on these patterns within the system's energy landscape, thereby effectively guiding the neural network's dynamic behavior and ensuring its convergence to the desired stable state.

\begin{examplebox}{Application Example of Discrete Attractor Networks: Image Association and Recognition}{image-association}
Discrete attractor networks can be used to achieve associative memory and pattern recovery. For example, if the network is trained with clean samples to recognize standard letter images $A,B,C,\cdots$, the corresponding attractors are stored as $P$ patterns $x_i^\mu (\mu = 1, \dots, P)$. Then, if an incomplete or noise-corrupted letter image is provided as input, the network will be able to stably converge to the stored pattern $x_i^\nu$ (if $x_i^\nu$ has a unique match with $x_i^\mu$). This process realizes an image recognition system that stably regresses to standard results \cite{olson1993handwritten}. This process can be viewed as a pattern recognition and memory association mechanism based on the Hebbian learning rule.
\end{examplebox}

\paragraph{(2) Continuous Attractor Neural Networks}

A \textbf{Continuous Attractor Neural Network (CANN)} \cite{amari1977dynamics} is a neural network capable of forming and stably controlling a low-dimensional continuous activity pattern within a high-dimensional state space through a specific local-excitation/global-inhibition connection pattern, creating a stable manifold in the high-dimensional space (corresponding to the stable points of discrete attractor networks). It can remain stationary at any point on a smooth circular track, while external inputs can push it to move smoothly along the track. This model profoundly simulates the potential mechanisms by which biological neural systems process continuous information such as spatial direction and position.

For continuous attractor networks, an intrinsic energy function (Lyapunov function) also exists (taking 1D as an example \cite{stringer2002self}):
\begin{equation}
E(u) = - \frac{1}{2}\iint w(x-x^\prime)\phi(u(x))\text{d}x\text{d}x^\prime - \int I(x)\phi(u(x))\text{d}x+ \iint_0^{u(x)}\phi^{-1}(z)\text{d}z\text{d}x.
\end{equation}
Here, $\tau$ is the neuron time constant, which determines the network's response speed; $u(x,t)$ is the state at position $x$ and time $t$; $\phi(\cdot)$ is the nonlinear activation function; $I(x,t)$ is the external input; $w(r) = Ae^{-r^2/\sigma_e^2}-Be^{-r^2/\sigma_i}$ is the connection weight kernel (where $A$ and $B$ are the corresponding amplitude constants). This is a necessary condition for bump-like attractors and determines the topological structure of the constructed attractor. Two types of connection weights for attractors are given below:
\begin{itemize}
\item Ring Attractor: $w(x) = A\cos(kx)-B,$
\item 2D Grid Attractor: $w(x,y) = \sum A_n e^{-\frac{(x\cos\theta_n+y\sin\theta_n)^2}{2\sigma^2}}-B.$
\end{itemize}

The energy function $E(u)$ satisfies the condition $\frac{\text{d}E}{\text{dt}} = -\frac{1}{\tau}\int(\frac{\partial u}{\partial t})^2\text{d}x \le 0$, ensuring the system evolves toward the energy minimum. The continuous attractor corresponds to a flat valley of the energy function (energy remains constant along the valley direction and decreases in the perpendicular direction).

Taking the partial derivative of the continuous attractor network's intrinsic energy with respect to the network parameters $w$ yields its dissipative neural field model in the form of integro-differential equations, as shown below:
\begin{equation}
\begin{aligned}
\tau \frac{\text{d}r_i}{\text{d}t} &= -\frac{\partial E}{\partial r_i} \\
\tau\frac{\partial u(x,t)}{\partial t} &= -u(x,t)+\int_{-\infty}^{\infty}w(x-x^\prime)\phi(u^\prime(x^\prime,t))\text{d}x^\prime+I(x,t).
\end{aligned}
\end{equation}
When the network is stable, the equation becomes a pure integral equation:
\begin{equation}
u(x,t) = \int_{-\infty}^{\infty}w(x-x^\prime)\phi(u(x^\prime,t))\text{d}x^\prime+I(x,t).
\end{equation}

\begin{figure}[H]
  \centering
  \includegraphics[width=1\textwidth]{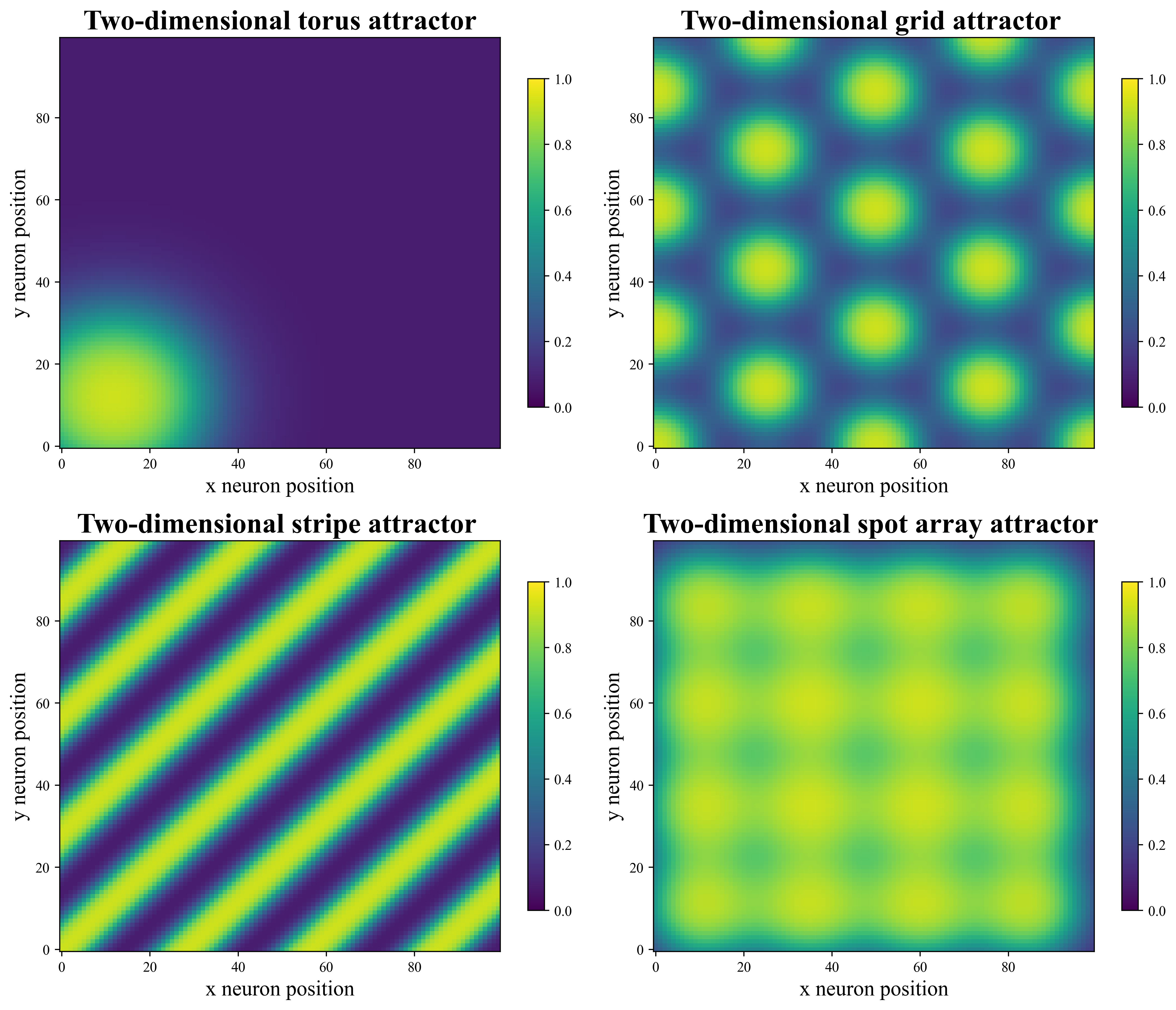}
  \caption{Visualization of neuron activities in various 2D attractors}
  \label{fig:attractors_visualization}
\end{figure}

Neuron Activities in Various 2D Attractors are shown in Figure \ref{fig:attractors_visualization}. To construct the stable manifold of the continuous attractor network, the network weights must be learned following the Hebbian rule to generate translation-invariant weights (where $\Delta w_{ij}$ is the update amount of the weights during learning), namely:
\begin{equation}
\Delta w_{ij} = \eta\left(\phi(u_i)\phi(u_j)-\frac{1}{N}\sum_{k,l}\phi(u_k)\phi(u_l)\right) - \gamma w_{ij}.
\end{equation}

\begin{table}[H]
  \centering
  \footnotesize
  \renewcommand{\arraystretch}{1.3} 
  \begin{tabular}{lccc}
    \toprule
    \textbf{Attractor Type} & \textbf{State Space Topology} & \textbf{Connection Weight} & \textbf{Biological Analogy} \\
    \midrule
    \textbf{2D Torus} & $x\in [0,L),y\in[0,L)$& $ J_0+J_1\sum_{m,n \in \mathbb{Z}}\text{exp}^{-(r_i-r_j)^2/\sigma_e^2} $ & Grid Cells (mEC) \\
    \cmidrule{1-4}
    \textbf{2D Grid} & $\mathbb{R}^2$ Rotation + Mirror & $ J_0+J_1\sum_{p \in G_{\textbf{hex}}}\text{exp}^{-\Delta r_p^2/\sigma_e^2} $ & Visual Cortex Orientation \\
    \cmidrule{1-4}
    \textbf{Spot Array} & $\mathbb{R}^2$ Rotation & $\begin{aligned} &J_0+J_1\sum_{i = 0,1,\cdots,5}\cos{\text{R}_{i\pi/6} q_0\cdot \Delta r}, \\ &q_0 = (4\pi/(\sqrt{3}a),0) \end{aligned}$ & Ocular Dominance Columns \\
    \cmidrule{1-4}
    \textbf{2D Stripes} & $\mathbb{R}^2$ Unidirectional Periodic & $ J_0+J_1\cos(q\cdot \Delta r),q = 2\pi/\lambda $ & Zebrafish Stripes \\
    \bottomrule
  \end{tabular}
  \caption{Characteristics of different types of attractors. The weight $w(\Delta r)$ is a function of relative coordinates, represented in the table as $r_i-r_j,\Delta r_p$, or $\Delta r$. In the specific function expressions, $J_0$ and $J_1$ are constant parameters, and $\text{R}_\alpha$ is the rotation matrix for an angle $\alpha$. \(G_{\textbf{hex}}\) denotes the set of symmetry transformations for the hexagonal lattice.}
  \label{tab:attractor_types}
\end{table}

\begin{examplebox}{Application Example of Continuous Attractor Networks: Brain-Inspired Navigation}{brain-navigation}
Continuous attractor networks are closely related to biological neural systems and can be used to simulate the process of brain navigation. The brain-inspired navigation system \cite{samsonovich1997path, couey2021continuous} translates multimodal sensory information into spatially invariant topological representations by simulating the encoding of linear and angular velocities by speed cells, directional encoding by head direction cells, spatial position encoding by place cells, and path integration encoding by grid cells. Relying on episodic memory simulation and reward prediction error mechanisms, it achieves goal-directed vector navigation. This system possesses capabilities for autonomous environmental perception, path planning, and online error correction, enabling efficient navigation in complex environments.
\end{examplebox}

Under conditions satisfying specific structural constraints, certain neural network dynamical systems can be modeled as dynamical systems with attractor structures. For example, in Hopfield networks with symmetric weights, no self-feedback, and asynchronous update rules, explicit energy functions can be constructed as Lyapunov functions. For general deep neural networks, the aforementioned energy functions and convergence properties do not automatically hold and require additional structural or training constraints.

\subsection{Relationship between Stabilization Learning and Reinforcement Learning}

\subsubsection{Mathematical Formulation of the Two Learning Paradigms}

    \begin{figure}[htbp]
        \centering
        \resizebox{\textwidth}{!}{
        \begin{tikzpicture} [
            node distance=1.5cm and 1cm,
            box/.style={
                draw=black, 
                thick, 
                fill=white, 
                minimum width=4cm, 
                minimum height=1.2cm, 
                align=center, 
                font=\bfseries 
            },
            dashedbox/.style={
                draw=black!80, 
                dashed, 
                thick, 
                fill=white, 
                align=left, 
                inner sep=10pt, 
                font=\small 
            },
            arrowstyle/.style={
                -{Triangle[width=6mm,length=4mm]}, 
                line width=3pt, 
                color=myyellow
            },
            bullet/.style={
                font=\large, 
                baseline
            }
        ]
        
        
            \node[inner sep=0pt] (bellman_img) at (0, 0) {
                \includegraphics[width=3cm, height=3cm, keepaspectratio]{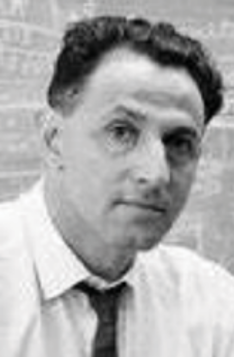}
            };
            \node[below=0.2cm of bellman_img] {Richard E. Bellman}; 
        
            \node[box, above=1cm of bellman_img] (bellman_box) {Bellman Optimality\\Equation};
            
            \draw[arrowstyle] (bellman_img) -- (bellman_box);
        
            \node[box, above=3.5cm of bellman_box] (rl_box) {Reinforcement\\Learning};
            
            \draw[arrowstyle] (bellman_box) -- (rl_box);
            
            \node[above=0.3cm of rl_box, font=\Huge, align=center] {$\vdots$};
        
            \node[dashedbox, right=2.2cm of bellman_box, anchor=west] (desc_bellman) {
                $\bullet$ Principle of Optimality \\
                $\bullet$ Solved via Recursive Iteration \\
                $\bullet$ Policy Evaluation \& Improvement \\
                $\bullet$ Existence \& Uniqueness of Solution
            };

            \node[dashedbox, right=2.2cm of rl_box, anchor=west] (desc_rl) {
                $\bullet$ Equality Constraint \\
                $\bullet$ Learn Value Function $V(s)$ \\
                $\bullet$ Driven by Maximizing \\ \quad Cumulative Reward \\
                $\bullet$ Value Function Guides \\ \quad System Behavior \\
                $\bullet$ Sequential Nature
            };
        
            \draw[arrowstyle, ->] (bellman_box.east) to[out=0, in=180] (desc_bellman.west);
            \draw[arrowstyle, ->] (rl_box.east) to[out=0, in=180] (desc_rl.west);

            
            \def\xoffset{15cm}
        
            \node[inner sep=0pt] (lyapunov_img) at (\xoffset, 0) {
                \includegraphics[width=3cm, height=3cm, keepaspectratio]{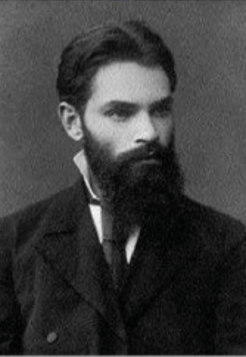}
            };
            \node[below=0.2cm of lyapunov_img] {Aleksandr Lyapunov}; 
        
            \node[box, above=1cm of lyapunov_img] (lya_cond_box) {Lyapunov Stability\\Condition};
            
            \draw[arrowstyle] (lyapunov_img) -- (lya_cond_box);
        
            \node[box, above=3.5cm of lya_cond_box] (lya_learn_box) {Lyapunov Function\\Learning};
            
            \draw[arrowstyle] (lya_cond_box) -- (lya_learn_box);
            
            \node[above=0.3cm of lya_learn_box, font=\Huge, align=center] {$\vdots$};
        
            \node[dashedbox, right=2.2cm of lya_cond_box, anchor=west] (desc_lya_cond) {
                $\bullet$ Stability Theory \\
                $\bullet$ Design via Lyapunov Function \\
                $\bullet$ Controller Redesign \\
                $\bullet$ Diversity of Solutions
            };
        
            \node[dashedbox, right=2.2cm of lya_learn_box, anchor=west] (desc_lya_learn) {
                $\bullet$ Inequality Constraint \\
                $\bullet$ Learn Lyapunov Function $V(x)$ \\
                $\bullet$ Driven by Satisfying Stability \\
                $\bullet$ Lyapunov Function Guides \\ \quad System Behavior \\
                $\bullet$ Instantaneous Nature
            };
        
            \draw[arrowstyle, ->] (lya_cond_box.east) to[out=0, in=180] (desc_lya_cond.west);
            \draw[arrowstyle, ->] (lya_learn_box.east) to[out=0, in=180] (desc_lya_learn.west);
        
        \end{tikzpicture}
        }
    \caption{Theoretical origins and characteristic comparison of reinforcement learning and stabilization learning. The left side shows the evolution path of reinforcement learning based on Bellman's optimality principle, emphasizing equality constraints and cumulative rewards; the right side shows the evolution path of stabilization learning based on Lyapunov stability theory, emphasizing inequality constraints and instantaneous stability.}
    \label{fig:comparison_diagram}
\end{figure}
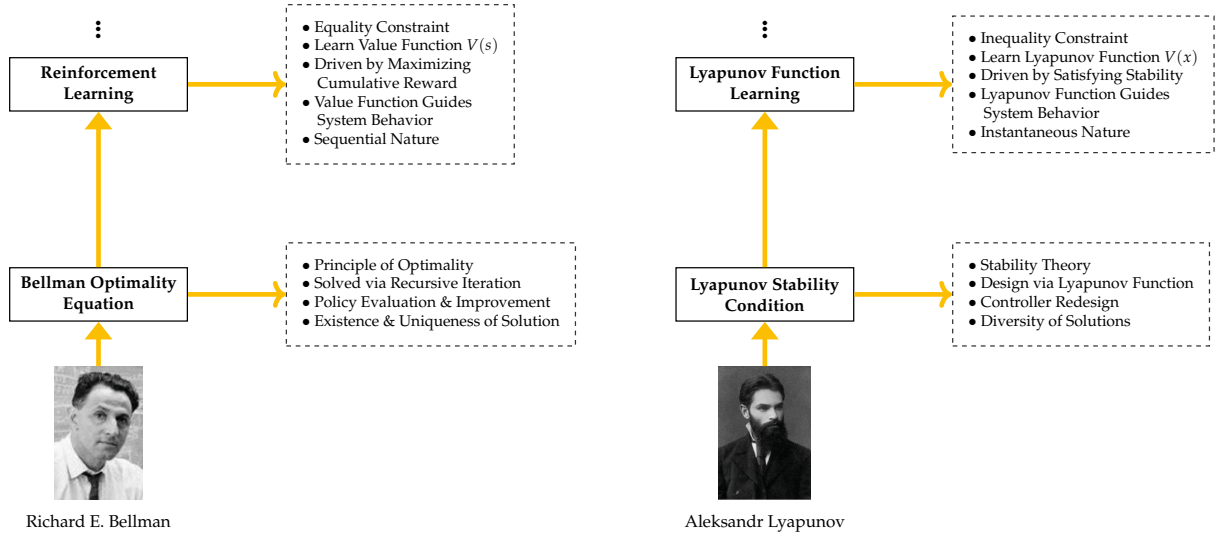

To deeply explore the intrinsic theoretical connection between stability and optimality, this section will formulate both paradigms mathematically under this unified framework. Figure \ref{fig:comparison_diagram} illustrates the theoretical origins and characteristic comparison of these two paradigms. It should be  noted that the scope of discussion in this article is limited to \textbf{Deterministic Continuous-time Systems} \cite{vrabie2013optimal}. Under this setting, the system is free from stochastic perturbations or observation noise, the state transition probability in reinforcement learning degenerates into a deterministic ODE, and the mathematical expectation operator $\mathbb{E}[\cdot]$ in the objective function transforms into a deterministic integral form. This assumption allows us to discuss stabilization learning and reinforcement learning within the same mathematical dimension.

Furthermore, it is necessary to clarify that generalized stabilization learning covers Lyapunov methods along with other technical stability criteria, and also includes implementation techniques such as deep feature extraction, dynamical system establishment, stable system construction, stabilization learning problem transformation, and generalization. However, when exploring its intrinsic connection with reinforcement learning (especially value-function-based methods), \textbf{Lyapunov Function Learning} \cite{dawson2023safe, chang2019neural, chang2021stabilizing, dawson2021safe} holds a corresponding relationship with the value function in reinforcement learning because it utilizes the Lyapunov function as a stability certificate. Therefore, unless otherwise specified, stabilization learning mentioned in this section and subsequent discussions specifically refers to certificate learning methods based on Lyapunov theory.

\subsubsection{Stability-Based Stabilization Learning}
    
Stability-based stabilization learning aims to find a control policy $a=\pi(s)$ and a corresponding scalar function $ V(s)$ such that the state of the closed-loop system asymptotically converges to the equilibrium point. Its core constraint is the energy decay condition \cite{khalil2002nonlinear}:
\begin{equation} \label{eq:clf_inequality} 
    \dot{V}(s) = \left(\nabla V \right)^\top f(s, \pi(s)) \leq -W(s),
\end{equation}
where $ V(s)$ is a positive definite function. If $W(s)$ is merely positive definite, then $\dot V\le 0$ only implies Lyapunov stability in general. To establish asymptotic stability, one needs additional arguments such as LaSalle's invariance principle. If $W(s)$ is positive definite, asymptotic stability follows
directly from Lyapunov's theorem.
 
Stabilization learning is essentially a feasible solution search problem based on \textbf{inequality constraints}. However, the selection of the auxiliary function $W(s)$ is artificial. If the decay rate set by $W(s)$ exceeds the physical limitations of the system (for instance, requiring exponential convergence when the system can only achieve polynomial convergence), it may lead to an empty feasible solution space.
	
\subsubsection{Optimality-Based Reinforcement Learning}
    
Optimality-based reinforcement learning \cite{sutton2018reinforcement, bertsekas2019reinforcement} is dedicated to solving for the optimal control policy $a = \pi^*(s)$ to minimize the cumulative discounted cost within an infinite horizon. Let $r(s,a): \mathbb{R}^n \times \mathbb{R}^m \to \mathbb{R}_{\geq 0}$ be defined as the running cost function of the task, used to quantify the instantaneous loss of the system. For any admissible policy $\pi$, its value function is defined as:
{
\begin{equation} \label{eq:value_function}
	J(\pi, s(t)) = \int_{t}^{\infty} e^{-\beta(\tau-t)} r(s(\tau), \pi(s(\tau))) \,\text{d}\tau,
\end{equation}
where $\beta > 0$ is the discount rate, and $\gamma = e^{-\beta} \in (0,1]$ is the discount factor. When $\gamma = 1$, it corresponds to the undiscounted infinite-horizon problem; when $\gamma < 1$, future costs are exponentially discounted. The optimal value function $V^*(s)$ is defined as the infimum over all admissible policies:
\begin{equation}
	V^*(s) = \inf_{\pi \in \Pi} J(\pi, s(t)).
\end{equation}}
Therefore, reinforcement learning is an extremum search problem based on an \textbf{optimality objective}.
	
\subsubsection{Complementary Characteristics Analysis}
	
Although both paradigms are dedicated to system optimization, there are significant differences in their constraint mechanisms and task applicability:
	
\begin{itemize}[leftmargin=*, itemsep=2pt]
	\item \textbf{Constraint Mechanism}: Stabilization learning provides a theoretical certificate of system stability through strict inequality constraints \eqref{eq:clf_inequality}, offering strong robustness guarantees. Reinforcement learning, on the other hand, typically converts constraints into penalty terms (soft constraints) within the cost function $r(s, a)$. Limited by function approximation errors and exploration mechanisms, it is often difficult to provide strict proofs of convergence.
	\item \textbf{Task Flexibility}: Stabilization learning usually requires an explicit desired state set, error system, or stability certificate, and formulates the task around convergence or boundedness. Reinforcement learning, in contrast, often specifies objectives through reward or cost functions and does not necessarily require an explicit desired state set. This difference gives RL greater flexibility in task formulation, while closed-loop stability typically requires additional structural assumptions, constraints, or post-hoc verification.
\end{itemize}
    
\subsubsection{Mutual Connection between Stabilization Learning and Reinforcement Learning} \label{sec:stability_optimality}
	
\paragraph{(1) Deriving Stabilization Learning from RL: Value Function as a Stability Certificate}

Assume the running cost function satisfies the following separable positive definite structure:
\begin{equation}
    r(s,a) = Q(s) + R(a),
\end{equation}
where $Q(s)$ and $R(a)$ are positive definite functions with respect to the state and control, respectively, requiring both to be zero at the origin ($Q(s) = 0 \text{ when } s = 0; R(a) = 0 \text{ when } a = 0$). The optimal value function $V^*(s)$ is absorbing at the origin and naturally satisfies positive definiteness ($V^*(s) > 0, \forall s \neq 0$ and $V^*(0)=0$), qualifying it as a Lyapunov candidate function. Without the prerequisite that $Q(s)$ and $R(a)$ evaluate to zero at the origin, positive definiteness alone cannot guarantee $V^\ast(0) = 0$, and thus it cannot be asserted to be a Lyapunov function.

For infinite-horizon\footnote{The infinite horizon indicates that the optimization objective covers the long-term behavior of the system's entire future trajectory starting from the initial moment, rather than the transient performance within a finite time period.} optimal control problems, the optimal value function $V^*(s)$ satisfies the Hamilton-Jacobi-Bellman (HJB) equation. 
When $\gamma = 1$ (undiscounted), the HJB equation is:
\begin{equation} \label{eq:HJB}
	\min_{a} \left\{ r(s, a) + (\nabla V^*)^\top f(s, a) \right\} = 0.
\end{equation}
Let $\pi^*(s)$ be the optimal policy; substituting it into the above equation yields the derivative of the closed-loop system along the optimal trajectory:
\begin{equation}
\begin{aligned}
	\dot{V}^*(s) &= (\nabla V^*)^\top f(s, \pi^*(s)) \\
    &= -r(s, \pi^*(s)) \\
    &= -Q(s) - R(\pi^*(s)).
\end{aligned}
\end{equation}
Since $R(a) \geq 0$, we have:
\begin{equation}
	\dot{V}^*(s) \leq -Q(s).
\end{equation}
This equation is completely consistent with the inequality constraint \eqref{eq:clf_inequality} of stabilization learning, where $W(s)$ corresponds to the state cost $Q(s)$. { Under these specific premises, namely an absorbing origin, no discounting,
and a positive definite cost, the optimal value function can serve as a Lyapunov candidate, thereby establishing a direct connection between optimality and stability.} Otherwise, lacking the prerequisite of the origin as an absorbing state, asymptotic stability of the system cannot be directly deduced from the positive definiteness of the cost.
	
\paragraph{(2) Deriving Reinforcement Learning from Stabilization Learning: Constraint Relaxation and Feasible Region Collapse}
	
Consider the Lyapunov-based stabilization learning problem, whose standard form seeks a policy $\pi(s)$ such that $\dot{V}(s) \leq -Q(s)$. To establish a connection with reinforcement learning, we introduce a slack term $\delta(s, a)$ containing both state and control, expanding the stability constraint into a stricter form:
\begin{equation} \label{eq:augmented_constraint}
	\dot{V}(s) \leq -Q(s) - \delta(s, a).
\end{equation}
Rearranging the terms, the above constraint is equivalent to satisfying non-positivity:
\begin{equation} \label{eq:hamiltonian_ineq}
	Q(s) + \delta(s, a) + (\nabla V)^\top f(s, a) \leq 0.
\end{equation}
	
Now suppose the chosen Lyapunov candidate function $V(s)$ happens to be the optimal value function $V^*(s)$ in reinforcement learning, and the slack term is set as the control cost $\delta(s, a) = R(a)$. At this point, the aforementioned problem of finding a feasible solution for stabilization learning transforms into finding $a$ such that:
\begin{equation} \label{eq:rl_constraint}
	Q(s) + R(a) + (\nabla V^*)^\top f(s, a) \leq 0.
\end{equation}
	
On the other hand, according to optimal control theory, the optimal value function $V^*(s)$ must satisfy the HJB equation. The HJB equation states that the optimal control minimizes the Hamiltonian function to $0$:
\begin{equation} \label{eq:HJB_min}
	\min_{a} \left\{ Q(s) + R(a) + (\nabla V^*)^\top f(s, a) \right\} = 0.
\end{equation}
This equation means that for any admissible control $a$, the actual value of the Hamiltonian function must be greater than or equal to $0$:
\begin{equation} \label{eq:hjb_lower_bound}
	Q(s) + R(a) + (\nabla V^*)^\top f(s, a) \geq 0.
\end{equation}
	
Comparing the feasibility constraint of stabilization learning \eqref{eq:rl_constraint} with the theoretical lower bound of the HJB equation \eqref{eq:hjb_lower_bound}, it is evident that the only possibility of satisfying both conditions simultaneously is \textbf{the inequality taking the equal sign}. In other words, by introducing a slack term $\delta(s,a)=R(a)$ consistent with the task cost, the originally broad feasible solution space for stability ``collapses" into deterministic points (or sets), and this solution is precisely the optimal policy $\pi^*(s)$ satisfying the HJB equation. This result indicates that: \textbf{under specific relaxed constraints, the feasible solution of stabilization learning is exactly the optimal solution of reinforcement learning.}

\paragraph{(3) Impact of Discount Factor}
When the discount factor satisfies $\gamma < 1$ in the reinforcement learning value function \eqref{eq:value_function}, the value function in the HJB equation no longer directly corresponds to the Lyapunov function under the undiscounted case. Consequently, global asymptotic stability of the system is no longer guaranteed; {however, additional conditions beyond the discounted HJB equation are required to establish boundedness or local stability; the discount factor alone does not guarantee either.}  In this scenario, by tuning $\gamma$ and the weights of the cost function, reinforcement learning effectively adjusts the decay characteristics of the Lyapunov function and the geometry of the domain of attraction. The objective generally places a greater emphasis on the discounted cumulative performance, and stability conclusions require the support of additional conditions.
	
\subsubsection{Paradigm Comparison in Discrete Systems: D-learning { vs.} Q-learning}
	
\begin{figure}[htbp]
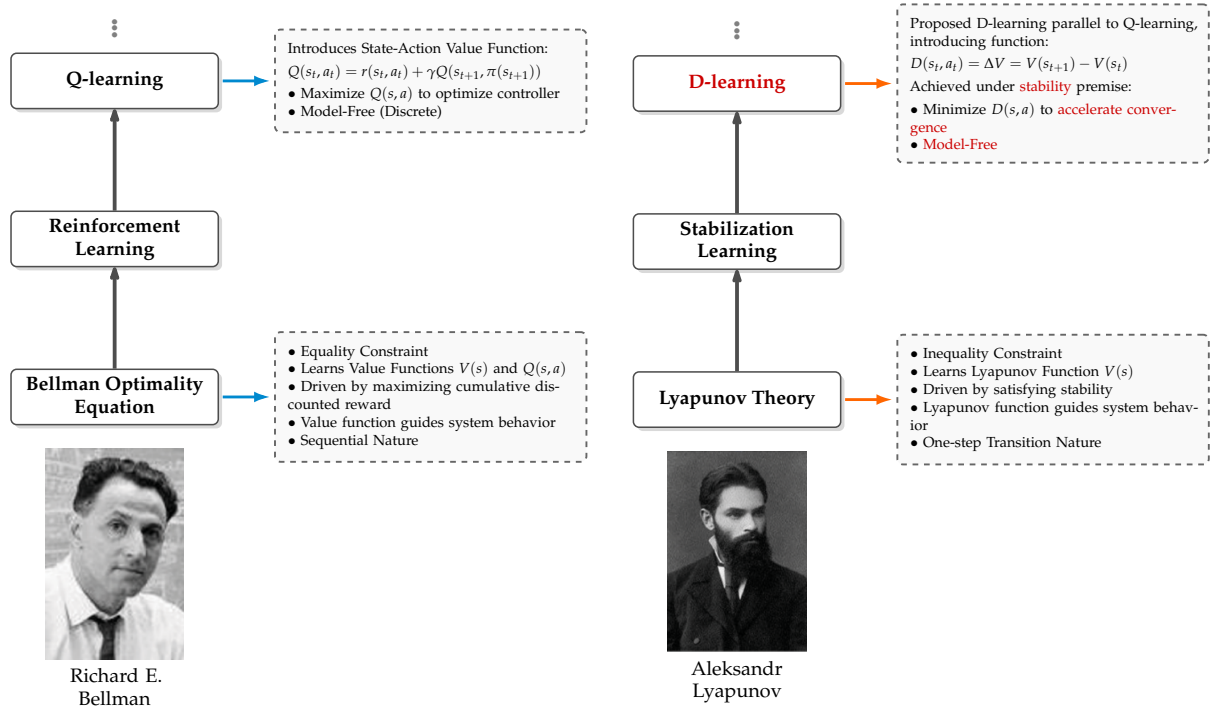

	\centering
	\resizebox{1.0\textwidth}{!}{
		\begin{tikzpicture}[
			scale=0.85, 
			transform shape,
			node distance=2.2cm and 1.2cm, 
			mainbox/.style={
				draw=black!70, 
				thick,
				rectangle,
				rounded corners=3pt, 
				minimum width=4.5cm,
				minimum height=1.2cm,
				align=center,
				font=\bfseries\large,
				fill=white,
				drop shadow={opacity=0.3, shadow xshift=2pt, shadow yshift=-2pt}
			},
			dashedbox/.style={
				draw=black!60,
				dashed,
				line width=1pt,
				rectangle,
				rounded corners=3pt,
				align=left,
				inner sep=8pt,
				font=\footnotesize,
				fill=gray!5
			},
			photobox/.style={
				minimum width=3cm,
				minimum height=3.5cm,
				inner sep=0pt,
				text width=3cm,
				align=center,
				font=\large
			},
			flow_arrow/.style={
				->,
				>={Stealth[round, length=4mm, width=2.5mm]}, 
				line width=2pt,
				draw=black!70
			},
			note_arrow_left/.style={
				->,
				>={LaTeX[width=2mm, length=3mm]},
				line width=1.5pt,
				color=cyan!70!blue, 
				shorten >= 2pt, 
				shorten <= 2pt
			},
			note_arrow_right/.style={
				->,
				>={LaTeX[width=2mm, length=3mm]},
				line width=1.5pt,
				color=orange!80!red, 
				shorten >= 2pt,
				shorten <= 2pt
			}
			]
			
			
			\node[photobox] (bellman_photo) at (0,0) {
				\includegraphics[width=3cm]{graph/bellman.png} 
				\\ Richard E. Bellman
			};
			
			\node[mainbox, above=0.5cm of bellman_photo] (bellman_eq) {Bellman Optimality\\Equation};
            \node[mainbox, above=of bellman_eq] (rl) {Reinforcement\\Learning};
			\node[mainbox, above=of rl] (q_learn) {Q-learning};
			
			\node[above=0.2cm of q_learn, font=\huge\bfseries, text=black!50] (dots_left) {$\vdots$};
			
			\node[dashedbox, right=of q_learn, text width=6.2cm] (q_note_top) {
				Introduces State-Action Value Function:\\[3pt]
				$Q(s_t, a_t) =  r({s}_{t}, {a}_{t}) + \gamma Q(s_{t+1},\pi(s_{t+1}))$\\[3pt]
				$\bullet$ Maximize $Q(s,a)$ to optimize controller\\
				$\bullet$ Model-Free (Discrete)
			};
			\node[dashedbox, right=of bellman_eq, text width=6.2cm] (q_note_bottom) {
				$\bullet$ Equality Constraint\\
				$\bullet$ Learns Value Functions $V(s)$ and $Q(s,a)$\\
				$\bullet$ Driven by maximizing cumulative discounted reward\\
				$\bullet$ Value function guides system behavior\\
				$\bullet$ Sequential Nature
			};
			\draw[note_arrow_left] (q_learn) -- (q_note_top);
			\draw[note_arrow_left] (bellman_eq) -- (q_note_bottom);
			
			\draw[flow_arrow] (bellman_eq.north) -- (rl.south);
			\draw[flow_arrow] (rl.north) -- (q_learn.south);
			
			\node[photobox, right=10.5cm of bellman_photo] (lyap_photo) {
				\includegraphics[width=3cm]{graph/lyapunov.png} 
				\\ Aleksandr Lyapunov
			};
            \node[mainbox, above=0.5cm of lyap_photo] (lyap_theory) {Lyapunov Theory};
			\node[mainbox, above=of lyap_theory] (lyap_learn) {Stabilization\\Learning};
			\node[mainbox, above=of lyap_learn] (d_learn) {\textcolor{red!80!black}{D-learning}};
            \node[above=0.2cm of d_learn, font=\huge\bfseries, text=black!50] (dots_right) {$\vdots$};
			
			
			\node[dashedbox, right=of d_learn, text width=6.2cm] (d_note_top) {
				Proposed D-learning parallel to Q-learning, introducing function:\\[3pt]
				$D({s}_t,{a}_t) = \Delta V = V({s}_{t+1}) - V({s}_t)$\\[3pt]
				Achieved under \textcolor{red!80!black}{stability} premise:\\[3pt]
				$\bullet$ Minimize $D({s},{a})$ to \textcolor{red!80!black}{accelerate convergence}\\
				$\bullet$ \textcolor{red!80!black}{Model-Free}
			};
			\node[dashedbox, right=of lyap_theory, text width=6.2cm] (d_note_bottom) {
				$\bullet$ Inequality Constraint\\
				$\bullet$ Learns Lyapunov Function $V({s})$\\
				$\bullet$ Driven by satisfying stability\\
				$\bullet$ Lyapunov function guides system behavior\\
				$\bullet$ One-step Transition Nature
			};
			\draw[note_arrow_right] (d_learn) -- (d_note_top);
			\draw[note_arrow_right] (lyap_theory) -- (d_note_bottom);
			
			\draw[flow_arrow] (lyap_theory.north) -- (lyap_learn.south);
			\draw[flow_arrow] (lyap_learn.north) -- (d_learn.south);
        \end{tikzpicture}
  }
	\caption{Comparison of Q-learning and D-learning \cite{quan2024control}}
  \label{fig:DQ_comparison}
\end{figure}

This section takes policy learning in deterministic discrete systems $s_{t+1}=f(s_t,a_t)$ as an example to further demonstrate the similarities and differences in the mathematical essence and evolutionary paths of Stabilization Learning and Reinforcement Learning. Figure \ref{fig:DQ_comparison} illustrates the theoretical origins and characteristic comparison of these two paradigms.
	
The core task of reinforcement learning is to find the optimal policy $\pi^*$ to maximize the \textbf{cumulative discounted reward} under the discount factor $\gamma\in[0,1)$.
In the classical Dynamic Programming framework, the core evaluation metric is the state value function $V^*(s)$, which satisfies the Bellman optimality equation:
\begin{equation}
	V^*(s_t)= \max_{a}\left( r(s_t, a) + \gamma V^*(f(s_t, a)) \right).
\end{equation}
However, performing policy improvement directly based on the state-value function $V(s)$ necessitates comparing the successor states induced by different actions and their corresponding values. Consequently, the policy improvement process relies not only on the value estimation of the current state but also requires access to the relationship between actions and state transitions—specifically, a computable system transition model or state transition information. In environments characterized by unknown models, complex dynamics, or transition relationships that are difficult to model precisely, this dependency constrains the direct application of dynamic programming methods.

To overcome the limitations of model dependency, Watkins and Dayan \cite{watkins1992q} introduced the state-action value function $Q(s,a)$, which directly encodes the long-term performance following the execution of a specific action in a given state onto the state-action pair. Utilizing the identity $V^*(s_{t+1}) = \max_{a'} Q^*(s_{t+1}, a')$, the aforementioned Bellman equation can be expanded as:
\begin{equation}
	\begin{aligned}
		Q^*(s_t, a_t) &\triangleq r(s_t, a_t) + \gamma V^*(s_{t+1}) \\
		&= r(s_t, a_t) + \gamma \max_{a'} Q^*(s_{t+1}, a').
	\end{aligned}
\end{equation}
Through this paradigm shift, obtaining the optimal policy transforms into an extremum search directly targeting the $Q$ function:
\begin{equation}
\pi^*(s) = \arg\max_{a} Q^*(s,a).
\end{equation}
In this manner, policy evaluation and policy improvement can be updated via temporal-difference methods based on sampled state transitions, thereby eliminating the need to explicitly construct a complete system dynamics model. However, if policy improvement is executed directly based on the state-value function, it necessitates comparing the subsequent states induced by different actions and their corresponding values, which consequently relies on a computable system transition model or state transition information. In complex environments where models are unknown or difficult to obtain precisely, this dependency constrains the direct application of dynamic programming methods.
	
As system complexity increases, subsequent algorithms combined deep learning on the foundation of Q-learning, achieving policy learning from discrete to continuous spaces, and from low-dimensional to high-dimensional state spaces:
\begin{itemize}[leftmargin=*, itemsep=2pt]
	\item \textbf{Stability Improvement in Deep Q-Network (DQN)}: Mnih et al. \cite{mnih2015human} introduced the \textbf{Experience Replay} mechanism and the \textbf{Target Network}. This method achieved the estimation of the $Q$ function using neural networks.
	\item \textbf{Continuous Domain Expansion in Deep Deterministic Policy Gradient (DDPG)}: Lillicrap et al. \cite{lillicrap2016continuous} proposed DDPG combining the Actor-Critic architecture. This algorithm evaluates action values using the $Q$ function (Critic) and performs backpropagation updates using the deterministic policy gradient (Actor).
\end{itemize}
	
Unlike reinforcement learning, which pursues the maximization of cumulative rewards, the primary goal of stabilization learning is to guarantee system \textbf{stability}. This condition requires not only that the current state satisfies constraints but also that the system state asymptotically converges along its trajectory. In classical control theory, verifying Lyapunov stability conditions requires calculating the difference of the Lyapunov function along the system trajectory. This calculation process explicitly includes the next state $s_{t+1} = f(s_t, a_t)$.

To solve this problem, D-learning \cite{quan2024control} uses sampled data $(s_t, a_t, s_{t+1})$ to learn the $D$ function $D(s,a)$, aiming to approximate the natural evolution of the Lyapunov function along the system trajectory:
\begin{equation}
	D(s_t, a_t) = V(s_{t+1}) - V(s_t) = V(f(s_t, a_t)) - V(s_t).
\end{equation}
During the learning process, the algorithm iteratively learns the $D$ function directly. On this basis, D-learning transforms the stability differential inequality into an inequality constraint based on the $D$ function: $D(s, \pi(s)) \leq -W(s)$. This equation enables D-learning to learn control policies with stability guarantees in unknown dynamic environments.
	
After establishing the basic D-learning paradigm, addressing the issues of expensive sampling and high real-time requirements in practical applications, subsequent studies have made significant improvements in sample efficiency and computational robustness:
\begin{itemize}[leftmargin=*, itemsep=2pt]
	\item \textbf{Sample Efficiency (DOPT)}: DOPT \cite{shen2025dopt}, proposed by Shen et al., introduced an off-policy target mechanism. This method, \textbf{analogous to the Experience Replay mechanism in DQN \cite{mnih2015human}}, effectively solves the inefficiency issue of traditional stabilization learning (which can only utilize on-policy data) by reusing historical data in the buffer to update the $D$ function.
	\item \textbf{Online Robustness (DL-Clip)}: DL-Clip \cite{liu2025dlclip}, proposed by Liu et al., introduced a Clipping operation. This mechanism \textbf{draws inspiration from the trust-region clipping idea in PPO \cite{schulman2017proximal}}, suppressing fitting errors by limiting the update magnitude of the $D$ function, thereby enhancing the robustness of online control while guaranteeing closed-loop stability.
	\item \textbf{Cross-Task Adaptability (Reptile-D-Learning)}: Reptile-D-learning \cite{cao2026rdlearning}, proposed by Cao et al., \textbf{integrates D-Learning and meta-learning \cite{hospedales2022meta, finn2017model, nichol2018first} for variable-parameter systems}. It optimizes shared network initializations via multi-subtask training and first-order meta-updates to capture general system dynamics, thus quickly adapting to unknown parameters with little data and boosting Lyapunov control robustness against mass variation, payload interference and model errors.
\end{itemize}

Building on D-learning, which directly characterizes Lyapunov differences, DQ-learning \cite{Wang2026DQLearning} further introduces the state-action value function  used in Q-learning, providing a learning framework for jointly considering stability constraints and performance optimization. The basic idea is to use the D function to characterize whether a one-step transition satisfies practical stability or a desired convergence trend, while using the Q function to evaluate the policy in terms of long-term cumulative cost or reward. In this way, policy updates are not driven solely by reward maximization, but are performed under stability constraints or safety boundaries.

\subsubsection{Section Summary}
    
\begin{figure}[htbp]
    \centering
    \begin{tikzpicture}
        \definecolor{stabColor}{RGB}{100, 149, 237} 
        \definecolor{rlColor}{RGB}{255, 127, 80}    
        
        \def\R{4.5} 
        \def\D{2.0} 
        
        \begin{scope}
            \clip (-\D,0) circle (\R);
            \fill[stabColor, opacity=0.3] (-\D,0) circle (\R);
        \end{scope}
        
        \begin{scope}
            \clip (\D,0) circle (\R);
            \fill[rlColor, opacity=0.3] (\D,0) circle (\R);
        \end{scope}
        
        \draw[stabColor!80!black, thick] (-\D,0) circle (\R);
        \draw[rlColor!80!black, thick] (\D,0) circle (\R);
        
        \node[align=center, text width=3cm, color=black!80] at (-4.2, 0) {
            \textbf{Stabilization Learning}\\[0.5em]
            \scriptsize
            \textbf{Feasibility-oriented}\\
            Based on stability\\
            $\dot{V} \leq -W$\\
            Inequality constraint
        };
        
        \node[align=center, text width=3cm, color=black!80] at (4.2, 0) {
            \textbf{Reinforcement Learning}\\[0.5em]
            \scriptsize
            \textbf{Optimality-oriented}\\
            Performance metric extremum search\\
            $\min \int l(s,a) \text{d} t$\\
            Soft constraint/generalized task
        };
        
        \node[align=center, text width=3.8cm, color=black!90] at (0, 0) {
            \textbf{Unified Paradigm}\\[0.5em]
            \scriptsize
            $\bullet$ Positive definite cost corresponds to positive definite value function $V^*$ \\
            $\bullet$ Reinforcement learning policy satisfies stabilization learning stability through $V^*$ \\
            $\bullet$ Stabilization learning policy satisfies reinforcement learning optimality through $V^*$ \\
        };
        
        \node[above, xshift=-0.5cm, font=\bfseries\large, color=stabColor!60!black] at (-\D, \R) {Stability Oriented};
        \node[above, xshift=0.5cm, font=\bfseries\large, color=rlColor!60!black] at (\D, \R) {Optimality Oriented};
    \end{tikzpicture}
    \caption{Paradigm unification and complementary relationship between Stabilization Learning and Reinforcement Learning. The diagram illustrates the mathematical intersection of the two paradigms: under the settings of positive definite costs and an infinite horizon, the optimal value function $V^*$ of reinforcement learning naturally constitutes a Lyapunov certificate for stabilization learning; conversely, by introducing slack variables, the feasible solution space of stabilization learning collapses into the optimal solution of reinforcement learning. The regions on both sides respectively represent pure stabilization tasks (relaxing optimality for robustness) and pure reinforcement learning tasks (relaxing stability to accommodate generalized objectives).}
    \label{fig:venn_relationship}
\end{figure}
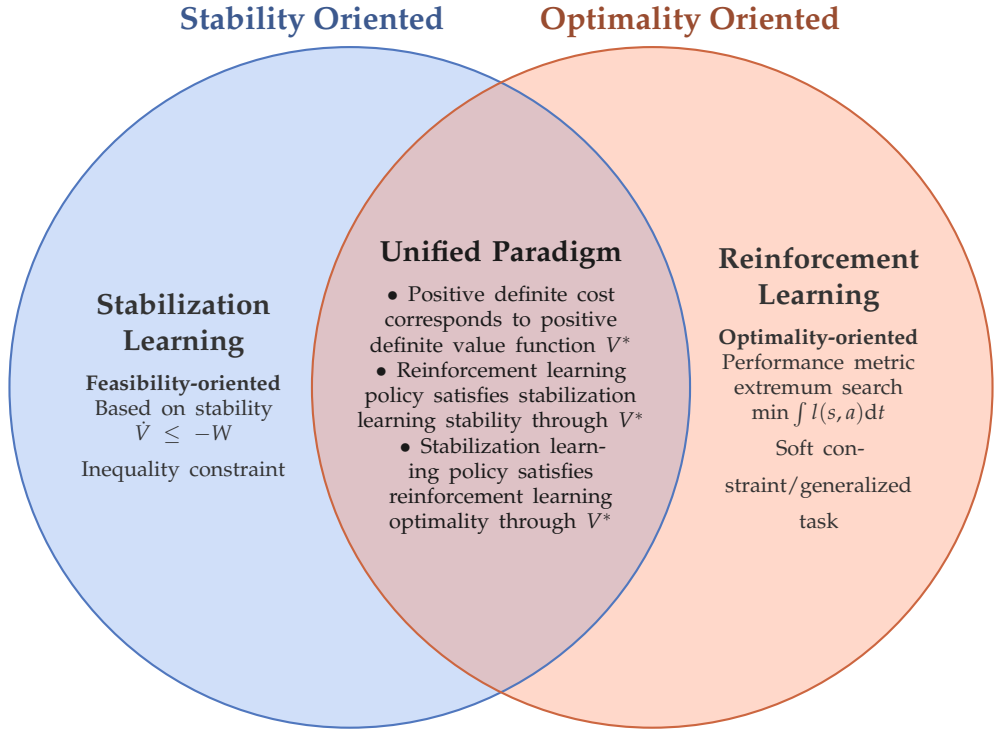

Under the unified framework of deterministic continuous-time systems, stabilization learning and reinforcement learning share an intrinsic theoretical connection and mathematical consistency.
\begin{itemize}[leftmargin=*, itemsep=2pt]
    \item \textbf{From a feasibility perspective}: Stabilization learning aims to solve the feasible solution space of Lyapunov inequalities. Optimal control can be viewed as a unique specific solution within this space that satisfies the HJB equality constraint; that is, by introducing a slack term consistent with the task cost, the broad stability boundary shrinks to the optimal trajectory.
    \item \textbf{From an optimality perspective}: Reinforcement learning is dedicated to the extremum search of the value function. Under the settings of an infinite horizon and positive definite cost functions, the optimality condition (HJB equation) naturally entails the Lyapunov stability constraint, thereby achieving an endogenous unity of performance optimization and system stabilization.
\end{itemize}
In summary, the two constitute complementary paradigms in control theory: stabilization learning trades inequality relaxation for adaptability to generalized tasks, while reinforcement learning naturally satisfies system stability requirements under certain conditions. Their relationship is illustrated in Figure \ref{fig:venn_relationship}.

\section{Transformation of Stabilization Learning Problems}

The previous sections elaborated on the definition and mathematical framework of stabilization learning. This section will explore how to uniformly transform various real-world problems into the mathematical framework of stabilization learning, thereby achieving a paradigm shift from control theory to machine learning.

\subsection{Multi-Agent Cooperative Pursuit}

\begin{figure}[H]
  \centering
  \includegraphics[width=1\textwidth]{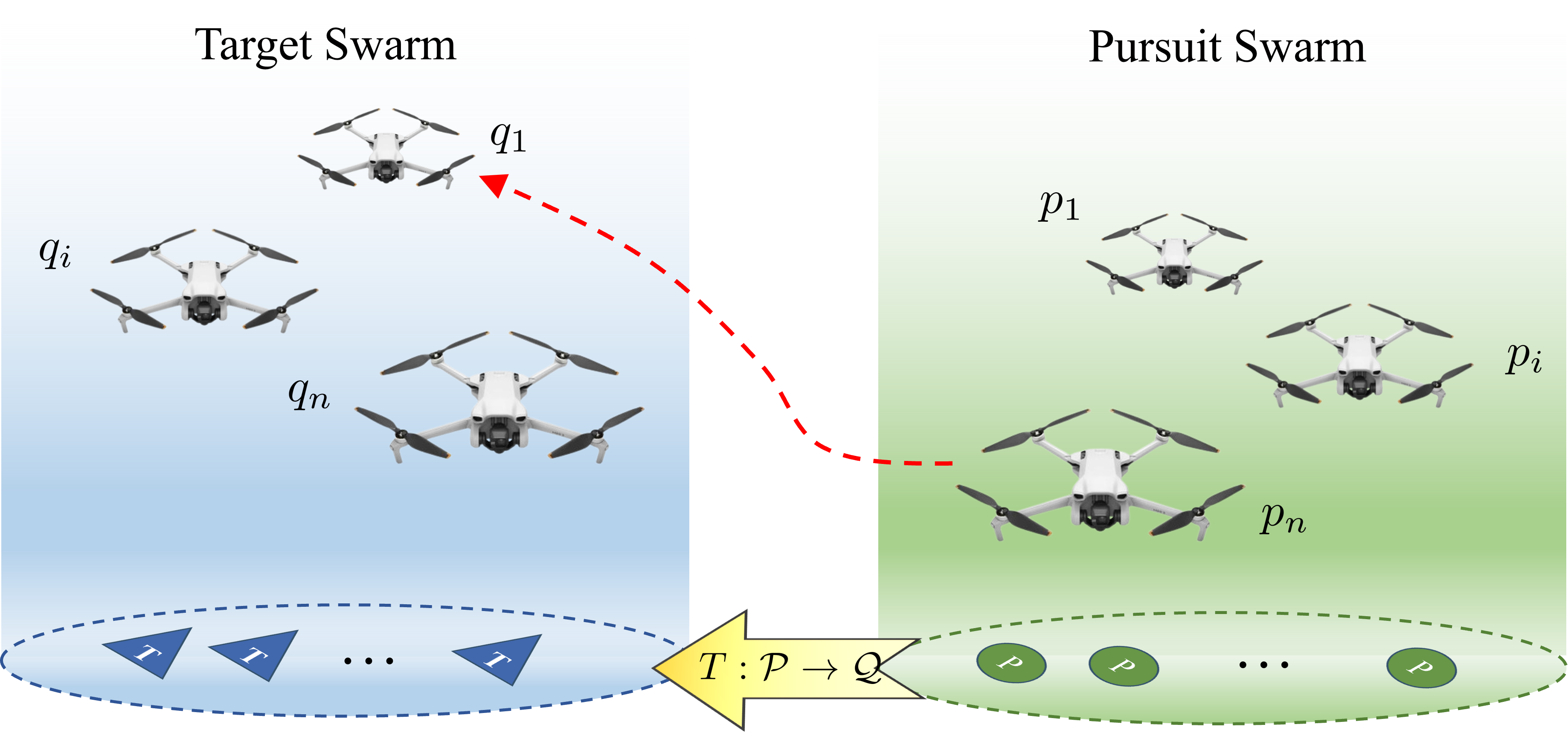}
  \caption{Schematic diagram of multi-agent cooperative pursuit task}
  \label{fig:cluster}
\end{figure}

\textbf{Problem Description}:
Consider a scenario in three-dimensional space where $N$ pursuing UAVs (Unmanned Aerial Vehicles) are chasing $N$ target UAVs, as shown in Figure \ref{fig:cluster}. The objective is to achieve a one-to-one tracking mapping from the pursuing swarm to the target swarm, ultimately allowing the pursuing UAVs to approach and capture the target UAVs.

\begin{itemize}[leftmargin=*, itemsep=2pt]
  \item Spatial coordinates of the target UAV swarm: $q_i(t) \in \mathbb{R}^3, i = 1,2,3,\cdots,N.$
  \item Spatial coordinates of the pursuing UAV swarm: $p_i(t) \in \mathbb{R}^3, i = 1,2,3,\cdots,N.$
\end{itemize}

During the pursuit, online negotiation and dynamic assignment are required to ensure that each pursuing UAV uniquely locks onto and continuously tracks a specific unassigned target. Throughout the task, conflicts such as ``many-to-one'' or ``one-to-many'' pursuits must be completely avoided, thereby maintaining a strict one-to-one assignment. This relationship can be described as:
\begin{equation}
T: \mathcal{P} \to \mathcal{Q},
\end{equation}
where $\mathcal{P}$ is the pursuing swarm, $\mathcal{Q}$ is the target swarm, {and $T[p_i] = q_j \ (i,j = 1,2,\cdots,N)$ indicates that pursuer $p_i$ is tracking target $q_j$.}

The core control variable for the pursuing swarm during the chase is velocity. The objective is to adjust the velocity control variables of the pursuing swarm:
\[ v_i \in \mathbb{R}^3, i = 1,2,\cdots,N, \]
So that the positions of the pursuing swarm gradually approach those of the target swarm, realizing the encirclement and capture of the target swarm by the pursuing swarm.

{ The ultimate objective of stabilization learning is to achieve global positional synchronization between the two swarms, relaxing the strict constraints on the specific distances of predefined ``pursuer-target" UAV pairs. In other words, within the stabilization learning framework, our primary concern is the generalized distance between the pursuer UAV coordinate set \(\{p_1,p_2,p_3,\dots,p_N\}\) and the target UAV coordinate set \(\{q_1,q_2,q_3,\ldots,q_N\}\). Since we solely evaluate whether the elements of the two sets are sufficiently close on a one-to-one basis, an \textbf{assignment-aware metric} is introduced to quantify the overall spatial proximity between the two swarms, formulated as follows:
\begin{equation}
    d = \min_{T}\max_{i}\|p_i(t) - T[p_i(t)]\| = \min_{T}\max_{i}\|p_i(t) - q_j(t)\|,
\end{equation}
where \(T\) denotes the one-to-one mapping (assignment) between pursuers and targets. Given a predefined threshold \( \epsilon > 0 \), the pursuit task is deemed successful when this metric between the two swarms falls below the threshold.}

\begin{highlightbox}{Multi-Agent Cooperative Pursuit Problem}{multi-agent-pursuit}
  Transforming the aforementioned multi-agent cooperative pursuit problem into the six-tuple $ (\mathcal{S},\mathcal{A}, \mathcal{P}, \pi, \mathcal{S}_{\mathrm{d}}, d)$ under the stabilization learning framework yields the following representations.
  \begin{itemize}[leftmargin=*, itemsep=2pt]
    \item \textbf{State Space ($\mathcal{S}$) and State ($s$)}\\
    Defined as the coordinate set of the pursuing UAV swarm, encompassing all possible positions the UAVs can reach, as shown below:
    \[
    \mathcal{S} = \left\{ s \mid s = [p_1^\top \quad p_2^\top \quad \cdots \quad p_N^\top]^\top \in \underbrace{\mathbb{R}^3 \times \mathbb{R}^3 \times \cdots \times \mathbb{R}^3}_{N} \right\}.
    \]
    \item \textbf{Action Space ($\mathcal{A}$)}\\
    Defined as the set of velocity control variables for the pursuing UAV swarm, as shown below:
    \[
    \mathcal{A} = \left\{ a \mid a = [v_1^\top \quad v_2^\top \quad \cdots \quad v_N^\top]^\top \in \underbrace{\mathbb{R}^3 \times \mathbb{R}^3 \times \cdots \times \mathbb{R}^3}_{N} \right\}.
    \]
    \item \textbf{Plant ($\mathcal{P}$)}\\
    The evolutionary model of the controlled pursuing UAV swarm is its kinematic equation. The state evolution at the next moment is deterministically decided by the state and control variables at the current moment, as shown below:
    \[
    \dot{s} = a, \quad s \in \mathcal{S}, a \in \mathcal{A}. 
    \]
    \item \textbf{Desired State Set ($ \mathcal{S}_{\mathrm{d}}$)}\\
    Defined as the real-time position coordinate set of the target UAV swarm, which continuously updates over time, as shown below:
    \[
    \mathcal{S}_{\mathrm{d}}(t) = \left\{ s_{\mathrm{d}} = [\, q_1^\top(t) \quad q_2^\top(t) \quad \cdots \quad q_N^\top(t)\, ]^\top \in \mathcal{S} \right\}.
    \]
    \item \textbf{Metric Function ($d$)}\\
    Measures the overall closeness between the pursuing UAV swarm set and the target UAV swarm set. The assignment-aware metric can be employed, as shown below:
    \[
    d(s(t), \mathcal{S}_{\mathrm{d}}(t)) = \min_{T}\max_{i}\|p_i(t) - T[p_i(t)]\| = \min_{T}\max_{i}\|p_i(t) - q_j(t)\|.
    \]
    \item \textbf{Policy ($\pi$)}\\
    The role of policy $\pi$ is to adjust the velocity control variable $a$ based on the system state $s(t)$ and the desired target state $\mathcal{S}_{\mathrm{d}}(t)$. It aims to flexibly adjust the pursuers' velocities (action $ a(t) = \pi(s(t),\mathcal{S}_{\mathrm{d}}(t))$) so that the system state $s(t)$ gradually approaches the desired state set $ \mathcal{S}_{\mathrm{d}}(t)$ over time. That is, as $t \to \infty$, $ d(s(t), \mathcal{S}_{\mathrm{d}}(t)) \to 0$.
  \end{itemize}
\end{highlightbox}

\subsection{Visual Servo-Based Position Stabilization of Multi-Rotor UAVs}\label{sec:drone_hover}

\begin{figure}[H]
	\centering
	\resizebox{1.0\textwidth}{!}{
		\begin{tikzpicture}[
			>=latex, 
			node distance=2cm, 
			font=\small,
			block/.style={draw, rectangle, rounded corners, minimum height=3em, minimum width=4.5em, align=center, fill=blue!5, draw=blue!40, thick},
			process/.style={draw, rectangle, minimum height=3em, minimum width=4.5em, align=center, fill=orange!10, draw=orange!40, thick},
			image_frame/.style={draw, rectangle, minimum size=2.8cm, fill=white, draw=gray, thick}, 
			feature_point/.style={circle, fill=red, inner sep=1.8pt},
			target_point/.style={circle, fill=green!60!black, inner sep=1.8pt},
			line/.style={->, thick, draw=gray!80},
			dashed_line/.style={->, dashed, thick, draw=gray!60}
			]
			
			\node (ground_center) at (0, -3.5) {};
			\draw[fill=gray!10, draw=none] (-2.2, -4.0) rectangle (2.2, -3.0);
			\node at (0, -4.3) {Physical World (3D)};
			
			\coordinate (P1) at (-1.2, -3.5);
			\coordinate (P2) at (0.6, -3.3);
			\coordinate (P3) at (-0.2, -3.7);
			\foreach \p in {P1,P2,P3} \fill[black] (\p) circle (2pt);
			\node[right, font=\footnotesize] at (P2) {Features};
			
			\node (drone) at (0, 0) [inner sep=0] {
				\tikz {
					\draw[thick, fill=gray!20] (-0.9,0) -- (0.9,0);
					\draw[thick] (0,0) -- (0,0.4);
					\draw[thick, fill=white] (-0.35,0.4) rectangle (0.35,0.55); 
					\draw[thick] (-0.9,0) -- (-0.9,0.3); \draw[thick] (-1.1,0.3) -- (-0.7,0.3);
					\draw[thick] (0.9,0) -- (0.9,0.3); \draw[thick] (0.7,0.3) -- (1.1,0.3); 
					\draw[fill=black] (-0.15,-0.15) rectangle (0.15,0);
					\draw[fill=gray] (-0.15,-0.15) -- (-0.3,-0.4) -- (0.3,-0.4) -- (0.15,-0.15) -- cycle;
				}
			};
			\node[left=0.6cm of drone, align=right] (state_label) {Current Pose\\ $s(t)$};
			
			\draw[dashed, gray] (0, -0.4) -- (-1.4, -3.0);
			\draw[dashed, gray] (0, -0.4) -- (1.4, -3.0);
			
			\node[image_frame, right=3.5cm of drone] (img_current) {};
			\node[below=2pt, font=\footnotesize] at (img_current.north) {Current Image $I$};
			
			\coordinate (y1) at ($(img_current.center) + (-0.6, 0.4)$);
			\coordinate (y2) at ($(img_current.center) + (0.3, 0.5)$);
			\coordinate (y3) at ($(img_current.center) + (-0.2, 0.0)$);
			\foreach \p in {y1,y2,y3} \node[feature_point] at (\p) {};
			\node[feature_point, label=right:{$y(t)$}] at (y2) {};
			
			\draw[line] (drone) -- node[above, font=\footnotesize] {Imaging} node[below, font=\footnotesize] {$h_{\text{cam}}(\cdot)$} (img_current);
			
			\node[block, right=1.6cm of img_current] (extraction) {Feature\\Extraction\\$\phi_I(\cdot)$};
			\draw[line] (img_current) -- (extraction);
			
			\node[circle, draw, thick, right=1.2cm of extraction, inner sep=3pt] (sum) {$\Sigma$};
			\node[below right=-3pt of sum] {$-$};
			\node[above left=-3pt of sum] {$+$};
			\draw[line] (extraction) -- node[above] {$y$} (sum);
			
			\node[image_frame, below=3.5cm of img_current] (img_desired) {};
			\node[below=2pt, font=\scriptsize] at (img_desired.north) {Reference Image $I_{\mathrm{d}}$};
			
			\coordinate (yd1) at ($(img_desired.center) + (-0.3, 0.2)$);
			\coordinate (yd2) at ($(img_desired.center) + (0.4, 0.3)$);
			\coordinate (yd3) at ($(img_desired.center) + (0.0, -0.2)$);
			\foreach \p in {yd1,yd2,yd3} \node[target_point] at (\p) {};
			\node[target_point, label=right:{\color{green!60!black}$y_{\mathrm{d}}$}] at (yd2) {};
			
			\node[block, right=1.8cm of img_desired] (extraction_d) {Feature\\Extraction};
			\draw[line] (img_desired) -- (extraction_d);
			
			\draw[line] (extraction_d) -| node[right, pos=0.9] {Desired $y_{\mathrm{d}}$} (sum);
			
			\node[process, right=1.5cm of sum, align=center] (controller) {Control Policy\\ $\pi(a|s, y, y_{\mathrm{d}})$};
			\draw[line] (sum) -- node[above, align=center, font=\footnotesize] {Error\\$e = y - y_{\mathrm{d}}$} (controller);
			
			\draw[line] (controller.south) -- ++(0,-2.0) -|
			node[above, pos=0.25, font=\footnotesize] {Control Input $a$ (Thrust, Torque)} ($(drone.south)+(0,-1.8)$) -- (drone.south);
			
			\node[fit=(img_current) (extraction), dashed, draw=blue!30, inner xsep=35pt, inner ysep=10pt, xshift=-33pt, label={[yshift=2pt]above:{Composite Mapping $h(s)$}}] {};
			
			\node[align=left, font=\small, color=black!80, anchor=north west] at (-3.5, 2.5) {
				\textbf{System Dynamics:}\\
				$\dot{s} = f(s, a)$\\
				$y = h(s)$
			};
			
			\node[align=left, font=\small, color=black!80, anchor=west] at (5.5, -4.3) {
				\textbf{Control Objective:}\\
				$\lim_{t \to \infty} \|y(t) - y_{\mathrm{d}}\| = 0$
			};
		\end{tikzpicture}
	}
	\caption{Schematic diagram of closed-loop control architecture for visual servo-based position stabilization of multi-rotor UAVs.}
	\label{fig:visual_servo_structure}
\end{figure}
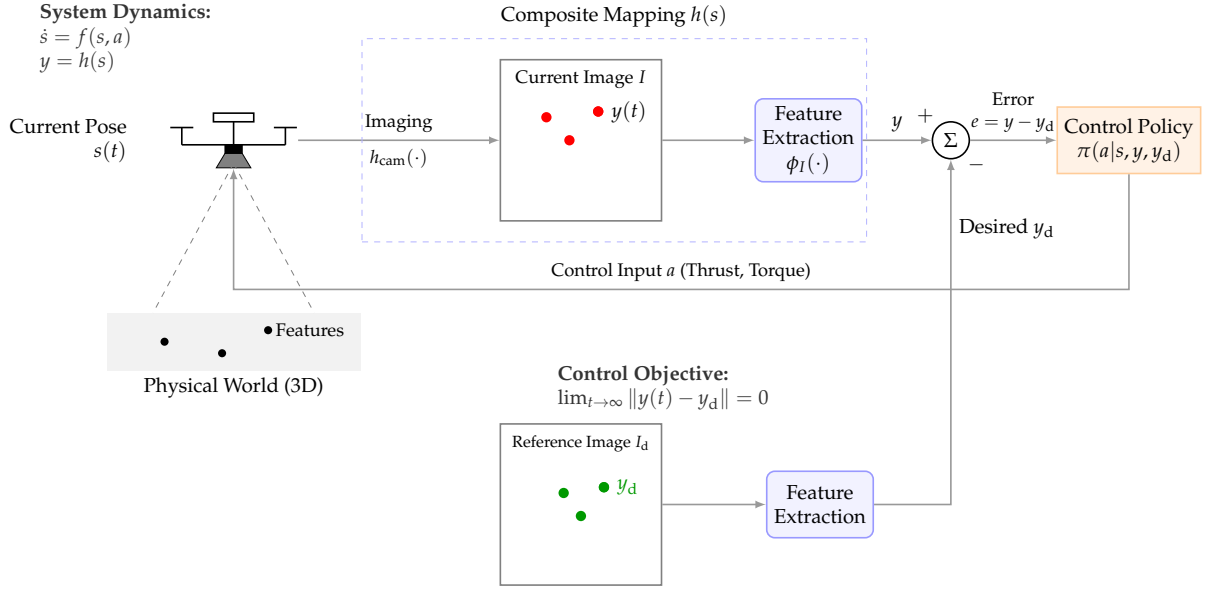
	
\textbf{Problem Description}:
Consider visual servo-based position stabilization of a multi-rotor UAV \cite{chaumette2006visual}, which refers to utilizing feedback from onboard visual sensors to construct a closed-loop control policy, as shown in Figure \ref{fig:visual_servo_structure}. This policy drives the UAV to asymptotically converge from an arbitrary initial state and stabilize at the desired relative pose $ (p_{\mathrm{d}}, R_{\mathrm{d}})$ defined by the reference image $ I_{\mathrm{d}}$. The core of this problem lies in effectively stabilizing the UAV state by minimizing the error in the image feature space.

First, a general multi-rotor UAV dynamics model is established \cite{quan2017introduction}. The UAV state includes the position $p \in \mathbb{R}^3$ in the inertial frame, the velocity $v \in \mathbb{R}^3$ in the inertial frame, and the rotation matrix $R \in \text{SO}(3)$ of the body frame relative to the inertial frame. Adopting the Computed Torque and Body Rate (CTBR) control mode, the control inputs are the total axial body thrust $F \in \mathbb{R}$ and the body angular velocity $\omega \in \mathbb{R}^3$:
\begin{equation} \label{eq:uav_dynamics}
  \begin{cases}
    \dot{p} = v \\
    \dot{v} = g e_3 - \frac{F}{m} R e_3 \\
    \dot{R} = R [\omega]_\times 
  \end{cases},
\end{equation}
where $m$ is the mass, $g$ is the gravitational acceleration, $e_3=[0\quad 0\quad 1]^\top$ denotes the unit vector, and $[\cdot]_\times$ denotes the skew-symmetric matrix mapping of a vector.

The UAV is equipped with a camera, and its imaging process can be abstracted as a composite mapping. The camera imaging function $h_\text{cam}(\cdot)$ maps the UAV pose to the image $I$; the feature extraction function $\phi_{I}(\cdot)$ extracts low-dimensional structured feature information $y\in\mathbb{R}^p$ (such as feature point coordinates, line parameters, image moments, etc.) from the image \cite{quan2024control}:
\begin{equation} \label{eq:cam_func}
	y = \phi_{I}(h_\text{cam}(p, R)) \equiv h'(p, R).
\end{equation}
Here, $h^{\prime}(p, R)$ is introduced because although the UAV state $(p,v,R)$ contains velocity information, the feature observation of a single frame essentially depends only on the spatial geometric properties of the camera (position $p$ and pose $R$), independent of the linear velocity $v$.

The control objective based on visual servoing is to design a policy $\pi$ such that the current feature $y$ tracks the desired feature $ y_{\mathrm{d}}$ extracted from the desired image $ I_{\mathrm{d}}$. That is, as $t \to \infty$, the feature error converges to zero:
\begin{equation} \label{eq:tracking_error} 
	\lim_{t \to \infty} \| y(t) - y_{\mathrm{d}} \| = 0.
\end{equation}
Based on the system dynamics equation \eqref{eq:uav_dynamics} and the output function \eqref{eq:cam_func}, this problem can be formulated as a tracking problem based on output features.

\begin{highlightbox}{Visual Servo-Based Position Stabilization of Multi-Rotor UAVs: Formulation as a Tracking Problem}{stability-control-tracking}
  Transformed into a seven-tuple $ (\mathcal{S}, \mathcal{A}, \mathcal{P}, \pi, h, \mathcal{Y}_{\mathrm{d}}, d_\mathcal{Y})$ tracking problem under the stabilization learning framework, it is represented as follows.
  \begin{itemize}[leftmargin=*, itemsep=2pt]
    \item \textbf{State Space (\(\mathcal{S}\)) and State (\(s\))}\\
    Defined as the set of the full physical states of the multi-rotor UAV, including position, velocity, and pose information:
    $$\mathcal{S}=\{s \mid s=(p, v, R) \in \mathbb{R}^3 \times \mathbb{R}^3 \times \text{SO}(3)\}.$$
    
    \item \textbf{Action Space (\(\mathcal{A}\))}\\
    Defined as the set of computed thrust and angular velocity control variables for the UAV:
    $$\mathcal{A}=\{a \mid a=(F, \omega) \in  \mathbb{R} \times \mathbb{R}^3\}.$$
    
    \item \textbf{Plant (\(\mathcal{P}\))}\\
    The evolution model of the plant is the dynamics model of the multi-rotor UAV \eqref{eq:uav_dynamics}, where the state evolution at the next moment is deterministically decided by the state and control variables at the current moment:
    $$\dot{s} = f(s, a), \quad s\in\mathcal{S},a\in\mathcal{A}.$$
    
    \item \textbf{Output Function (\(h\))}\\
    Establishes the mapping from the state space to the output space. The output $y$ is defined as the features extracted from the image. The single-frame feature depends only on the geometric state (position $p$ and pose $R$):
    $$h(s) \triangleq h'(p, R).$$
    
    \item \textbf{Tracking Target (\( \mathcal{Y}_{\mathrm{d}} \))}\\
    Consists of the desired feature $ y_{\mathrm{d}}$ extracted from the desired image $ I_{\mathrm{d}}$:
    $$ \mathcal{Y}_{\mathrm{d}} = \{ y_{\mathrm{d}} \in \mathbb{R}^p \}.$$
    
    \item \textbf{Metric Function (\(d_\mathcal{Y}\))}\\
    Measures the difference between the current output feature $y$ and the desired feature $ y_{\mathrm{d}}$. The Euclidean distance is typically used:
    $$ d_\mathcal{Y}(y, \mathcal{Y}_{\mathrm{d}}) = \| y - y_{\mathrm{d}} \|.$$
    
    \item \textbf{Policy (\(\pi\))}\\
    The objective of policy $\pi$ is to design the control input $a$ using the state information $s$ and the desired feature $ y_{\mathrm{d}}$. It aims to progressively drive the system output $y$ to approach the tracking target $ \mathcal{Y}_{\mathrm{d}}$ over time by adjusting the thrust and angular velocity of the UAV. That is, as $t \to \infty$, $ d_\mathcal{Y}(y(t), \mathcal{Y}_{\mathrm{d}}(t)) \to 0$, while ensuring that the state $s$ remains bounded.
  \end{itemize}
\end{highlightbox}

\subsection{Interception Task under Field of View Constraints}

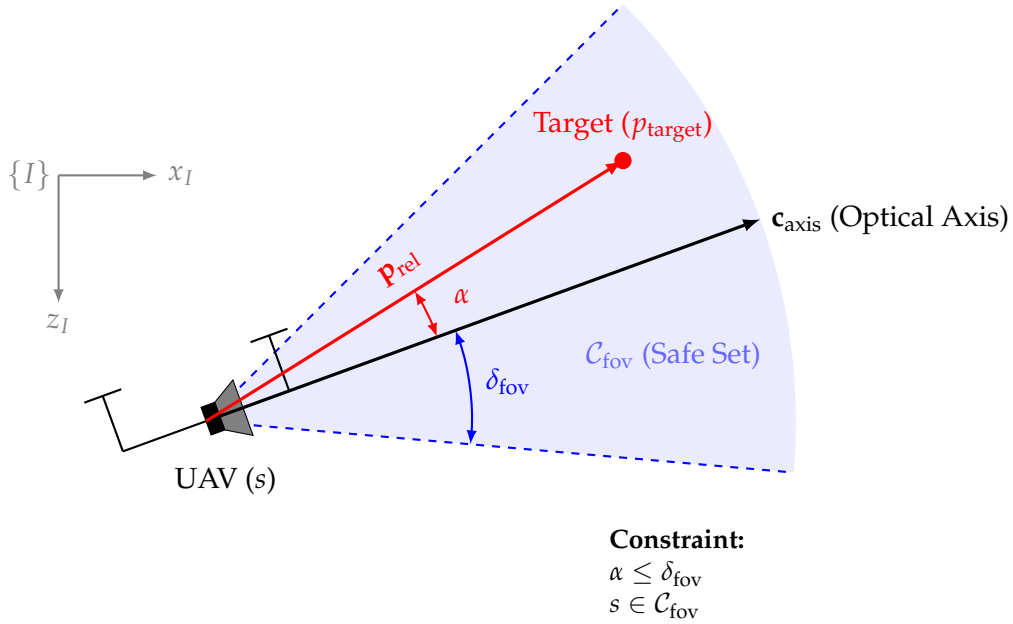
\begin{figure}[H]
    \centering
    \begin{tikzpicture}[>=latex, scale=1.3]
        
        \def\angBody{20}      
        \def\angFOV{25}       
        \def\angTarget{32}    
        \def\radDist{6.0}     
        \def\targetDist{5.0}  
        
        \fill[blue!8] (0,0) -- (\angBody-\angFOV:\radDist) arc (\angBody-\angFOV:\angBody+\angFOV:\radDist) -- cycle;
        \node[blue!60] at (\angBody-12:\radDist*0.8) {$\mathcal{C}_\text{fov}$ (Safe Set)};
        
        \draw[blue, dashed, thick] (0,0) -- (\angBody+\angFOV:\radDist);
        \draw[blue, dashed, thick] (0,0) -- (\angBody-\angFOV:\radDist);
        
        \draw[<->, blue, thick] (\angBody:2.7) arc (\angBody:\angBody-\angFOV:2.7);
        \node[blue] at (\angBody - 0.5*\angFOV : 3.1) {$\delta_\text{fov}$};
        
        \draw[->, thick, gray] (-1.5, 2.5) -- (-0.5, 2.5) node[right] {$x_I$};
        \draw[->, thick, gray] (-1.5, 2.5) -- (-1.5, 1.2) node[below] {$z_I$};
        \node[gray] at (-1.8, 2.5) {$\{I\}$};
        
        \begin{scope}[rotate=\angBody]
            \draw[thick, fill=gray!20] (-0.9,0) -- (0.9,0);
            
            \draw[thick] (-0.9,0) -- (-0.9,0.6);
            \draw[thick] (-1.1,0.6) -- (-0.7,0.6); 
            
            \draw[thick] (0.9,0) -- (0.9,0.6);
            \draw[thick] (0.7,0.6) -- (1.1,0.6); 
            
            \draw[fill=black] (0,-0.15) rectangle (0.15,0.15);
            
            \draw[fill=gray] (0.15, 0.15) -- (0.4, 0.3) -- (0.4, -0.3) -- (0.15, -0.15) -- cycle;
        \end{scope}
        
        \node at (0.2, -0.6) {UAV ($s$)};
        
        \draw[->, line width=1.2pt, black] (0,0) -- (\angBody:\radDist) node[right] {$\mathbf{c}_\text{axis}$ (Optical Axis)};
        
        \coordinate (Target) at (\angTarget:\targetDist);
        \fill[red] (Target) circle (2.5pt) node[above=3pt] {Target ($p_\text{target}$)};
        \draw[->, line width=1.2pt, red] (0,0) -- (Target) node[midway, above, sloped, yshift=2pt] {$\mathbf{p}_\text{rel}$};
        
        \draw[<->, red, thick] (\angBody:2.5) arc (\angBody:\angTarget:2.5);
        \node[red] at (\angBody*0.5 + \angTarget*0.5 : 2.9) {$\alpha$};
        
        \node[anchor=north west, align=left, font=\small] at (4.0, -1.0) {
            \textbf{Constraint:}\\
            $\alpha \leq \delta_\text{fov}$\\
            $s \in \mathcal{C}_\text{fov}$
        };
    \end{tikzpicture}
    \caption{Schematic diagram of geometric relationships in the interception task under FOV constraints. The blue area represents the FOV safe set $\mathcal{C}_\text{fov}$, and the UAV model includes the fuselage and the forward-facing camera.}
    \label{fig:Interception_FOV}
\end{figure}

\textbf{Problem Description}:
Consider a UAV system executing an interception task in three-dimensional space. The dynamics model of the interception system directly adopts the multi-rotor rigid body model \eqref{eq:uav_dynamics} described in the previous section. The UAV state includes the position $p \in \mathbb{R}^3$ in the inertial frame, the velocity $v \in \mathbb{R}^3$ in the inertial frame, and the rotation matrix $R \in \text{SO}(3)$ of the body frame relative to the inertial frame. Adopting the Computed Torque and Body Rate control mode, the control inputs are the total axial body thrust $F \in \mathbb{R}$ and the body angular velocity.

In the interception task under field of view (FOV) constraints, the monocular camera mounted on the UAV is the sole source of guidance information, and its imaging optical axis is fixed to the body frame (denoted as the unit vector $c_\text{axis}$, e.g., the forward axis of the body). The target position to be intercepted is represented as $p_\text{target}$, and the position vector relative to the UAV is $p_\text{rel} = p_\text{target} - p$. The geometric relationship is illustrated in Figure \ref{fig:Interception_FOV}. The core control objective of the interception task is to adjust the UAV's pose so that its optical axis aligns with the target or follows the desired pose generated by the guidance law. We transform the interception problem into a pose tracking problem, primarily based on the following physical facts:
\begin{itemize}[leftmargin=*, itemsep=2pt]
\item The multi-rotor UAV is an underactuated system, requiring a change in pose to direct the thrust vector toward the desired direction;
\item Visual guidance requires that the target must always remain within the camera's FOV, which means the body pose must be strictly constrained.
\end{itemize}

Therefore, in this subtask, the output function is defined as the pose extraction mapping \cite{yang2025jgcd,yang2025tcst}:
{
\begin{equation}\label{eq:hrr}
    R = h_R(s), \quad s=(p,v,R),
\end{equation}
 where $h_R$ extracts the rotation-matrix
component from the state.} The control objective is to design a policy such that the current pose $R$ asymptotically converges to the desired pose $R_\text{d}$. Utilizing the orthogonal property of rotation matrices (i.e., $R_\text{d} R_\text{d}^\top = I$), this tracking objective is mathematically equivalent to making the pose error matrix approach the identity matrix over time. That is, the Frobenius norm $\| \cdot \|_F$ of the difference between the two matrices satisfies:
\begin{equation} 
	\lim_{t \to \infty} \| I - R^\top(t) R_\text{d} \|_F = 0.
\end{equation}
Meanwhile, the system faces a hard FOV safety constraint. The line-of-sight angle deviation $\alpha$ must always be less than the FOV half-angle $\delta_\text{fov}$, as shown in Figure \ref{fig:Interception_FOV}. The set of all physical states satisfying this geometric condition is defined as the \textbf{FOV Safe Set} $\mathcal{C}_\text{fov}$. The policy must ensure that the physical state always belongs to $\mathcal{C}_\text{fov}$.

{Let $b_{\mathrm{fov}}(s)=\delta_{\mathrm{fov}}-\alpha(s)$. We explicitly define the FOV safe set as
\[
\mathcal{C}_{\mathrm{fov}}=\{s\in\mathcal{S} \mid b_{\mathrm{fov}}(s)\ge 0\}=\{s\in\mathcal{S} \mid \alpha(s)\le\delta_{\mathrm{fov}}\}.
\]
Assuming that $\alpha(s)$ is continuous, $\mathcal{C}_{\mathrm{fov}}$ is closed relative to the state
space $\mathcal{S}$. The boundary $\partial\mathcal{C}_{\mathrm{fov}}=\{s\in\mathcal{S} \mid \alpha(s)=\delta_{\mathrm{fov}}\}$ is included in the safe set, where the FOV constraint becomes active.

To guarantee safety invariance, the control input is required to satisfy
the barrier condition
\begin{equation}\label{eq:cbf_condition}
    \dot{b}_{\mathrm{fov}}(s,u)+\kappa b_{\mathrm{fov}}(s)\ge 0,\quad \kappa>0.
\end{equation}
Thus, if $s(0)\in\mathcal{C}_{\mathrm{fov}}$, then $b_{\mathrm{fov}}(s(t))\ge 0$ for all $t\ge 0$, which implies
$s(t)\in\mathcal{C}_{\mathrm{fov}}$ and the trajectory never enters the barrier space $\mathcal{B}$.}

In summary, the interception control objective under FOV constraints can be formalized as: finding the control inputs (body thrust $F$ and angular velocity $\omega$) to drive the body pose $R$ to asymptotically converge to the desired pose $R_\text{d}$ under the premise of strictly satisfying the FOV geometric constraints. That is, solving for a control policy to satisfy the following simultaneous conditions:
\begin{equation}
    \begin{cases}
    \lim_{t \to \infty} \| I - R^\top(t) R_\text{d} \|_F = 0 \\
    (p(t), v(t), R(t)) \in \mathcal{C}_\text{fov}, \quad \forall t \geq 0
    \end{cases},
\end{equation}
where the first equation represents the asymptotic stability requirement of pose tracking, utilizing the Frobenius norm to characterize the degree of coincidence between the current pose matrix and the desired pose matrix; the second equation represents the FOV safety requirement during the interception process, ensuring that the line-of-sight angle deviation $\alpha(t)$ is kept within the allowable range $\delta_\text{fov}$ at any moment to prevent target loss.

\begin{highlightbox}{Formalization of the Interception Task under FOV Constraints}{interception-fov-formal}
    Transforming this problem into a standard constrained control eight-tuple $(\mathcal{S}, \mathcal{A}, \mathcal{P}, \pi, h, \mathcal{Y}_{\mathrm{d}}, \mathcal{B}, d_\mathcal{Y})$ yields the following representations.
    \begin{itemize}[leftmargin=*, itemsep=2pt]
        \item \textbf{State Space (\(\mathcal{S}\)) and State (\(s\))}\\
        Defined as the set of the full physical states of the multi-rotor UAV:
        \[ \mathcal{S} = \{ s \mid s = (p, v, R) \in \mathbb{R}^3 \times \mathbb{R}^3 \times \text{SO}(3) \}. \]
        
        \item \textbf{Action Space (\(\mathcal{A}\))}\\
        Defined as the set of computed thrust and angular velocity control variables for the UAV:
        \[ \mathcal{A} = \{ a \mid a = (F, \omega) \in \mathbb{R} \times \mathbb{R}^3 \}. \]
        
        \item \textbf{Plant (\(\mathcal{P}\))}\\
        Driven by the multi-rotor rigid body dynamics equation \eqref{eq:uav_dynamics}, describing the evolution of the state with the control inputs:
        \[ \dot{s} = f(s, a), \quad s \in \mathcal{S}, a \in \mathcal{A}. \]
        
        \item \textbf{Output Function (\(h\))}\\
        Defined as the pose extraction mapping based on \eqref{eq:hrr}. The output $y$ is the current body pose $R$:
        \[ y = h(s) \triangleq h_R(p,v, R) = R \in \text{SO}(3). \]
        
        \item \textbf{Desired State Set (\(\mathcal{Y}_{\mathrm{d}})\)}\\
        Consists of the desired interception pose $R_\text{d}$ for the task:
        \[  \mathcal{Y}_{\mathrm{d}} = \{ y_{\mathrm{d}} = R_\text{d} \in \text{SO}(3) \}. \]
        
        \item \textbf{Barrier Space (\(\mathcal{B}\))}\\
        Defined as the complement of the FOV safe set $\mathcal{C}_\text{fov}$ (i.e., the state region leading to target loss):
        \[ \mathcal{B} = \mathcal{S} \setminus \mathcal{C}_\text{fov} = \{ s \in \mathcal{S} \mid \alpha(s) \geq \delta_\text{fov} \}. \]
        {Safety invariance guarantee comes from the Equation \eqref{eq:cbf_condition}.}
        
        \item \textbf{Metric Function (\(d_\mathcal{Y}\))}\\
        Utilizing the orthogonality of rotation matrices, it measures the deviation between the current pose $R$ and the desired pose $R_\text{d}$. A distance metric on the manifold is typically employed, using the Frobenius norm:
        \[ d_\mathcal{Y}(y, \mathcal{Y}_{\mathrm{d}}) = \| I - y^\top y_{\mathrm{d}} \| = \| I - R^\top R_\text{d} \|_F. \]
        
        \item \textbf{Policy (\(\pi\))}\\
        The objective of policy $\pi$ is to design the control input $a$ using the state information $s$ and the desired feature $y_{\mathrm{d}}$. Under the strict premise of ensuring that the system state never enters the barrier space (i.e., $s(t) \notin \mathcal{B}$), it aims to drive the output $y$ to asymptotically converge to the desired set $\mathcal{Y}_{\mathrm{d}}$, while ensuring that $s$ remains bounded and $s \in \mathcal{C}_\text{fov}$:
        \[ \lim_{t \to \infty} d_\mathcal{Y}(y(t), \mathcal{Y}_{\mathrm{d}}) = 0, \quad \text{s.t.} \quad s(t) \in \mathcal{C}_\text{fov} \text{ and } s \in \mathcal{L}_\infty, \forall t \geq 0. \]
    \end{itemize}
\end{highlightbox}

\subsection{The Chess Game Problem}

\begin{figure}[H]
  \centering
  \includegraphics[width=1\textwidth]{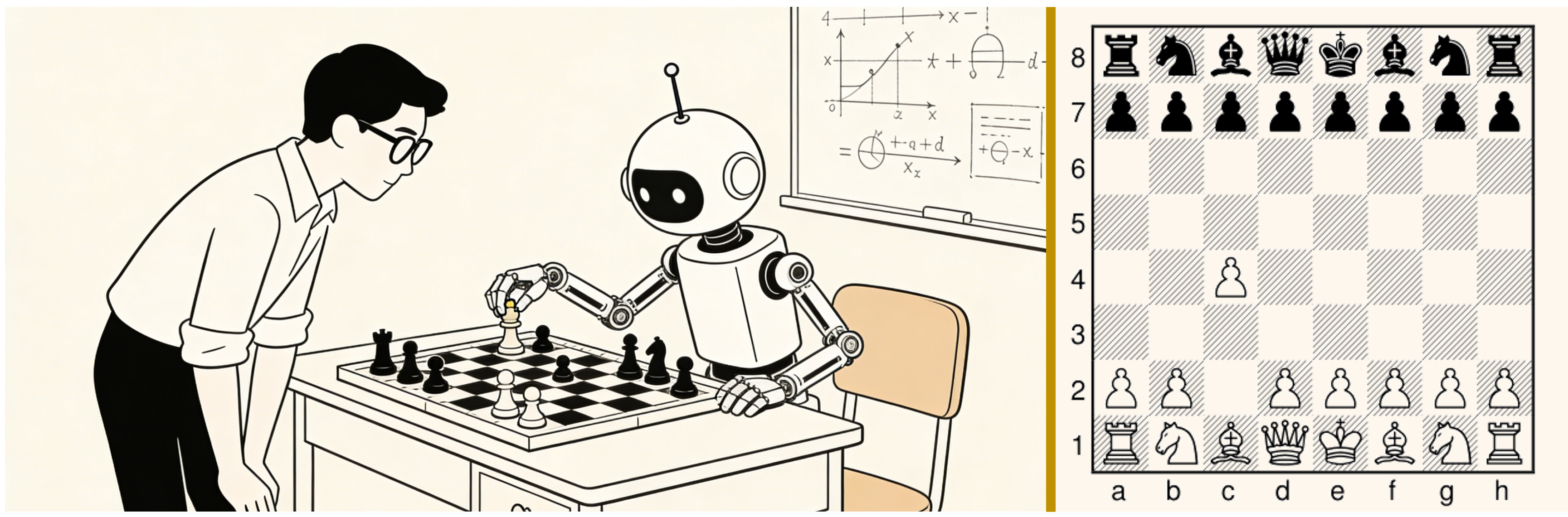}
  \caption{The chess game problem}
  \label{fig:chess}
\end{figure}

\textbf{Problem Description}: Consider the game problem of a player/agent in a chess match. In this game, both sides control black and white pieces on an \( 8 \times 8 \) chessboard. Each side has 16 pieces divided into 6 types: Pawn (P), Knight (N), Bishop (B), Rook (R), Queen (Q), and King (K). Figure \ref{fig:chess} provides a schematic of the overall game. The movement of the pieces must comply with the rules of chess. The game is won when one side checkmates the opponent's King; if neither side is in check and no legal moves are available, the game is declared a draw (stalemate).

In the context of stabilization learning, the player/agent needs to formulate a policy \( \pi(a|s) \) (which correlates with the current game state \(s\)) to ensure that executing moves according to this policy maximizes the probability of winning or drawing, while minimizing the chances of losing. The challenge of this task lies in the fact that, as one side of the game, the player/agent can only control its own pieces, while the evolution of the overall game system, compounded by the opponent's behavior, is highly unpredictable. Essentially, chess is a combinatorial game characterized by perfect information, zero-sum, and non-cooperation. The state space of chess is approximately \( 10^{40} \), and { the game-tree complexity reaches the order of $10^{120}$, often referred to as the Shannon number}; therefore, traditional exhaustive search or hand-crafted feature methods are largely inapplicable.

To address this problem, it is first necessary to establish a model \cite{silver2018general} that describes the real-time state of the chess game, which can be encoded as an \( N \times N \times (MT+L) \) state tensor \(s\):
\begin{itemize}[leftmargin=*, itemsep=2pt]
\item \( N = 8 \): Represents the size of the chessboard, with the 8 squares in horizontal and vertical directions numbered \( \{a,b,c,d,e,f,g,h\} \) and \( \{1,2,3,4,5,6,7,8\} \), respectively.
\item \( M = 6+6 = 12 \): Represents the feature planes used to describe the distribution of pieces on the board. Each feature plane provides the spatial distribution of a specific type of piece. The tensor element \( s_{i,j,\text{Plane}} = 0/1, i,j = 1,2,\ldots,8 \) indicates whether a piece of that type exists at position \( (i,j) \), where \( s = 0 \) means absent and \(s = 1\) means present. The 6 types of pieces for both our side and the opponent (P, N, B, R, Q, K) each occupy one plane. For example, our King's feature plane can be denoted as the ``King Plane", and the opponent's Queen's plane as the ``Opponent Queen Plane".
\item \( T \): Represents historical time steps, incorporating the board states of the previous \( T \) time steps including the current one.
\item \( L \): Represents constant feature planes used to denote information such as total steps and special rule states.
\end{itemize}

In chess, piece movements are discrete, finite, and strictly governed by rules. A legal action can be encoded as an \( N \times N \times K \) policy tensor \( a \):
\begin{itemize}[leftmargin=*, itemsep=2pt]
\item \( N = 8 \): Represents the size of the chessboard.
\item \( K \): Represents the possible action types of a piece, including movement directions\allowbreak \( \{\text{N, NE, E,}\allowbreak \text{SE, S, SW, W, NW}\} \), movement distances \( \{1,2,3,4,5,6,7\} \), and possible underpromotions.
\end{itemize}

Considering that the strategy the opponent might adopt is unknown and stochastic, a decision function \( g(s) \) is employed to abstract the possible actions the opponent might take under the game state \( s \) (essentially describing the probability distribution of the opponent's various next moves). Within a single round (where both our side and the opponent each make one move), the evolution of the game state can be described as follows: in state \( s_t \), our side executes action \( a_t \), yielding an intermediate state \( s_t^\prime \); the opponent then executes action \( a_t^\prime = g(s_t^\prime) \) under \( s_t^\prime \), ultimately resulting in the next state \( s_{t+1} \). This series of operations represents a Markov process, expressed as:
\begin{align}
g(s) &:= \mathbb{P}(a|s), \\
s_t \xrightarrow{a_t} s_t^\prime  & \xrightarrow{a_t^\prime = g(s_t^\prime)} s_{t+1}.
\end{align}

To ensure that the policy adopted by the player/agent guarantees stable convergence toward winning, a deep neural network \( f_\theta(s) \) trained via self-play learning is typically used to assist decision-making. Upon inputting the game state \( s \) into the network, it returns a move probability vector \( p \) and a scalar value \( v \). The vector \( p \) quantifies the probability \( \mathbb{P}(a|s) \) of executing an action under the current situation, while the scalar \( v = \mathbb{V}(s) \in [-1,1] \) represents the evaluation network \( \mathbb{V} \)'s win-rate estimation for the current state \( s \), where \( z = -1 \) denotes a loss, \( z = 1 \) denotes a win, and \( z = 0 \) denotes a draw:
\begin{align}
(p, v) &= f_\theta(s), \\
p &= \left\{p_a \mid p_a = \mathbb{P}(a|s) \in [0,1], a \in \mathcal{A}, \sum p_a = 1 \right\}.
\end{align}

Furthermore, the Monte Carlo Tree Search (MCTS) algorithm is utilized. Each search comprises a series of simulated self-play games, traversing from the root node \( s_\text{root} \) to a leaf node \( s_\text{leaf} \). Each simulation selects a move \( a \) at each state \( s \) that features a low visit count, high move probability, and high value (based on the current neural network \( f_\theta \)'s average evaluation of the simulated leaf node states resulting from choosing \( a \) at \( s \)).

\begin{highlightbox}{The Chess Game Problem}{chess-game}
  Transforming the chess game problem into a six-tuple $(\mathcal{S}, \mathcal{A}, \mathcal{P}, \pi, \mathcal{S}_{\mathrm{d}}, d)$ under the stabilization learning framework yields the following.
  \begin{itemize}[leftmargin=*, itemsep=2pt]
  \item \textbf{State Space ($\mathcal{S}$) and State ($s$)}\\
    Defined as the set of all possible game states. The positions of each type of piece are characterized by corresponding feature planes, and the combination of these planes constitutes the game state:
    \[ \mathcal{S} = \{s \mid s\in \mathbb{R}^{N\times N\times (MT+L)}, s_{i,j,k} \in \{0,1\} \} . \]
  \item \textbf{Action Space ($\mathcal{A}$)}\\
    Defined as the set of all legal moves. The size of the action space is determined by the game state:
    \[ \mathcal{A} = \{a \mid a \in \mathbb{R}^{N\times N\times K}\}. \]
  \item \textbf{Plant ($\mathcal{P}$)}\\
    The evolution of the plant's state is driven by two parts: our deterministic action \( a_t \) and the opponent's probabilistic random action \( a_t^\prime = g(s_t^\prime) \). The state transitions from \( s_t \) to \( s_{t+1} \) in one round:
    \[
    \begin{aligned}
    g(s) &:= \mathbb{P}(a|s), \\
    s_t \xrightarrow{a_t} s_t^\prime  & \xrightarrow{a_t^\prime = g(s_t^\prime)} s_{t+1}.
    \end{aligned}
    \]
  \item \textbf{Desired State Set ($ \mathcal{S}_{\mathrm{d}}$)}\\
    {Defined as the set of all game states where our side wins, characterized by our King surviving while the opponent's King is checkmated:
    \[ 
\mathcal{S}_{\mathrm{d}} = \left\{s_{\mathrm{d}} \,\middle|\, \begin{aligned}
    &s_{\mathrm{d}} \xrightarrow{a} s^\prime,\, s^\prime_{i^\ast, j^\ast, M^\prime} = {s_{\mathrm{d}}}_{i^\ast,j^\ast, \text{Opponent King Plane}}=1 , \exists \,a \in \mathcal{A}; \\ 
    &{s_{\mathrm{d}}}_{i,j,\text{King Plane}} = 1, \exists\, i,j \in \{1,2,\ldots,8\}
\end{aligned} \right\},
\]where $(i^\ast, j^\ast)$ denotes the spatial coordinate of the opponent's King in state $ s_{\mathrm{d}}$, and $M^\prime$ represents the specific feature plane of our piece that executes the capture transition. Under this formulation, a state is classified into the desired set if our King remains active on the board, and there exists a feasible action capable of driving the system into a successor state where our piece occupies the opponent's King position, thereby finalizing the stabilization objective.}

  \item \textbf{Metric Function ($d$)}\\
    Measures the win rate of the current game state evolving to the end game. Typically, the evaluation network \( \mathbb{V} \) maps all states to a value function \( v = \mathbb{V}(s) \in [-1,1] \), where \(v = -1\) means loss, \(v = 1\) means win, and \(v = 0\) means draw:
    \[  d(s,\mathcal{S}_{\mathrm{d}}) = 1 - \mathbb{V}(s). \]
  \item \textbf{Policy ($\pi$)}\\
    The objective of the policy is to provide the move \(a_t\) that has the highest probability of winning based on the state tensor \(s_t\). By providing the maximum-value action \(a_t\), it aims to drive the board situation \(s\) towards a winning end-game result \(\mathcal{S}_{\mathrm{d}}\) over multiple rounds, i.e., as \( t \to \infty \), \( d(s_t,\mathcal{S}_{\mathrm{d}}) \to 0 \).
  \end{itemize}
\end{highlightbox}

\subsection{Interaction with the Environment: Taking the Push-T Task as an Example} \label{sec:push-t}

\begin{figure}[H]
    \centering
    \includegraphics[width=0.8\textwidth]{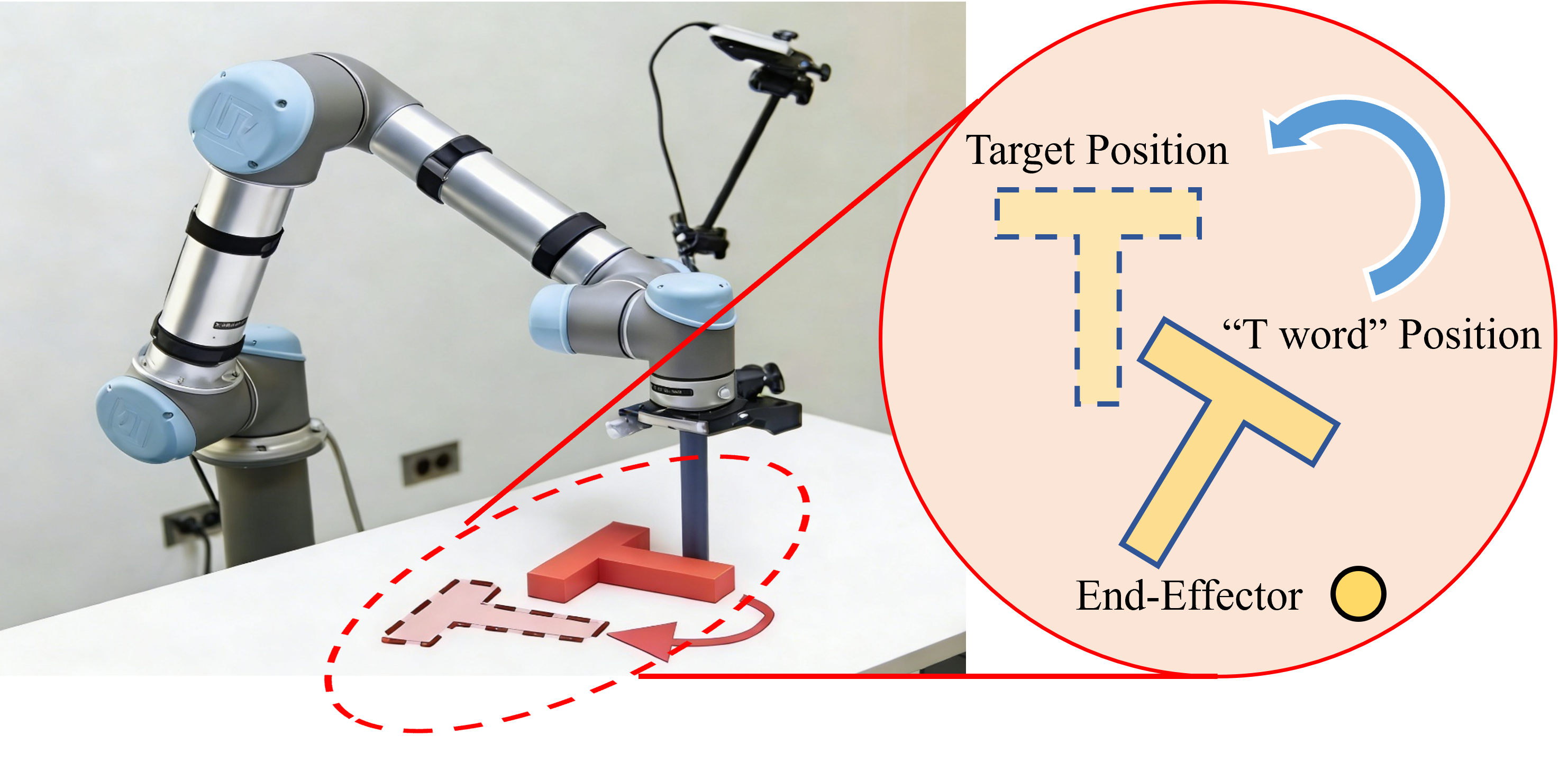}
    \caption{Schematic diagram of the Push-T task. The robot uses its end-effector to push a T-shaped object so that the object's pose converges to a target pose(dashed box) while respecting contact and collision constraints.}
    \label{fig:push-t-task}
\end{figure}

\textbf{Problem Description:}
Consider the Push-T task \cite{chi2023diffusion}, a typical non-prehensile robotic manipulation problem. It requires the robot to exert a contact pushing force on a T-shaped block via its end-effector, moving it across a plane to precisely reach a predefined target pose, as shown in Figure \ref{fig:push-t-task}. This task simulates common object-pushing operations in scenarios like industrial assembly and logistics sorting. Its core challenge lies in the complexity of contact dynamics: the robot must control the object through indirect contact rather than direct grasping, involving non-holonomic constraints, friction cone constraints, and contact state transitions.

Specifically, the task objective is to make the center-of-mass position $p_T = [x_T\quad y_T]^\top$ and the orientation angle $\theta_T$ of the T-shaped block simultaneously converge within a neighborhood of the target values $p_T^{*}$ and $\theta_T^{*}$, and ultimately bring the system to rest. This target means we need to design a control policy such that within a finite time, the state satisfies:
\begin{equation}
\|p_T - p_T^{*}\| \leq \varepsilon_p, \quad \|\theta_T - \theta_T^{*}\| \leq \varepsilon_\theta, \quad v_T = 0, \quad \omega_T = 0,
\label{eq:success_criteria}
\end{equation}
where $\varepsilon_p, \varepsilon_\theta$ are predefined allowable error thresholds. This criterion implies the control policy must not only achieve precise pose control but also ensure the system stabilizes near the target point to avoid persistent oscillation.

To mathematically describe this control problem, we define the full system state $s \in \mathcal{S} \subseteq \mathbb{R}^{10}$, which includes the pose and velocity information of the manipulated object (T-block) and the End-Effector:
\begin{align}
s &= [s_T^\top \quad s_{ee}^\top]^\top, \\
s_T &= [p_T^\top\quad v_T^\top\quad \theta_T\quad \omega_T]^\top, \\
s_{ee} &= [p_{ee}^\top\quad v_{ee}^\top]^\top.
\label{eq:state_vector_align}
\end{align}
Here, $p_T, v_T \in \mathbb{R}^2$ are the center-of-mass position and linear velocity of the T-block; $\theta_T, \omega_T \in \mathbb{R}$ are its orientation angle and angular velocity; $p_{ee}, v_{ee} \in \mathbb{R}^2$ are the position and velocity of the end-effector's contact point in the plane. The state space has a dimensionality of 10, encompassing the complete kinematic information of the system.

The control input is the velocity command $v_{\text{cmd}} \in \mathcal{A} \subseteq \mathbb{R}^2$ of the end-effector. In real physical systems, the actuator has inertia, and its dynamics can be modeled as a first-order lag system:
\begin{equation}
\dot{v}_{ee} = -\frac{1}{\tau}(v_{ee} - v_{\text{cmd}}),
\label{eq:actuator_dynamics}
\end{equation}
where $\tau > 0$ is the time constant reflecting the actuator's response speed. Under high-bandwidth control or ideal simulation conditions, this equation is often simplified to $v_{ee} \approx v_{\text{cmd}}$.

The dynamics of the T-block are governed by contact mechanics. When the end-effector pushes the T-block, the magnitude and direction of the contact force are subjected to friction cone constraints, closely relating to the relative position of the contact point, the object's geometry, and inertial parameters. This contact dynamic can be generally described as:
\begin{align}
\dot{s}_{T} &= f_T(s_T, s_{ee}) = 
\left[\begin{array}{c}
v_{T} \\ a_{T}(s_T, s_{ee}) \\ \omega_{T} \\ \alpha_{T}(s_T, s_{ee})
\end{array}\right],
\label{eq:system_dynamics}
\end{align}
where $a_T$ and $\alpha_T$ are the translational and angular accelerations of the T-block's center of mass, representing the kinematic tendencies induced by the end-effector's contact. For the end-effector, the dynamical system can be expressed as:
\begin{align}
\dot{s}_{ee} &= f_{ee}(v_{\text{cmd}})=
\left[\begin{array}{c}
v_{ee} \\ -\frac{1}{\tau}(v_{ee} - v_{\text{cmd}})
\end{array}\right].
\label{eq:ee_dynamics}
\end{align}

\begin{highlightbox}{The Push-T Task}{push-t-task}
Formulating the Push-T task into a six-tuple $(\mathcal{S}, \mathcal{A}, \mathcal{P}, \pi, \mathcal{S}_{\mathrm{d}}, d)$ under the stabilization learning framework.
\begin{itemize}[leftmargin=*, itemsep=2pt]
    \item \textbf{State Space ($\mathcal{S}$) and State (\( s\)) } \\
    Defined as a 10-dimensional vector space containing the poses and velocities of both the T-block and the end-effector:
    \[ \mathcal{S} = \{ s \mid s = [p_T^\top \quad v_T^\top \quad \theta_T \quad \omega_T \quad p_{ee}^\top \quad v_{ee}^\top]^\top \in \mathbb{R}^{10} \}. \]
    
    \item \textbf{Action Space ($\mathcal{A}$) } \\
    Defined as the velocity command space of the end-effector:
    \[ \mathcal{A} = \{a \mid a = v_{\text{cmd}} \in \mathbb{R}^2\}. \]
    
    \item \textbf{Plant ($\mathcal{P}$)} \\
    The dynamic evolution of the plant is jointly described by the actuator dynamics \eqref{eq:ee_dynamics} and the T-block's contact dynamics \eqref{eq:system_dynamics}. The closed-loop dynamic equation is:
    \[ \dot{s} = \left[\begin{array}{c} f_T(s_T, s_{ee}) \\ f_{ee}(v_{\text{cmd}}) \end{array}\right]. \]
    
    \item \textbf{Desired State Set ($\mathcal{S}_{\mathrm{d}}$) } \\
    Defined as the set of exact system states that satisfy the joint constraints of position, angle, and velocity for task success:
    \[  \mathcal{S}_{\mathrm{d}} = \{ s_{\mathrm{d}} \in \mathcal{S} \mid p_T = p_T^{*}, \theta_T = \theta_T^{*}, v_T=0, \omega_T=0 \}. \]   
    
    \item \textbf{Metric Function ($d$)} \\
    A positive scalar function $d: \mathcal{S} \to \mathbb{R}_{\geq 0}$ measuring the distance between the current state and the target set. An Euclidean-based approach can be adopted:
    \[  d(s_{\mathrm{d}},\mathcal{S}_{\mathrm{d}}) = \left\| \begin{bmatrix} p_T - p_T^* \\ \theta_T - \theta_T^* \\ v_T \\ \omega_T \end{bmatrix} \right\|. \]
   
   \item \textbf{Policy ($\pi$)}\\
    The objective of policy \( \pi \) (control law) is to adjust the end-effector's velocity based on current observations. It aims to drive the system state into the target set $\mathcal{S}_{\mathrm{d}}$ under model uncertainty and contact nonlinearities by modulating the end-effector velocity, i.e., as \( t \to \infty \), $d(s(t),\mathcal{S}_{\mathrm{d}}) \to 0$.
\end{itemize}
\end{highlightbox}

\subsection{Stability Problems in Numerical Computation}

\begin{figure}[H]
  \centering
  \includegraphics[width=0.5\textwidth]{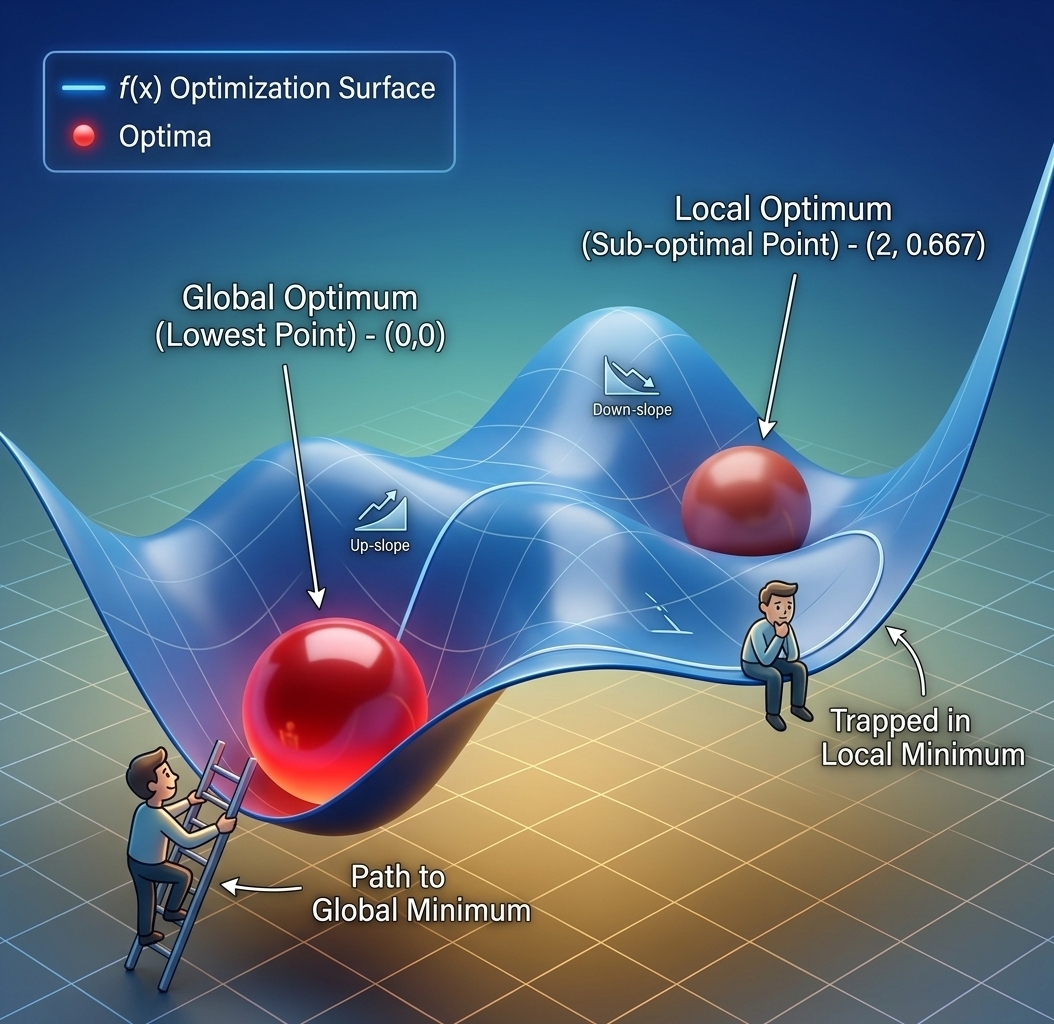}
  \caption{Schematic diagram of finding the kernel of a non-convex function}
  \label{fig:imgboth_num}
\end{figure}

\textbf{Problem Description}: Consider using numerical methods to solve for the roots $\text{Ker}[\,f\,]$ of a complex equation, where
\[ \text{Ker}[\,f\,] = \{s \mid f(s)=0\}, \quad f:\mathbb{R}^n \to \mathbb{R}^m. \]

Gradient descent is an effective optimization method capable of finding the unique minimum of a scalar convex function. However, conventional scalar functions typically possess multiple local minima (as shown in Figure \ref{fig:imgboth_num}). Even if the general equation-solving problem is reformulated into the aforementioned optimization problem, there typically exist no closed-form analytical solutions for such general non-convex problems. Numerically solving an equation is a process of continuously iterating the current state to converge toward a target objective, and the evolution of the state can be formulated using a discrete dynamical system as follows:
\begin{equation} \label{eq:numerical_method}
  s_{t+1} = s_t + \pi(f(s_t), s_t),
\end{equation}
where $\pi$ satisfies:
\begin{equation} \label{eq:pi_condition}
  \pi(f(s_{\mathrm{d}}),s_{\mathrm{d}})  = 0\Leftrightarrow s_{\mathrm{d}} \in \text{Ker}[\,f\,].
\end{equation}
Thus, if Equation \eqref{eq:numerical_method} converges, then
\[ \lim_{t\to \infty} \pi(f(s_t), s_t) = 0. \]
This limit achieves the goal of stabilizing the dynamical system, thereby accomplishing the root-finding for the function.

There are currently numerical computation algorithms designed to guarantee stable convergence, such as Runge-Kutta methods and Adams methods (classical numerical analysis methods for solving differential equations) \cite{hairer1993solving}. These aim to enhance the stability and convergence of numerical solution systems, ensuring that the solving results still stabilize towards the true solution even when initial conditions are polluted, step sizes are sub-optimal, or rounding errors exist during the process { under suitable step-size and stability-region conditions}.

\begin{highlightbox}{Stability Problems in Numerical Computation}{numerical-stability}
  Transforming the numerical computation problem into a six-tuple $(\mathcal{S}, \mathcal{A}, \mathcal{P}, \pi, \mathcal{S}_{\mathrm{d}}, d)$ under the stabilization learning framework.
  \begin{itemize}[leftmargin=*, itemsep=2pt]
  \item \textbf{State Space ($\mathcal{S}$) and State ($s$)}\\
    Defined as the natural domain reachable by the iterative solution vector under unconstrained conditions:
    \[ \mathcal{S} = \{s \mid s\in \mathbb{R}^n\}. \]
  \item \textbf{Action Space ($\mathcal{A}$)}\\
    Defined as the difference between the iterative solution vector $s_t$ at the current step and $s_{t+1}$ at the next step:
    \[ \mathcal{A} = \{a \mid a = \pi(f(s),s) \in \mathbb{R}^n\}. \]
 \item \textbf{Plant ($\mathcal{P}$)}\\
    The iterative process of the plant's solution vector is described by a difference equation:
    \[ s_{t+1} = s_t + a_t. \]
  \item \textbf{Desired State Set ($\mathcal{S}_{\mathrm{d}}$)}\\
    Defined as the set of true solution vectors of the original equation:
    \[  \mathcal{S}_{\mathrm{d}} = \{s_{\mathrm{d}} \mid f(s_{\mathrm{d}})=0\}. \]
  \item \textbf{Metric Function ($d$)}\\
    Measures the distance between the current state (solution vector) and the desired state set, or the distance after mapping the vector to the set $f(\mathcal{S})$, defined by a norm:
    \[  d(s_{\mathrm{d}},\mathcal{S}_{\mathrm{d}}) = \left\| f(s_{\mathrm{d}}) \right\|. \] 
    { Since the policy $\pi$ adopted during the iterative process satisfies \eqref{eq:pi_condition}, this metric function is equivalent to the gradual convergence of the iterated vector over successive steps, i.e.,
    \[ \lim_{t\to \infty} \left\| s_{t+1} - s_t \right\| = 0. \]}

  \item \textbf{Policy ($\pi$)}\\
    The objective of the policy is to adjust the difference vector for the next iteration step based on the current state solution vector. By flexibly adjusting the difference vector \( a = \pi(f(s_t),s_t) \), it aims to avoid falling into local optima as iterations increase, gradually approaching the true solution vector. That is, as \( t\to \infty \), \( d(s,\mathcal{S}_{\mathrm{d}}) \to 0 \).
  \end{itemize}
\end{highlightbox}

\subsection{Relative Pose Estimation of Target UAV Based on Visual Sensors} \label{sec:RelativePose}

\begin{figure}[H]
  \centering
  \includegraphics[width=0.7\textwidth]{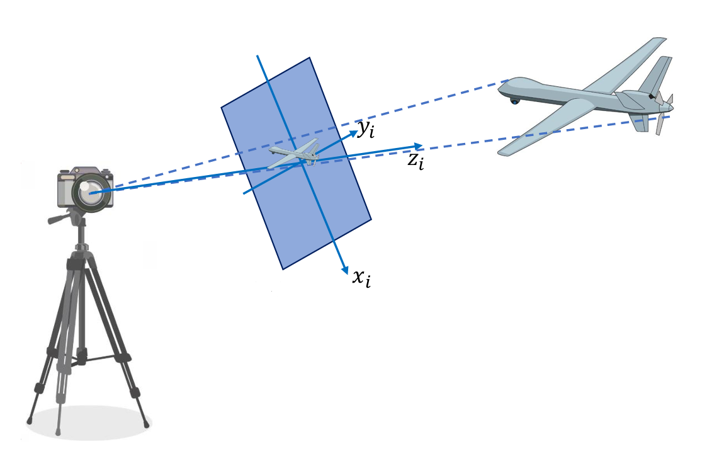}
  \caption{Schematic diagram of relative pose estimation of a target UAV based on visual sensors}
  \label{fig:RelativePose_img}
\end{figure}

\textbf{Problem Description}: Consider observing a fixed-wing UAV flying in the air using a visual sensor mounted on the ground. By utilizing the image sequence \( I \in \mathbb{R}^{C\times H\times W} \) (\( C, H \), and \( W \) being the number of channels, height, and width of the image respectively) acquired by the sensor, the state quantities of the observed UAV relative to the visual sensor are estimated. The actual scenario is shown in Figure \ref{fig:RelativePose_img}. Let $s_x^\prime=[p^\top\quad v^\top\quad \Theta^\top\quad \omega^\top]^\top$ where $\dot{s}_x^\prime=f_x^\prime(s_x^\prime,u)$, readers are referred to the literature \cite{beard2012small}. For simplicity of analysis, assuming the fixed-wing UAV is engaged in steady-state motion, we assume:
\[ \dot{u}=0. \]
Then the system can be transformed into an autonomous system:
\begin{equation}
    \dot{s}_x = f_x(s_x),
\end{equation}
where \( s_x = [p^\top\quad v^\top\quad \Theta^\top\quad \omega^\top\quad u^\top]^\top \), with specific components as follows:
\begin{itemize}[leftmargin=*, itemsep=2pt]
    \item Position state component \(p = [x\quad y\quad z]^\top \in \mathbb{R}^3\);
    \item Linear velocity state component \(v = [v_x\quad v_y\quad v_z]^\top \in \mathbb{R}^3\);
    \item Attitude state component \( \Theta = [\phi \quad\theta\quad\psi]^\top \in \mathbb{R}^3 \), representing roll angle \( \phi \in [-\pi,\pi] \), pitch angle \( \theta \in [-\frac{\pi}{2},\frac{\pi}{2}] \), and yaw angle \( \psi \in [-\pi,\pi] \);
    \item Angular velocity state component \(\omega = [\omega_x\quad\omega_y\quad\omega_z]^\top \in \mathbb{R}^3\);
    \item Input component \(u = [\delta_t\quad\delta_e\quad\delta_a\quad\delta_r]^\top \in \mathbb{R}^4\), corresponding to throttle thrust, elevator, aileron, and rudder respectively.
\end{itemize}

Furthermore, since directly using the visual sensor output $I$ in the observer would result in excessively high observation space dimensions, the onboard camera's observation process can be abstracted as a composite mapping. First, the camera imaging function $h_{\rm cam}(\cdot)$ maps the UAV pose to the image $I$; then the feature extraction function $\phi_I(\cdot)$ extracts low-dimensional structured feature information $y\in\mathcal{Y}\subset\mathbb{R}^p$ (such as feature point coordinates, line parameters, etc.) from the image, i.e.:
\begin{equation}
y = \phi_I(h_{\rm cam}(s_x,\mathcal{M})) \triangleq h(s_x),
\end{equation}
where $\mathcal{M}$ represents known prior information such as the 3D model of the target fixed-wing UAV and camera intrinsics.

The observer can be designed as:
\[
    \begin{aligned}
    \dot{\hat{s}}_x &= f_x(\hat{s}_x) + l(y, \hat{y}),\\
    \hat{y} &= h(\hat{s}_x),
    \end{aligned}
\]
where $l(\cdot)$ is the observer function describing the observer's processing and feedback of the observed images.

Defining the observation error as $\tilde{s}_x=s_x-\hat{s}_x$, the dynamical system model of the observation problem can be expressed as:
\begin{equation}
  \begin{aligned}
  \dot{\tilde{s}}_x &= f_x(s_x) - f_x(\hat{s}_x) - l(y,\hat{y}) \triangleq f(\tilde{s}_x,l(y,\hat{y})).
  \end{aligned}
  \label{eq:RelativePose_SystemModel}
\end{equation}
The goal of this problem is to ensure that the observer's estimation result $\hat{s}_x$ converges to the actual state $s_x$ by designing $l(\cdot)$, i.e.:
\[ \lim_{t\to\infty} \|\tilde{s}_x(t)\| = 0. \]

This technique plays an important role in fields such as airport safety supervision \cite{fan2021estimating}, UAV-assisted takeoff and landing/recovery \cite{falanga2017vision}, low-altitude airspace defense and countermeasures \cite{yang2025unified}, and swarm coordination/distributed measurement \cite{lu2024fast}.

\textbf{Note}: In this problem formulation, $\mathcal{M}$ represents the 3D model prior information of the target fixed-wing UAV, which is generally considered known. Since the monocular camera imaging process is usually modeled as a mapping from 3D Euclidean space $\mathbb{R}^3$ to 2D projective space $\mathbb{P}^2$, depth information is lost. Without the target's prior size information, recovering its 3D position is impossible—a phenomenon generally known as ``scale ambiguity". However, in this modeling, the dimensions of the target UAV are considered known, so the 3D position can be recovered utilizing camera intrinsics, making the problem observable. Under the stabilization learning framework, designing a reasonable policy $\pi$ guarantees the state converges to the desired state set.

\begin{highlightbox}{Relative Pose Estimation of Target UAV Based on Visual Sensors}{target-pose}
  Transforming this problem into a six-tuple $(\mathcal{S}, \mathcal{A}, \mathcal{P}, \pi, \mathcal{S}_{\mathrm{d}}, d)$ under the stabilization learning framework.
  \begin{itemize}[leftmargin=*, itemsep=2pt]
  \item \textbf{State Space ($\mathcal{S}$) and State ($s$)}\\
    Defined as the state requiring control during observation \( s = \tilde{s}_x \). The state space is:
    \[ \mathcal{S} = \{s \mid \tilde{s}_x \in \mathbb{R}^{16} \}. \]
  \item \textbf{Action Space ($\mathcal{A}$)}\\
    Defined as the system action \( a \), which is the control input of the system model (the observer gain $l(y,\hat{y})$):
    \[ \mathcal{A} = \{a \mid a  = l(y,\hat{y}) \in \mathbb{R}^{16}\}. \]
  \item \textbf{Desired State Set ($ \mathcal{S}_{\mathrm{d}}$)}\\
    Defined as the set where the estimation result equals the true state:
    \[ \mathcal{S}_{\mathrm{d}} = \{s_{\mathrm{d}} \mid s_{\mathrm{d}} = 0 \}. \]
  \item \textbf{Plant ($\mathcal{P}$)}\\
    Driven by the evolution law of state $s$ under the action of control input $a$, $\dot{s}=f(s,a)$, which is Equation \eqref{eq:RelativePose_SystemModel}.
  \item \textbf{Metric Function ($d$)}\\
    Measures the distance between the current state $s$ and the desired state set $  \mathcal{S}_{\mathrm{d}}$ using a norm:
    \[ d(s, \mathcal{S}_{\mathrm{d}}) = \| s_{\mathrm{d}} \|. \]
  \item \textbf{Policy ($\pi$)}\\
    The function of policy \(\pi\) is to design the observer correction function \(l(y,\hat{y})\) based on observed features \(y\). It aims to ensure the estimated result \(\hat{s}_x\) converges to the true state \(s_x\) via robust observer design despite external disturbances, i.e., as \( t\to\infty \), \( d(s(t),\mathcal{S}_{\mathrm{d}}) \to 0\).
  \end{itemize}
\end{highlightbox}

\subsection{Ego-Motion Estimation of UAV Based on Visual Sensors}

\begin{figure}[H]
  \centering
  \includegraphics[width=0.6\textwidth]{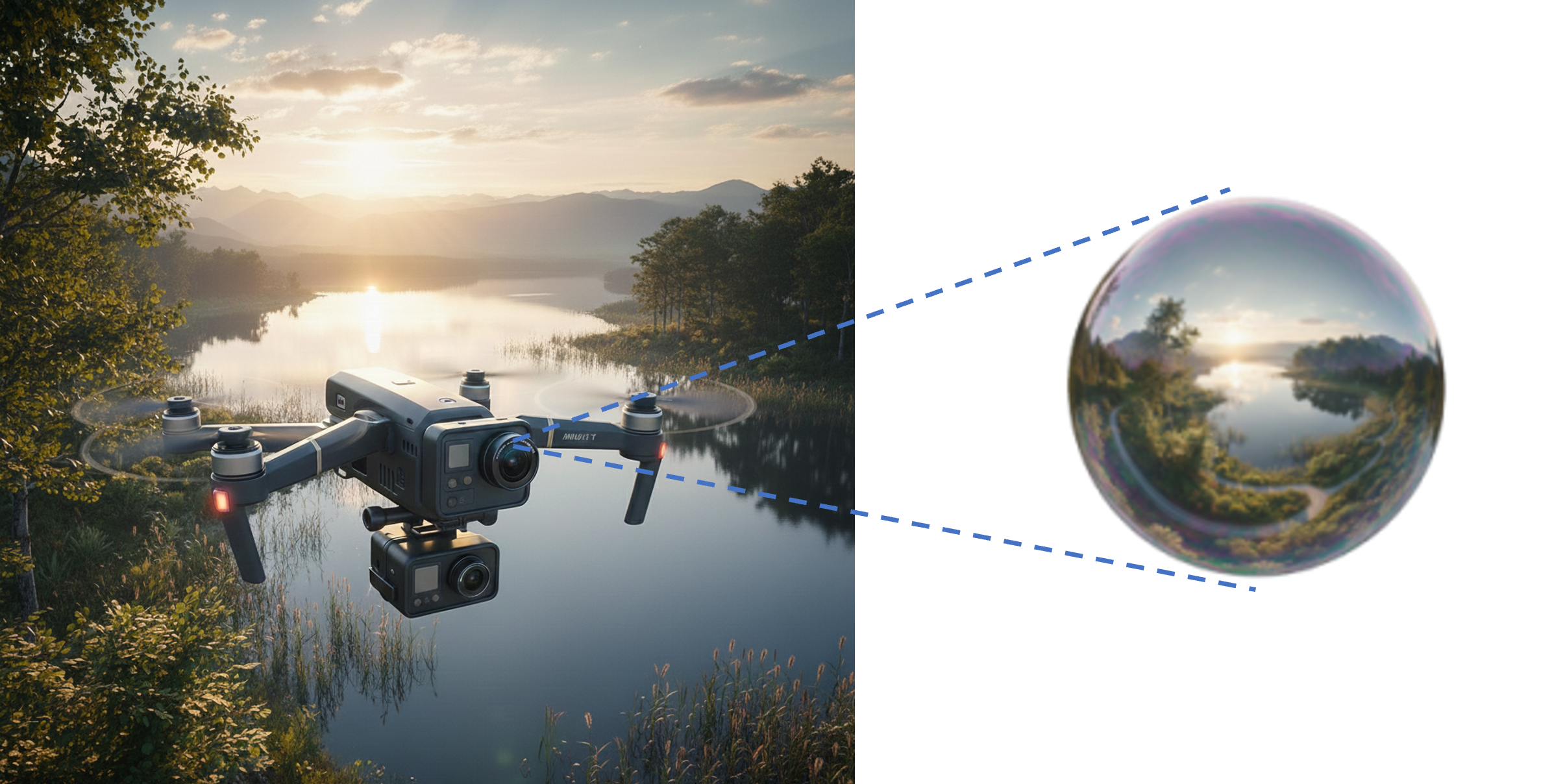}
  \caption{Schematic diagram of ego-motion estimation of UAV based on visual sensors}
  \label{fig:EgoMotion_img}
\end{figure}

\textbf{Problem Description}: Consider a multi-rotor UAV flying in the air, using the image sequence \( I \in \mathbb{R}^{C\times H\times W} \) (\( C, H \), and \( W \) being the number of channels, height, and width respectively) acquired by a visual sensor rigidly attached to it to estimate its own state \( s_x = [p^\top\quad v^\top\quad  \Theta^\top\quad  \omega^\top\quad  f\quad \tau^\top]^\top \)(the actual scenario is shown in Figure \ref{fig:EgoMotion_img}), where:
\begin{itemize}[leftmargin=*, itemsep=2pt]
    \item Position state component \(p = [x\quad y\quad z]^\top \in \mathbb{R}^3\);
    \item Linear velocity state component \(v = [v_x\quad v_y\quad v_z]^\top \in \mathbb{R}^3\);
    \item Attitude state component \( \Theta = [\phi\quad \theta\quad \psi]^\top \in \mathbb{R}^3 \). Roll angle \( \phi \in [-\pi,\pi] \), pitch angle \( \theta \in [-\frac{\pi}{2},\frac{\pi}{2}] \), and yaw angle \( \psi \in [-\pi,\pi] \);
    \item Angular velocity state component \(\omega = [\omega_\phi\quad \omega_\theta\quad \omega_\psi]^\top \in \mathbb{R}^3\);
    \item Thrust state component \(f \in \mathbb{R}\);
    \item Torque state component \(\tau = [\tau_x\quad  \tau_y\quad  \tau_z]^\top \in \mathbb{R}^3\).
\end{itemize}

The model of a quadrotor UAV can be expressed as \cite{quan2017introduction}:
\begin{equation}
\begin{cases}
  \dot{p} &= v \\ 
  \dot{v} &= ge_3 - \frac{fRe_3}{m} \\ 
  \dot{R} &= R[\omega]_\times \\ 
  J\dot{\omega} &= -\omega\times(J\omega) + \tau
\end{cases}
\end{equation}
where $g$ is the gravitational acceleration, $e_3=[0\quad 0\quad 1]^\top$, $m$ is mass, and $[\cdot]_\times$ denotes the skew-symmetric matrix. In practice, $f$ and $\tau$ are hard to obtain directly. To simplify the analysis, assuming the quadrotor is in steady-state motion:
\[
  \begin{aligned}
  \dot{f} &= 0,\\
  \dot{\tau} &= 0,
  \end{aligned}
\]
then the state model can be denoted as:
\[ \dot{s}_x = f_x(s_x). \]

Furthermore, directly using visual output $I$ in the observer causes high dimensions. The onboard camera's observation is abstracted as a composite mapping. The imaging function $h_{\rm cam}(\cdot)$ maps pose to $I$, and feature extraction $\phi_I(\cdot)$ extracts low-dimensional features $y\in\mathcal{Y}\subset\mathbb{R}^p$:
\begin{equation}
y_I = \phi_I(h_{\rm cam}(s_x,\mathcal{M})) \triangleq h^\prime(s_x, \mathcal{M}),
\end{equation}
where $\mathcal{M}$ represents prior scene information and camera intrinsics, which might be unknown. However, relying purely on vision in unknown scenes makes it difficult to ensure scale observability due to scale ambiguity \cite{hesch2014consistency}. Thus, a barometer measurement of altitude is incorporated to enhance observability:
\[ y_z = z. \]
The system output becomes:
\begin{equation}
  y = \begin{bmatrix} y_I \\ y_z \end{bmatrix} = \begin{bmatrix} h^\prime(s_x, \mathcal{M}) \\ z \end{bmatrix} \triangleq h(s_x,\mathcal{M}).
\end{equation}

The observer can be designed as:
\[
    \begin{aligned}
    \dot{\hat{s}}_x &= f_x(\hat{s}_x) + l(y, \hat{y}),\\
    \hat{y} &= h(\hat{s}_x,\mathcal{M}),
    \end{aligned}
\]
where $l(\cdot)$ is the observer correction function.

Defining observation error $\tilde{s}_x=s_x-\hat{s}_x$, the dynamic model is:
\begin{equation}
  \begin{aligned}
  \dot{\tilde{s}}_x &= f_x(s_x) - \hat{f}_x(\hat{s}_x) - l(y,\hat{y}) \triangleq f(\tilde{s}_x,l(y,\hat{y})).
  \end{aligned}
  \label{eq:EgoMotion_SystemModel}
\end{equation}
The goal is to design $l(\cdot)$ so that $\hat{s}_x$ converges to $s_x$:
\[ \lim_{t\to\infty} \|\tilde{s}_x(t)\| = 0. \]
This technique is commonly applied in GNSS-denied environments like indoors, forests, or urban canyons, where the onboard camera calculates real-time pose using pixel info or projective geometry \cite{murartal2017orbslam2, engel2018direct, forster2017svo}.

\textbf{Note}: { Distinct from Section \ref{sec:RelativePose}, $\mathcal{M}$ in this problem denotes the prior information of the scene, which is typically unknown. Consequently, the system suffers from severe scale unobservability, meaning that it is insufficient to determine the physical distance of its ego-motion relying solely on visual information. In this problem, the altitude $y_z$ provided by a barometer is introduced as a metric reference to forcefully decouple the scale factor, thereby ensuring the convergence of the estimated state. Within the framework of Stabilization Learning, the designed policy $\pi$ guarantees the convergence of the state to the desired state set and eliminates the influence of the unknown $\mathcal{M}$.}

\begin{highlightbox}{Ego-Motion Estimation of UAV Based on Visual Sensors}{ego-motion}
  Transforming this problem into a six-tuple $(\mathcal{S}, \mathcal{A}, \mathcal{P}, \pi, \mathcal{S}_{\mathrm{d}}, d)$ under the stabilization learning framework.
  \begin{itemize}[leftmargin=*, itemsep=2pt]
  \item \textbf{State Space ($\mathcal{S}$) and State ($s$)}\\
    The required controlled state during observation is \( s=\tilde{s}_x \):
    \[ \mathcal{S} = \{s \mid s = [p^\top\quad v^\top\quad \Theta^\top\quad \omega^\top\quad u^\top]^\top \in \mathbb{R}^{16} \}. \]
  \item \textbf{Action Space ($\mathcal{A}$)}\\
    The system action \( a \) is the control input (observer gain $l(y,\hat{y})$):
    \[ \mathcal{A} = \{a \mid a = l(y,\hat{y}) \in \mathbb{R}^{16}\}. \]
  \item \textbf{Plant ($\mathcal{P}$)}\\
    Driven by the evolution law $\dot{s}=f(s,a)$, which is Equation \eqref{eq:EgoMotion_SystemModel}.
  \item \textbf{Desired State Set ($ \mathcal{S}_{\mathrm{d}}$)}\\
    Defined as the set where estimation equals truth:
    \[ \mathcal{S}_{\mathrm{d}} = \{s_{\mathrm{d}} \mid \| s_{\mathrm{d}} \| = 0\}. \]
  \item \textbf{Metric Function ($d$)}\\
    Measures distance using the 2-norm:
    \[ d(s, \mathcal{S}_{\mathrm{d}}) = \| s_{\mathrm{d}} \|. \]
  \item \textbf{Policy ($\pi$)}\\
    The policy $\pi$ designs the proper observer function \(l(y,\hat{y})\) based on visual features. It aims to guarantee state $s$ converges to $\mathcal{S}_{\mathrm{d}}$ via robust observer design, i.e., as \(t\to\infty\), \( d(s(t),\mathcal{S}_{\mathrm{d}}) \to 0\).
  \end{itemize}
\end{highlightbox}

\subsection{Image Generation Guided by Text Prompts}

\begin{figure}[H]
  \centering
  \includegraphics[width=0.98\textwidth]{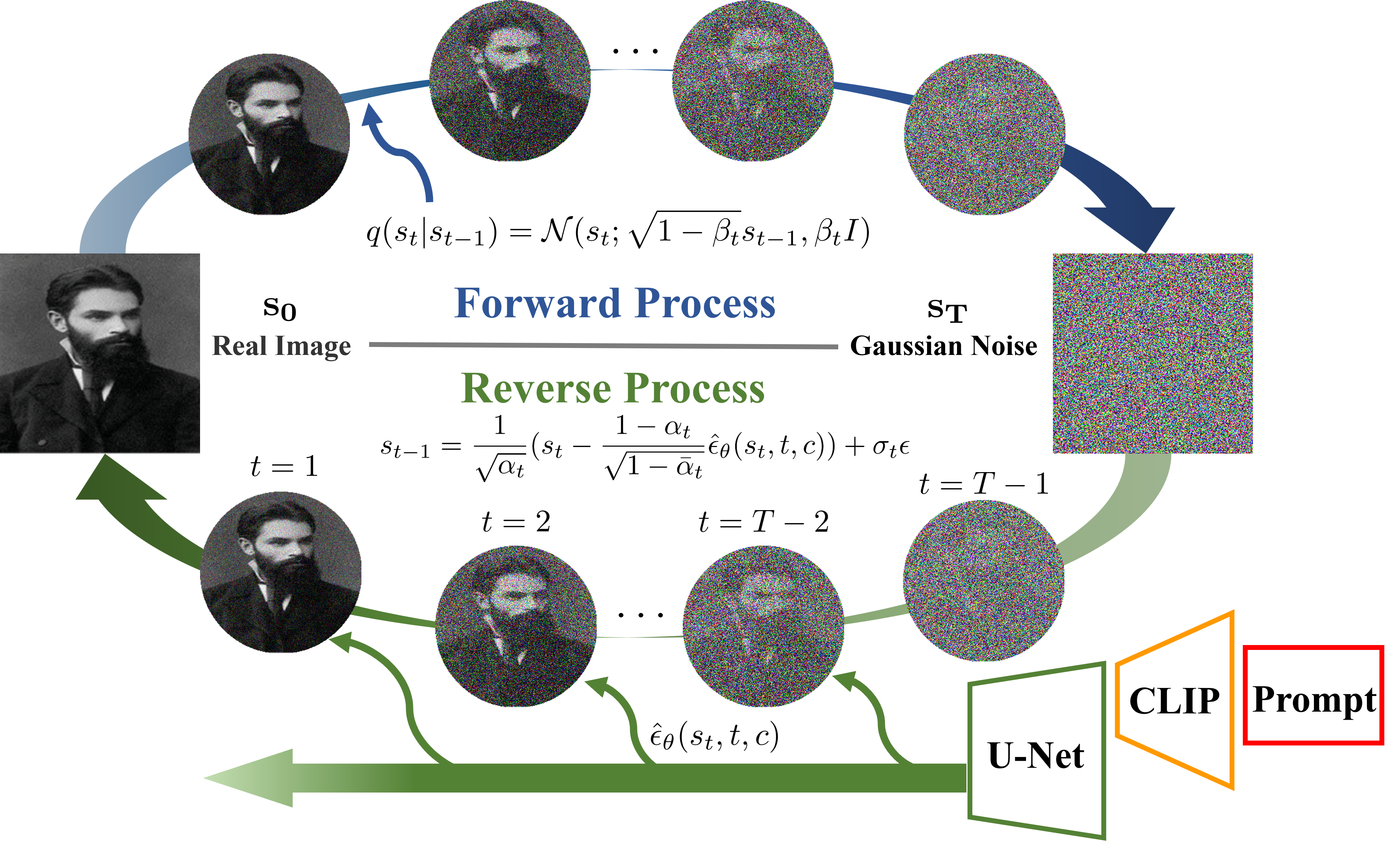}
  \caption{Schematic diagram of the forward diffusion process and the backward text-guided image generation process in diffusion models.}
  \label{fig:diffusion}
\end{figure}

\textbf{Problem Description}: Consider a text-to-image model that guides an agent to generate images conforming to text prompt requirements. Such models take natural language descriptions as inputs and rely on the step-by-step denoising generation mechanism of diffusion models. They iteratively reconstruct visual content from random noise in the latent space, continuously aligning with the text semantics during the generation process, and ultimately output images with reasonable structures, clear content, and high matching degrees with the input prompts, thereby realizing the automated transformation from text semantics to visual images. Diffusion models \cite{sohl2015deep,ho2020denoising} consist of two core mechanisms: the forward diffusion process and the reverse denoising process, as shown in Figure \ref{fig:diffusion}. The former is used to train the model, while the latter generates images based on prompt instructions.

In the forward diffusion process, given the latent space state $ s_0 $ of the initial image, a series of latent space states $ s_1, s_2, \cdots, s_T $ is generated by progressively adding Gaussian noise until the latent state approximates standard Gaussian noise. Here, $ T $ is the total number of diffusion steps, $ \beta_t $ is the predefined variance schedule for each step (typically varying linearly from $ \beta_1 = 10^{-4} $ to $ \beta_T = 0.02 $), and $ I $ is the identity matrix. The forward process as a whole constitutes a Markov chain:
\begin{align}
q(s_t | s_{t-1}) &= \mathcal{N}(s_t; \sqrt{1- \beta_t} s_{t-1}, \beta_t I), \\
q(s_T | s_0) &= \prod_{t = 1}^T q(s_t | s_{t-1}).
\end{align}
Through the reparameterization trick and the additivity of Gaussian distributions, one can directly sample the latent space state $ s_t $ at any time step $ t $ from the initial latent space state $ s_0 $. Defining the retention rate at each step as $ \alpha_t = 1 - \beta_t $:
\begin{align}
    \bar{\alpha}_t &= \prod_{i = 1}^t \alpha_i, \\
q(s_t | s_0) &= \mathcal{N}(s_t; \sqrt{\bar{\alpha}_t} s_0, (1 - \bar{\alpha}_t) I).
\end{align}

The reverse denoising process is essentially learning a denoising model of a Markov chain. By sampling from the initialized Gaussian noise $ s_T $ and progressively removing the noise, it restores the latent space state $ s_0 $ of the initial image:
\begin{align}
p(s_T) &= \mathcal{N}(s_T; 0, I), \\
p(s_{t-1} | s_t) &= \mathcal{N}(s_{t-1}; \mu_\theta(s_t), \Sigma_\theta(s_t)).
\end{align}
Here, $ \theta $ represents the parameters of the denoising model in the reverse process, $ \mu_\theta(s_t) $ is the output of the denoising model, {and $ \Sigma_\theta(s_t) $ may be a predefined variance schedule, a fixed matrix, or a learned variance, depending on the model configuration. }Typically, the denoising model does not predict $ \mu_\theta(s_t) $ directly; instead, it predicts the noise $ \hat{\epsilon}_\theta(s_t, t) $ and infers $ \mu_\theta(s_t) $ from it:
\begin{equation}
\mu_\theta(s_t) = \frac{1}{\sqrt{\alpha_t}}\left(s_t - \frac{1-\alpha_t}{\sqrt{1-\bar{\alpha}_t}}\hat{\epsilon}_\theta(s_t, t)\right).
\end{equation}

Text-to-image models generally utilize the reverse process of diffusion models and introduce conditional information (such as text prompt embeddings $ c $) to ensure the generated images satisfy the instruction requirements. First, the model samples an initial noise $ s_T \in \mathbb{R}^{H \times W \times C} $ satisfying a Gaussian distribution in the latent space (where $ H \times W \times C $ is the latent space scale). Then, through a conditional U-Net (parameterized by $ \theta $) and time step $ t $, it predicts the noise and denoises the current latent space state.

Specifically, the model converts the text prompt into a text embedding vector $ c $ via Contrastive Language-Image Pre-training (CLIP) \cite{radford2021learning}. At each time step $ t $, based on the current latent space state $ s_t $, time step $ t $, and conditional information $ c $, it predicts the noise $ \hat{\epsilon}_\theta(s_t, t, c) $ using the conditional U-Net. Subsequently, utilizing a scheduler (such as Denoising Diffusion Implicit Models (DDIM) or Diffusion Probabilistic Model Solver (DPM-Solver)), it infers the latent space state $ s_{t-1} $ for the previous time step based on the noise prediction:
\begin{equation} \label{eq:denoise}
s_{t-1} = \frac{1}{\sqrt{\alpha_t}}\left(s_t - \frac{1-\alpha_t}{\sqrt{1-\bar{\alpha}_t}}\hat{\epsilon}_\theta(s_t, t, c)\right)+\sigma_t\epsilon,
\end{equation}
where $ \sigma_t $ is the noise scale controlling the randomness of image generation, and $ \epsilon $ is random noise sampled from a standard Gaussian distribution. By iterating the above process ($ t \to 0 $), the final latent space state $ s_0 $ satisfying the condition is obtained, which is then decoded into a pixel-space image by a Variational Autoencoder (VAE) \cite{kingma2014auto} decoder.

To align the theoretical framework with the infinite-horizon asymptotic stability in stabilization learning, a time-mapping function $\tau = \varphi(t)$ can be formally introduced. For instance, one can define an infinite virtual time for the system evolution as $\tau = -\ln(t/T)$ (i.e., $t = T e^{-\tau}$). Consequently, as the time step of the denoising process $t \to 0$, the virtual time $\tau \to \infty$.

\begin{highlightbox}{Image Generation Guided by Text Prompts}{text-to-image-basic}
  Transforming this problem into a fundamental six-tuple $(\mathcal{S}, \mathcal{A}, \mathcal{P}, \pi, \mathcal{S}_{\mathrm{d}}, d)$ under the stabilization learning framework.
  \begin{itemize}[leftmargin=*, itemsep=2pt]
  \item \textbf{State Space ($\mathcal{S}$) and State ($s$)} \\
  State $s$ is the latent space representation obtained by encoding the image $I$ via a VAE encoder, containing the semantic and structural information of $I$. The state space $\mathcal{S}$ is the set of all possible latent space representations, typically a continuous space with dimensions much smaller than the original image space:
  \[
  \begin{aligned}
   s = \text{VAE}_\text{encode}(I) \in & \mathcal{S} = \mathbb{R}^{H \times W \times C}\\
   \text{(where } H, W, \text{ and } C &\text{ are the height, width, and number of channels respectively)}.
  \end{aligned}
  \]

  \item \textbf{Action Space ($\mathcal{A}$)}\\
  The action space shares the same dimensions as the state space, representing the denoising operation performed in the latent space. Each action $a_t$ corresponds to the noise $\hat{\epsilon}_\theta(s_t, t, c)$ predicted based on the conditional information $c$ at the current state $s_t$. Thus, the action space $\mathcal{A}$ is the set of all possible denoising operations:
  \[
  a_t = \hat{\epsilon}_\theta(s_t, t, c) \in \mathcal{A} = \mathbb{R}^{H \times W \times C}.
  \]

  \item \textbf{Plant ($\mathcal{P}$)}\\
  The state $s$ of the image in the latent space undergoes discrete evolution driven by the denoising operations performed by the scheduler, based on the noise $\hat{\epsilon}_\theta(s_t, t, c)$ predicted by the U-Net. The specific form is given by Equation \eqref{eq:denoise}.  

  \item \textbf{Desired State Set ($\mathcal{S}_{\mathrm{d}}$)}\\
  The desired state set $\mathcal{S}_{\mathrm{d}}$ refers to the set of latent space states of the target images that the prompt aims to generate, encompassing the latent representations of all real images $\mathcal{I}_\text{real}$ meeting the instruction requirements. Generally, an image $I$ is considered to meet the requirements if its semantic matching degree (CLIP score) with the text prompt $c$ exceeds a certain threshold $\eta$:
  \[
  \begin{aligned}
  \mathcal{I}_\text{real} &= \left\{I \,\middle|\, \frac{f^\top_\text{img}(I) f_\text{text}(c)}{\|f_\text{img}(I)\|\|f_\text{text}(c)\|} \geq \eta \right\},\\
  \mathcal{S}_{\mathrm{d}} &= \{ s_{\mathrm{d}} \in \mathcal{S} \mid s_{\mathrm{d}} = \text{VAE}_\text{encode}(I), I \in \mathcal{I}_\text{real} \},
  \end{aligned}
  \]
  where $f_\text{img}(\cdot)$ and $f_\text{text}(\cdot)$ are the feature vectors extracted by the CLIP image encoder and text encoder, respectively.
    
  \item \textbf{Metric Function ($d$)}\\
  The core of the metric function is to measure the proximity between the current generated image's latent state $s$ and the desired state set $\mathcal{S}_{\mathrm{d}}$. The distance can be compared directly in the latent space:
  \[
  d(s, \mathcal{S}_{\mathrm{d}}) = \min_{s_{\mathrm{d}} \in \mathcal{S}_{\mathrm{d}}} \|s - s_{\mathrm{d}}\|.
  \]
  In practice, it is often measured by the image generation quality, i.e., evaluating the semantic matching degree between the generated image and the set of valid images $\mathcal{I}_\text{real}$ using cosine similarity extracted by the CLIP image encoder:
  \[ 
  \begin{aligned}
  I = \text{VAE}_\text{decode}(s), I_{\mathrm{d}} = \text{VAE}_\text{decode}(s_{\mathrm{d}})\ \in \mathcal{I}_\text{real}, s_{\mathrm{d}} \in \mathcal{S}_{\mathrm{d}} \\
  d(s, \mathcal{S}_{\mathrm{d}}) = 1- \max_{s_{\mathrm{d}} \in \mathcal{S}_{\mathrm{d}}} \frac{f^\top_\text{img}(I) f_\text{img}(I_{\mathrm{d}})}{\|f_\text{img}(I)\|\|f_\text{img}(I_{\mathrm{d}})\|} \in [0,2].
  \end{aligned}
  \]
  A metric closer to 0 indicates a higher semantic match between the generated image and the text prompt, approaching the desired state set $\mathcal{S}_{\mathrm{d}}$.

  \item \textbf{Policy ($\pi$)}\\
  The objective of policy $\pi$ is to select appropriate denoising operations given the current latent state $s_t$. It aims to train the model to accurately predict the noise $\hat{\epsilon}_\theta(s_t,t,c)$, ensuring that as the latent space state iterates, the semantic matching degree improves, i.e., as $\tau \to \infty$ (corresponding to the state timestep $t \to 0$ in the diffusion and denoising processes), $  d(s_t, \mathcal{S}_{\mathrm{d}}) \to 0 $.
  \end{itemize}
\end{highlightbox}

This problem can also be viewed as a tracking problem of the image generation model with respect to text semantics, translating it into a tracking problem under the stabilization learning framework.

\begin{highlightbox}{Image Generation Guided by Text Prompts}{text-to-image-tracking}
  Transforming this problem into a tracking problem seven-tuple $ (\mathcal{S}, \mathcal{A}, \mathcal{P}, \pi, h, \mathcal{Y}_{\mathrm{d}}, d_\mathcal{Y})$ under the stabilization learning framework.
  \begin{itemize}[leftmargin=*, itemsep=2pt]
  \item \textbf{State Space ($\mathcal{S}$) and State ($s$)} \\
  State $s$ is the latent space representation obtained via the VAE encoder, defined as:
  \[
  \begin{aligned}
  & s = \text{VAE}_\text{encode}(I) \in \mathcal{S} = \mathbb{R}^{H \times W \times C}.
  \end{aligned}
  \]

  \item \textbf{Action Space ($\mathcal{A}$)}\\
  Represents the denoising operations in the latent space:
  \[
  a_t = \hat{\epsilon}_\theta(s_t, t, c) \in \mathcal{A} = \mathbb{R}^{H \times W \times C}.
  \]

  \item \textbf{Plant ($\mathcal{P}$)}\\
  Driven by discrete evolution according to Equation \eqref{eq:denoise}. 

  \item \textbf{Output Function ($h$)}\\
  Establishes the mapping between the latent space state $s$ and the semantic feature vector of the prompt:
  \[
  y_t = h(s_t) = f_\text{img}(\text{VAE}_\text{decode}(s_t)),
  \]
  where $f_\text{img}(\cdot)$ is the feature vector extracted by the CLIP image encoder.

  \item \textbf{Tracking Target ($ \mathcal{Y}_{\mathrm{d}}$)}\\
  The tracking target is the semantic feature vector $ y_{\mathrm{d}}$ of the text prompt:
  \[
  \mathcal{Y}_{\mathrm{d}} = \{ y_{\mathrm{d}} = f_\text{text}(c) \}.
  \]
    
  \item \textbf{Metric Function ($d_\mathcal{Y}$)}\\
  Measures the matching degree between the output image feature vector $y_t$ and the prompt semantic feature vector $ y_{\mathrm{d}}$ using cosine similarity:
  {
  \[
  d_\mathcal{Y}(y_t, \mathcal{Y}_{\mathrm{d}}) = 1 - \frac{y^\top_t y_{\mathrm{d}}}{\|y_t\|\|y_{\mathrm{d}}\|} \in [0, 2], \quad y_{\mathrm{d}} \in \mathcal{Y}_{\mathrm{d}}.
  \]}

  \item \textbf{Policy ($\pi$)}\\
  Aims to train the model to select optimal denoising operations to ensure that the generated images progressively track the expected image set designated by the prompt. As $\tau \to \infty$ (corresponding to the state timestep $t \to 0$ in the diffusion and denoising processes), $ d_\mathcal{Y}(y_t, \mathcal{Y}_{\mathrm{d}}) \to 0 $.
  \end{itemize}
\end{highlightbox}

\subsection{Multi-Robot Swarm Navigation in Virtual Tubes}

\begin{figure}[H]
  \centering
  \includegraphics[width=0.8\textwidth]{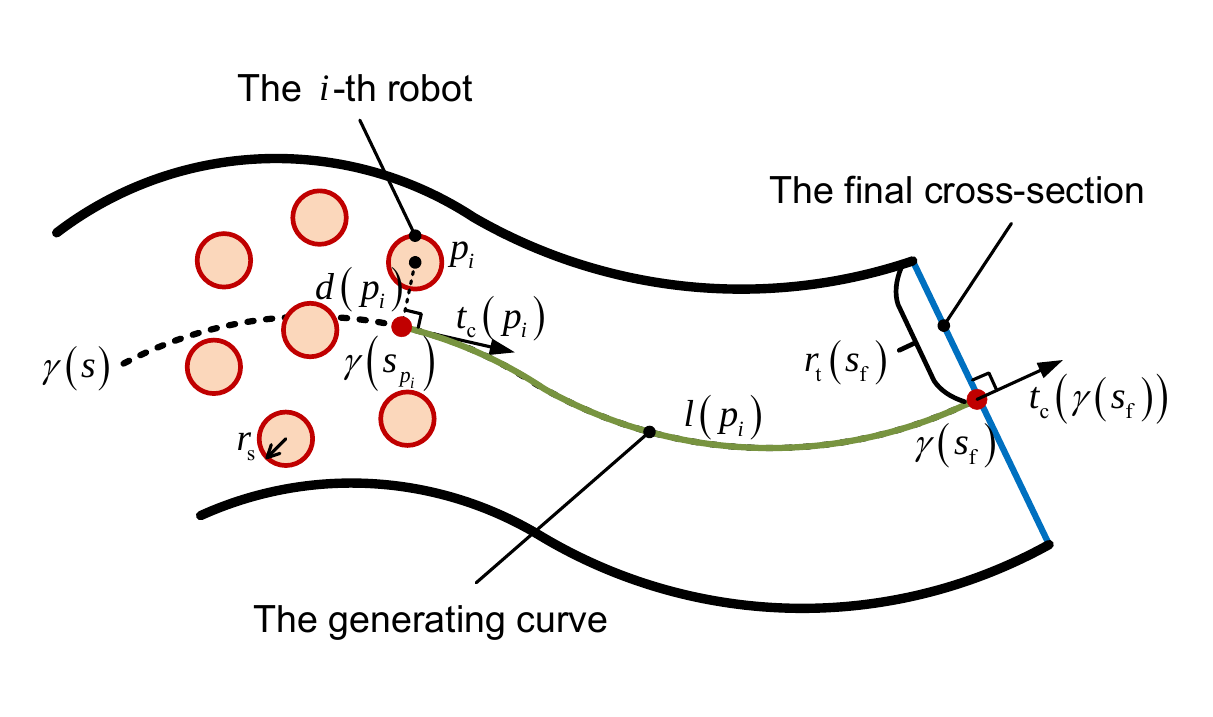}
  \caption{Schematic diagram of a virtual tube}
  \label{f1}
\end{figure}

\textbf{Problem Description}:
Consider the problem of guiding multiple robots through a virtual tube, i.e., directing a robot swarm to travel inside a virtual tube along the direction of the tube's generating curve, without colliding with other robots and without exceeding the boundaries of the virtual tube. The core of this problem lies in ensuring that all individuals in the robot swarm reach the terminal cross-section of the virtual tube.

For a swarm consisting of $N$ robots, the spatial coordinates of the robots are \(p_i \in \mathbb{R}^2, \,\, i = 1,\cdots,N\). Taking velocity as the control variable for the robot swarm, a first-order particle model is established:
\begin{equation*}
  \begin{aligned}
  & {\dot{p}}_{i} = \text{sat}\left( {{v}_{\text{c},i}}, {{v}_{\max }} \right) \\
  & \text{sat}\left( v, {{v}_\text{max }} \right) \triangleq 
  \begin{cases}
    v & \|v\| \le {{v}_\text{max }}  \\
    {{v}_\text{max }} \frac{v}{\|v\|} & \|v\| > {{v}_\text{max }}  
  \end{cases},
  \end{aligned}
\end{equation*}
where ${{v}_{\max }}$ is the maximum velocity the robot can achieve due to physical limitations. The goal is to complete the passage through the virtual tube by adjusting the velocity control variable ${{v}_{\text{c},i}}$.

A virtual tube is formed by sweeping a cross-section along a generating curve \cite{quan2022practical}. Here, a cross-section refers to a plane or line segment perpendicular to the tangent direction at each point on the generating curve. As shown in Figure \ref{f1}, in a 2D plane, the terminal cross-section of the virtual tube is a line segment perpendicular to the tangent direction at the end point of the generating curve. Let the generating curve be denoted as $\gamma(s)$, the end point of the generating curve as $\gamma(s_\text{f})$, the unit tangent vector at position ${{p}}$ as ${t}_{\text{c}}(p)$, and the radius of the virtual tube as ${{r}_{\text{t}}}(s)$. The terminal cross-section of the virtual tube can be expressed as:
\begin{equation*}
\left\{ x \,\middle|\, (x-\gamma(s_{\text{f}}))^\top {t}_{\text{c}}(\gamma(s_{\text{f}})) = 0, \quad \|x-\gamma(s_{\text{f}})\| \le r_{\text{t}}(s_{\text{f}}) \right\}.
\end{equation*}
It should be noted that $\gamma(s)$ can be any curve, but it must ensure that the generated virtual tube is regular, meaning no two cross-sections of the virtual tube intersect, preventing boundary folding \cite{mao2022making}.

The ultimate goal of stabilization learning is to make every individual in the swarm navigate through the virtual tube along the guided direction of the generating curve. Therefore, a line integral is used to define the distance between the robot and the terminal cross-section of the virtual tube.
Specifically, the distance between the $i$-th robot and the terminal cross-section is represented by a line integral along $\gamma(s)$:
\begin{equation*}
  l(p_i) = \int_{\gamma(s)} \text{d}l(p_i) = \int_{\gamma(s)} {t}^\top_{\text{c}}(p_i) \text{d}p_i,
\end{equation*}
where ${t}_{\text{c}}(p_i)$ is the unit tangent vector at the projected point $\gamma(s_{p_i})$ of the $i$-th robot on the generating curve. This definition ensures that robots on the same cross-section share the same distance to the terminal cross-section.
Since the generating curve is parameterized by $\gamma(s)$, we have:
\begin{equation*}
  t_{\text{c}}(p_i) = \frac{\text{d}\gamma(s_{p_i})}{\text{d}s}.
\end{equation*}
Note that the direction of the line integral is from $\gamma(s_{p_i})$ to $\gamma(s_\text{f})$. Thus, when the $i$-th robot has not reached the terminal cross-section, the integral $l(p_i)$ is always greater than 0; when it reaches the terminal cross-section, $l(p_i)$ equals 0.

As a parametric example, if the generating curve in a 2D Cartesian coordinate system is a sine curve in the domain $[0,2\pi]$, it can be parameterized as $\gamma(x) = [x \quad \sin x]^\top$. Then ${t}_{\text{c}}\left( {p}_i\right) = \frac{{[ \begin{matrix} 1 & \cos x  \\\end{matrix} ]}^\top}{\sqrt{1+(\cos x)^2}}$. If the projected point of the $i$-th robot has an x-coordinate of $\pi$, the line integral is $l(p_i) = \int_{\pi}^{2\pi} \sqrt{1+(\cos x)^2} \text{d}x$.

The desired state of the $i$-th robot is defined as its proportionally mapped point from its current cross-section to the terminal cross-section. As shown in Figure \ref{f1}, defining the distance from the $i$-th robot to its projection point $\gamma(s_{p_i})$ as $d(p_{i})$, its proportional position on the current cross-section is $\kappa = d(p_{i}) / r_{\text{t}}(s_{p_{i}})$. Thus, the expected state for the $i$-th robot is:
\begin{equation}
  \begin{aligned}
    & { p_{i}^\prime} (t) = \gamma(s_{\text{f}}) + \kappa \cdot r_{\text{t}}(s_{\text{f}}) R \cdot {t}_{\text{c}}(\gamma(s_{\text{f}})), \\
    & R = \begin{bmatrix}
      0 & -1  \\
      1 & 0  \\
    \end{bmatrix},
  \end{aligned}
  \label{pstar}
\end{equation}
where $R$ is the rotation matrix for a 90-degree counterclockwise rotation.

To ensure robots do not collide with each other or exceed the boundaries, barrier spaces are established. To avoid inter-robot collisions, the barrier space for the $i$-th robot is set as:
\begin{equation*}
  \mathcal{B}_{i} = \{ x \mid \|x - p_{i}\| <  r_\text{s} \},
\end{equation*}
where \(r_\text{s}\) is the safety radius. For the entire swarm, the collision-avoidance barrier space is:
\begin{equation*}
  \mathcal{B}_{\text{r}} = \mathcal{B}_{1} \cup \mathcal{B}_{2} \cup \cdots \cup \mathcal{B}_{N}.
\end{equation*}

To ensure robots stay within the virtual tube boundaries, the barrier space is set as:
\begin{equation*}
  \mathcal{B}_{\text{t}} = \left\{ x \,\middle|\, \min_{y \in \gamma(s)} \|x - y\| > r_\text{t}(s) - r_\text{s} \right\}.
\end{equation*}
Therefore, the total barrier space for the swarm is:
\begin{equation*}
  \mathcal{B} = \mathcal{B}_{\text{r}} \cup \mathcal{B}_{\text{t}}.
\end{equation*}

\begin{highlightbox}{Multi-Robot Swarm Navigation in Virtual Tubes}{multi-robots}
  Transforming this problem into a seven-tuple \(  (\mathcal{S},\mathcal{A}, \mathcal{P}, \pi, \mathcal{S}_{\mathrm{d}}, \mathcal{B}, d )\) under the stabilization learning framework.
  \begin{itemize}[leftmargin=*, itemsep=2pt]
    \item \textbf{State Space (\( \mathcal{S} \)) and State (\( s \))} \\
        Defined as the set of robot coordinates:
        \[
        \mathcal{S}=\left\{ s \,\middle|\, s = [{p_{1}^\top} \quad {p_{2}^\top} \quad \cdots \quad {p_{N}^\top}]^\top \in \mathbb{R}^{2N} \right\}.
        \]

    \item \textbf{Action Space (\( \mathcal{A} \))} \\
    Defined as the set of velocity control variables for the robot swarm:
    \[
    \mathcal{A}=\left\{ a \,\middle|\, a = [\text{sat}({v_{\text{c},1}}, {v_{\max}})^\top \quad \cdots \quad \text{sat}({v_{\text{c},N}}, {v_{\max}})^\top]^\top \in \mathbb{R}^{2N} \right\}.
        \]
        
    \item \textbf{Plant (\( \mathcal{P} \))} \\
    Driven by the kinematic equation where the next state is deterministically decided by the current state and control variables:
    \[
    \dot{s} = a, \quad s \in \mathcal{S}, \quad a \in \mathcal{A}.
        \]

    \item \textbf{Desired State Set (\( \mathcal{S}_{\mathrm{d}} \))} \\
    Based on Equation \eqref{pstar}, defined as the set of expected robot coordinates:
    \begin{equation*} 
      \mathcal{S}_{\mathrm{d}}(t) = \left\{ s_{\mathrm{d}}\, \middle|\, s_{\mathrm{d}} = [{p_1^\prime}^\top(t) \quad {p_2^\prime}^\top(t) \quad \cdots \quad {p_N^\prime}^\top(t)]^\top \in \mathbb{R}^{2N} \right\}.
        \end{equation*}
    
    \item \textbf{Barrier Space (\( \mathcal{B} \))} \\
    Defined as the union of the unsafe areas (inter-robot collisions) and the exterior of the tube boundaries, i.e., $\mathcal{B} = \mathcal{B}_{\text{r}} \cup \mathcal{B}_{\text{t}}$.    
        
    \item \textbf{Metric Function (\( d \))} \\
    A positive scalar function $d$ measuring the distance between the swarm's current state and the target set:
    \[
    d(s, \mathcal{S}_{\mathrm{d}}) = \sum\limits_{i=1}^{N}{l(p_i)}.
        \]
    When all robots reach the terminal cross-section, $d = 0$; otherwise, $d > 0$.
        
    \item \textbf{Policy (\( \pi \))} \\
    The objective of policy $\pi$ is to adjust the velocity control variable $a(t)$ based on the system state $s(t)$ and the desired state set $\mathcal{S}_{\mathrm{d}}(t)$. It aims to ensure that as $t \to \infty$, $ d(s(t), \mathcal{S}_{\mathrm{d}}(t)) \to 0$.    
  \end{itemize}
\end{highlightbox}

\subsection{End-to-End Motion Planning Problem Based on Flow Matching}

\begin{figure}[H]
		\centering
		\includegraphics[width=\columnwidth]{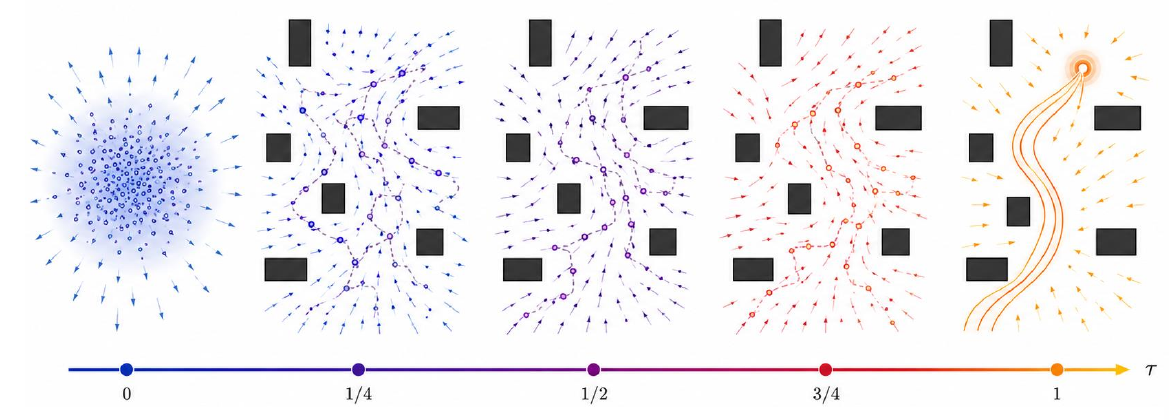}
		\caption{Schematic diagram of the end-to-end motion planning problem based on flow matching. During the trajectory generation process based on flow matching, the virtual integration times $\tau$ denoted as $1/4, 1/2, 3/4$, etc., represent the evolutionary progress of the system state flowing continuously from the initial prior distribution to the target trajectory manifold. As $\tau$ advances, under the strong conditional guidance of environmental perception, the time-varying velocity vector field constructed by the network deterministically drives the trajectory to undergo spatial deformation and reconstruction through ordinary differential equations; ultimately, when $\tau \to 1$, the system state smoothly and completely converges to a safe, collision-free expert demonstration trajectory shape.}
		\label{flow}
	\end{figure}
	
	\textbf{Problem Description:} With the rapid evolution of autonomous system technologies, it has become the norm for systems to perform tasks in complex, unstructured environments filled with unknown dynamic disturbances. Traditional motion planning methods usually adopt a serial modular structure of ``perception-mapping-planning-control". Although this paradigm possesses good interpretability, it is highly prone to cascading error accumulation between modules, and manually designed heuristic rules are often inadequate for the long-tail scenarios of high-dimensional environmental perception, leading to a severe deficiency in system robustness under extreme conditions.
	
	To break through the aforementioned bottlenecks, end-to-end generative motion planning has emerged. This paradigm advocates that the system directly maps high-dimensional environmental observations $O \in \mathbb{R}^{C \times H \times W}$ (where $C, H, W$ respectively represent the number of channels, height, and width of the observation features, such as multi-view camera array images or 3D LiDAR dense point clouds) end-to-end to continuous, collision-free trajectory sequences $y$ over a future period. In this context, Flow Matching (FM) \cite{Lipman2022FlowMF}, as an emerging generative framework based on continuous normalizing flows, provides advanced mathematical tools for end-to-end trajectory generation in complex environments, owing to its advantages in supporting optimal transport path construction, excellent numerical stability in solving ordinary differential equations, and the absence of simulated annealing assumptions.
	
	In the mathematical formulation of motion planning based on flow matching, the core task is to learn a continuous time-varying velocity vector field $v_\theta(s, \tau, O)$ conditioned on high-dimensional environmental observations $O$ (where $\theta$ represents the learnable parameters of the neural network), where $s$ is the state representation of the trajectory in the latent space or high-dimensional workspace, and $\tau$ is the virtual integration time during the flow matching model's generation process. Its state evolution, as shown in Figure \ref{flow}, strictly follows the ordinary differential equation below:
	\begin{equation}
		\frac{\mathrm{d} s}{\mathrm{d} \tau} = v_\theta(s, \tau, O), \quad \tau \in [0, 1].
	\end{equation}
	
	In this task, the evolution of the trajectory state is essentially a dynamic process of ``iterative optimization" or ``denoising stabilization." The system must start from a highly disordered random initial state (such as white noise from a standard Gaussian distribution), and under the strong guidance of the environmental prior $O$, continuously correct the geometric shape and spatiotemporal distribution of the trajectory through the vector field, until it asymptotically converges to the expert demonstration trajectory set $ \mathcal{Y}_{\mathrm{d}}$ that is safe, smooth, and compliant with physical laws. During this process, obstacle avoidance, as the most core rigid physical constraint in end-to-end motion planning, requires the network $v_\theta$ to deeply understand the 3D environmental topology and spontaneously form strong ``repulsive" or ``deflective" evolutionary gradients in state regions close to dangerous obstacle boundaries, ensuring that the trajectory is strictly confined within safe regions throughout its entire evolutionary cycle from disorder to order.
	
	Since directly computing and fitting the marginal vector field of the entire dataset is computationally infeasible, flow matching introduces the Conditional Flow Matching (CFM) mechanism \cite{wildberger2023flow}. Assuming the generic state variable of flow matching is $x$, CFM constructs a simple conditional probability path $p_\tau(x|x_1)$ and a corresponding conditional vector field $u_\tau(x|x_1)$ for each independent data sample $x_1$.
	
	In practical applications, the Optimal Transport Conditional Flow Matching (OT-CFM) \cite{tong2023improving} is widely adopted due to its stable training and straight evolutionary trajectories. It constructs a path from the initial noise sample $x_0 \sim p_0$ ($p_0$ is usually a standard Gaussian distribution) to the target data sample $x_1 \sim p_1$ ($p_1$ is the true expert data distribution) via linear interpolation:
	\begin{equation}
		x_{\tau} = (1-\tau)x_0 + \tau x_1.
	\end{equation}
	Under this definition, the derivative of the optimal transport path (i.e., the target vector field) manifests as a constant velocity:
	\begin{equation}
		u_\tau(x|x_1) = x_1 - x_0.
	\end{equation}
	The training objective of the neural network $v_\theta(x_\tau,\tau)$ is to fit this conditional vector field \cite{liu2022rectified}, with its loss function defined as the distance between the predicted vector field and the target vector field:
	\begin{equation}
		\mathcal{L}_{CFM}(\theta) = \mathbb{E}_{\tau\sim\mathcal{U}(0,1), x_1\sim p_1(x_1), x_0\sim p_0(x_0)} \left[ || v_\theta(x_\tau,\tau) - (x_1 - x_0) ||^2 \right].
	\end{equation}
	Where $\mathbb{E}$ denotes the mathematical expectation, and $\mathcal{U}(0,1)$ denotes the uniform distribution over the interval $[0,1]$. By minimizing the above objective function, the neural network $v_\theta$ can implicitly learn the ability to approximate the true marginal vector field. Extending this theoretical framework to motion planning problems conditioned on environmental perception merely requires replacing the state $x$ with the trajectory representation and inputting the environmental observations into the network as an additional condition.
	
	Regarding the mapping of virtual generation time to an infinite evolutionary domain: Stabilization learning frameworks typically require the system to reach an asymptotic steady state as evolutionary time $t \to \infty$. For flow matching, the generation process achieves precise transformation to the target distribution at $\tau=1$, which in engineering practice has already attained ``finite-time stabilization." To maintain rigorous consistency with the infinite-time asymptotic stability of stabilization learning in the theoretical architecture, a time mapping function $\tau = \phi(t)$ can be introduced, for example, defining the infinite virtual time of system evolution as $t = -\ln(1-\tau)$ (i.e., $\tau = 1 - e^{-t}$). As the original flow matching generation step $\tau \to 1$, the mapped stabilization time $t \to \infty$.
	
	Under this differential remapping, the dynamic equation of the original flow matching system is transformed into:
	\begin{equation}
		\frac{\mathrm{d} s}{\mathrm{d} t} = \frac{\mathrm{d} s}{\mathrm{d} \tau} \frac{\mathrm{d} \tau}{\mathrm{d} t} = v_\theta(s, 1-e^{-t}, O) \cdot e^{-t}.
	\end{equation}
	This transformation mathematically and rigorously ensures the smooth evolutionary behavior of the state system within the infinite time domain $t \in [0, \infty)$. Since the system has already converged to the desired set at the finite mapped moment $\tau=1$, it inherently possesses a stronger guarantee of asymptotic stabilization at the corresponding time $t \to \infty$. Therefore, the constrained trajectory generation process of flow matching can be reformulated as a state-constrained stabilization problem under the stabilization learning framework.

\begin{highlightbox}{End-to-End Motion Planning Based on Flow Matching}{flow-matching}
	
	Translating the above problem into a constrained tracking problem octuple $( \mathcal{S}, \mathcal{A},\allowbreak \mathcal{P}, \pi, h, \mathcal{Y}_{\mathrm{d}}, \mathcal{B}, d_{\mathcal{Y}})$ under the stabilization learning framework, expressed as follows.
	
	\begin{itemize}[leftmargin=*]
		\item \textbf{State Space ($\mathcal{S}$) and State ($s$)} \\
		Defined as the entire continuous set of trajectory representation sequences during the flow matching generation process. State $s$ contains the intermediate shapes of the entire flight trajectory dynamically deforming as evolutionary time $t$ progresses:
		$$\mathcal{S} = \{s \mid s \in \mathbb{R}^{T \times D}\}. $$
		In the case where the state is directly represented as a physical space state, $T$ can represent the discrete time steps of the predicted physical trajectory, and $D$ can represent the dimensions of a single-step motion pose or 3D coordinates. If the planning model extracts low-dimensional latent variables through a variational autoencoder for flow matching generation, then $T$ and $D$ represent the dimensions of the latent space state.
		
		\item \textbf{Action Space ($\mathcal{A}$)} \\
		Defined as the set of equivalent velocity vector fields output by the neural network after time mapping. At each evolutionary moment, action $a$ determines the direction and instantaneous rate at which the system state migrates toward the expert target set within the potential field:
		$$\mathcal{A} = \left\{a \;\middle|\; a = v_\theta(s, 1-e^{-t}, O) \cdot e^{-t} \in \mathbb{R}^{T \times D}\right\}.$$
		
		\item \textbf{Plant ($\mathcal{P}$)} \\
		The evolutionary model of the controlled plant is exactly the flow matching ODE dynamical system after time remapping, which is jointly driven by the current state, environmental observations, and policy control variables to continuously evolve the trajectory shape toward a steady state:
		$$\dot{s} = a, \quad s \in \mathcal{S}, a \in \mathcal{A}.$$
		
		\item \textbf{Output Function ($h$)} \\
		Usually implemented by a neural network Decoder, it establishes the mapping relationship from the latent space state during the generation process to the actual execution trajectory in physical space:
		$$y = h(s) = \text{Decoder}(s).$$
		The input to this function is the state $s \in \mathbb{R}^{T \times D}$, and the output is the trajectory sequence in the physical workspace, $y \in \mathbb{R}^{L \times N}$, where $L$ represents the number of discrete path steps, and $N$ represents the single-step output dimension (e.g., $N=3$ represents a path composed of a series of 3D spatial coordinates).
		
		\item \textbf{Tracking Target ($\mathcal{Y}_{\mathrm{d}}$)} \\
		Defined as the ideal trajectory manifold composed of expert demonstration data that is absolutely safe under the current environmental observation $O$:
		$$ \mathcal{Y}_{\mathrm{d}} = \{y_{\mathrm{d}} \in \mathbb{R}^{L \times N} \mid y_{\mathrm{d}} \text{ is the expert trajectory corresponding to environment } O\}. $$
		
		\item \textbf{Barrier Space ($\mathcal{B}$)} \\
		Defined as the set of all dangerous trajectories that cause physical collisions or exceed the safe flight zone. Considering that robots in the physical world have inherent geometric dimensions, define $P(y^{(k)}) \subset \mathbb{R}^3$ as the 3D physical space occupied by the fuselage when following the trajectory discrete step $y^{(k)}$, and let the physical obstacle region corresponding to the environmental observation $O$ be $\mathcal{W}_{\text{obs}} \subset \mathbb{R}^3$. As long as the volumetric footprint of the fuselage at any moment in the generated trajectory sequence overlaps with the physical obstacles (i.e., the intersection is not empty), the entire evolutionary trajectory is considered invalid and classified into the { barrier} space:
		$$\mathcal{B} = \left\{ y \in \mathcal{Y} \;\middle|\; \exists k \in \{1, \dots,L\}, \, P(y^{(k)}) \cap \mathcal{W}_{\text{obs}} \neq \emptyset \right\}.$$
		
		\item \textbf{Metric Function ($d_{\mathcal{Y}}$)} \\
		Measures the overall spatial deviation between the currently generated physical trajectory sequence $y$ and the expert target manifold $ \mathcal{Y}_{\mathrm{d}}$. The infinite-dimensional supremum norm is used to characterize the maximum error cost across the entire planning period:
		$$ d_{\mathcal{Y}}(y, \mathcal{Y}_{\mathrm{d}}) = \sup_{k \in \{1, \dots, L\}} ||y^{(k)} - y^{(k)}_{\mathrm{d}}||,$$
		where $y^{(k)}_{\mathrm{d}}$ represents the state of the expert trajectory $y_{\mathrm{d}}$ at the $k$-th discrete step.
		
		\item \textbf{Policy ($\pi$)} \\
		The core task of the policy is to train and optimize the time-varying vector field network $v_\theta$. It aims to precisely construct a continuous flow path for the target probability distribution, enabling the vector field to utilize observation $O$ to provide repulsive gradients to repel the { barrier} space $\mathcal{B}$ during state evolution, and stably and asymptotically converge to the expert target set as system evolutionary time $t$ advances. That is, when $t \to \infty$ (corresponding to the original flow matching generation step $\tau \to 1$), it rigorously satisfies $ d_{\mathcal{Y}}(y(t), \mathcal{Y}_{\mathrm{d}}) \to 0$.
	\end{itemize}
  \end{highlightbox}

\section{Future Development Directions}

With the continuous escalation of industrial intelligence and the control demands of complex systems, traditional control theory and modern machine learning each face their own bottlenecks. The former relies on accurate system models and struggles to adapt to complex scenarios characterized by prominent nonlinearity, hybrid dynamics, and high uncertainty. The latter, despite possessing data-driven flexibility, lacks strict theoretical support for stability, making its robustness and generalization capabilities difficult to meet engineering practice requirements. Against this backdrop, stabilization learning has emerged as an innovative cross-disciplinary paradigm bridging control theory and machine learning. Its core essence lies in utilizing ``stability constraints" as a nexus to integrate the rigor of traditional control with the flexibility of machine learning, providing a unified theoretical framework and solution path for complex problems across multiple domains. This section systematically expounds on the future development directions of stabilization learning, focusing on two core dimensions: ``construction of a unified problem framework" and ``realization of efficient and robust learning," clarifying key research contents, technical routes, and academic value to provide a systematic reference for subsequent research in this field.

\subsection{Stabilization Learning Framework: A Stabilization-Based Integration of Control and Learning Problems}

The core innovation of stabilization learning lies in breaking the problem barriers between the control and learning domains. It uniformly transforms various scattered control tasks and learning objectives into stabilization learning problems. By leveraging core tools such as Lyapunov stability theory and Lyapunov function modeling, it achieves structured solutions across different scenarios and tasks. This unified framework primarily encompasses two major dimensions: the stabilization transformation of control domain problems and the stabilization modeling of learning domain problems.

\subsubsection{Stabilization Transformation of Control Domain Problems}

Core tasks and multi-type system control demands in the traditional control domain can all be efficiently solved via data-driven methods under the stabilization learning framework, while inheriting the theoretical rigor of traditional control.

\paragraph{(1) Transformation of Core Control Tasks}
Basic control tasks such as state stabilization, output stabilization, state tracking, and output tracking, as well as complex control problems like adaptive control and robust control, can be uniformly transformed into stabilization learning problems through appropriate mathematical tools. For example, the need to compensate for parameter uncertainty in adaptive control can be addressed by dynamically adjusting the control policy through the online learning mechanism of stabilization learning while ensuring system stability. For external disturbances and unmodeled dynamics faced by robust control, the anti-disturbance design within stability constraints can achieve a synergy between data-driven flexibility and robust guarantees.

\paragraph{(2) Adaptation and Extension to Multi-Type Systems}
For controlled plants with different characteristics—such as linear, nonlinear, and hybrid systems—stabilization learning can transfer mature theoretical frameworks from traditional control and upgrade them with data-driven methods. For nonlinear systems, non-quadratic construction methods of Lyapunov functions can be used alongside learning models (like neural networks) to approximate nonlinear terms, achieving stabilizing control. For hybrid systems (e.g., stance-flight mode transitions in human running, multi-modal flight switching in UAVs), the design philosophy of multiple/switched Lyapunov functions can be adopted. Through the mode recognition and switching mechanisms of stabilization learning, stable control and smooth transitions across various modes are achieved. Additionally, for special scenarios like distributed systems and time-delay systems, extended stability criteria (such as distributed Lyapunov functions and delay-dependent stability conditions) can be incorporated into the unified stabilization learning framework.

\paragraph{(3) Extension and Fusion of Cross-Scenario Tasks}
Atypical control problems, such as path planning and trajectory optimization, can be transformed into stability problems through Lyapunov function construction, achieving unified solutions alongside traditional control tasks. Taking UAV path planning as an example: by constructing a comprehensive Lyapunov function encompassing path length, obstacle avoidance constraints, and energy consumption costs, the path optimization problem is transformed into a stabilization learning problem of ``making the system state converge to the optimal energy point." This transformation ensures flight stability and safety while guaranteeing path optimality. Similarly, tasks like robot collaborative control and mobile robot navigation can be transformed into stabilization learning problems using a similar approach, enhancing task adaptability and solution efficiency through a unified solving paradigm.

\subsubsection{Stabilization Modeling of Learning Domain Problems}

The core performance objectives of machine learning (such as optimality, generalization, and continual learning capabilities) can be transformed into structured optimization problems through the stability constraints of stabilization learning. This transformation provides strict theoretical guarantees for machine learning while enhancing its engineering practicality.

\paragraph{(1) Stabilization Modeling of Optimality Objectives (Oriented towards Reinforcement Learning)}
The core of reinforcement learning \cite{sutton2018reinforcement,bertsekas2019reinforcement} is to solve for the optimal control policy, whose essence can be described by the Hamilton-Jacobi-Bellman (HJB) equation. As introduced in Section \ref{sec:stability_optimality}, a profound mathematical duality exists between the optimal value function of the HJB equation and the Lyapunov function in stabilization learning. When the Lyapunov function is selected as the optimal value function of the HJB equation, the Lyapunov stability conditions (positive definite function, negative definite derivative) naturally transform into constraints for optimal control. That is, the system simultaneously achieves cumulative cost minimization during the stabilization process. This connection not only provides stability guarantees for reinforcement learning (preventing system divergence during training) but also transforms the optimal policy solving of reinforcement learning into a structured problem of ``finding the optimal Lyapunov function satisfying stability constraints," thereby enhancing the theoretical rigor and convergence reliability of the algorithm.

\paragraph{(2) Stabilization Modeling of Generalization and Continual Learning Capabilities (Oriented towards Transfer Learning and Continual Learning)}
The core of transfer learning \cite{pan2010survey, zhang2015deep} is migrating learned knowledge to new scenarios, where insufficient generalization performance is a key bottleneck; continual learning faces the challenge of catastrophic forgetting. Stabilization learning provides solutions for both through stability constraints. For transfer learning, the Lyapunov function learned in the source domain serves as prior knowledge. In the target domain, stability constraint parameters are optimized through data fine-tuning to ensure stable convergence of system performance during transfer, avoiding sudden performance drops caused by scenario differences and thus improving generalization. For continual learning \cite{kirkpatrick2017overcoming, delange2021continual}, incremental Lyapunov functions are constructed to retain the stability constraints of original tasks while learning new ones. This approach ensures that learning new tasks does not disrupt the stable performance of existing tasks, achieving knowledge accumulation without forgetting.

\paragraph{(3) Stabilization Modeling of Interactive Objectives (Oriented towards Imitation Learning)}
The core of imitation learning (including learning from demonstration) \cite{abbeel2004apprenticeship, celemin2024survey} is learning optimal policies from demonstration data. Its essence can be transformed into a problem of ``stable tracking and replication of demonstration trajectories." Stabilization learning transforms imitation learning into a stabilization problem of ``making the system state converge to the demonstration trajectory" by constructing a Lyapunov function tailored to the demonstration trajectory. By minimizing the Lyapunov function value, the learning system is guided to progressively approximate the dynamic characteristics of the demonstration trajectory while guaranteeing tracking stability, avoiding oscillations or divergence. This transformation not only simplifies the policy optimization process in imitation learning but also enhances the robustness of the learned policy through stability constraints, enabling it to maintain good performance even under noisy demonstration data or minor environmental perturbations.

\subsection{Efficient and Robust Stabilization Learning}

Building upon the unified problem framework, the practical engineering application of stabilization learning must address key issues such as ``low data utilization, parameter redundancy, and insufficient robustness." Therefore, another core future development direction of stabilization learning revolves around two major goals: ``Efficient Learning" and ``Robust Learning." It explores key technologies like Lyapunov function modeling, full-information utilization, and cross-domain constraint fusion to achieve deep integration of theoretical rigor and engineering practicality.

\subsubsection{Efficient Learning: Improving Data and Parameter Efficiency}

The core goal of efficient learning is to guarantee the convergence speed and control performance of stabilization learning while reducing data dependence and parameter scale. The key technical paths are as follows:

\paragraph{(1) Integrated Learning Driven by Energy Functions}
Using the Lagrangian function (the difference between kinetic and potential energy) as the core energy representation, combined with the principle of least action, this approach achieves the integrated learning of ``system model, control policy, and feature representation." As a physical prior describing system dynamics, the Lagrangian function directly correlates with system states, inputs, and energy changes. By embedding it into the stabilization learning framework, there is no need to independently learn models, policies, and features; instead, all three are simultaneously derived through the optimization of the energy function. Feature representations emerge naturally from the energy function, the control policy is determined by the gradient descent direction of the energy function, and the system model is obtained by fitting the energy function to actual system data. This integrated design drastically reduces parameter redundancy associated with independent modules and deeply couples feature representations with control tasks, improving learning efficiency and model adaptability. For example, in UAV flight control, energy modeling based on the Lagrangian function can synchronously learn aerodynamic models, attitude control policies, and flight state features without training multiple separate models, significantly lowering parameter scales and training data requirements.

Facing learning-based control problems in robotic systems, ensuring control stability and efficient data utilization is a core challenge. While reinforcement learning, as a mainstream data-driven method, possesses high flexibility, it often suffers from massive data requirements and a lack of stability guarantees. As an innovative control paradigm combining control theory and data-driven approaches, L-learning proposes a control framework based on Lagrangian model learning for ubiquitous Lagrangian dynamical systems\cite{quan2026llearning}. By embedding Lagrangian mechanics into deep neural networks for model learning and combining it with Lyapunov stability to design controllers, it achieves stable, precise, and efficient state tracking control for general robotic systems.

\begin{figure}[H]
	\centering
	\includegraphics[width=0.75\textwidth]{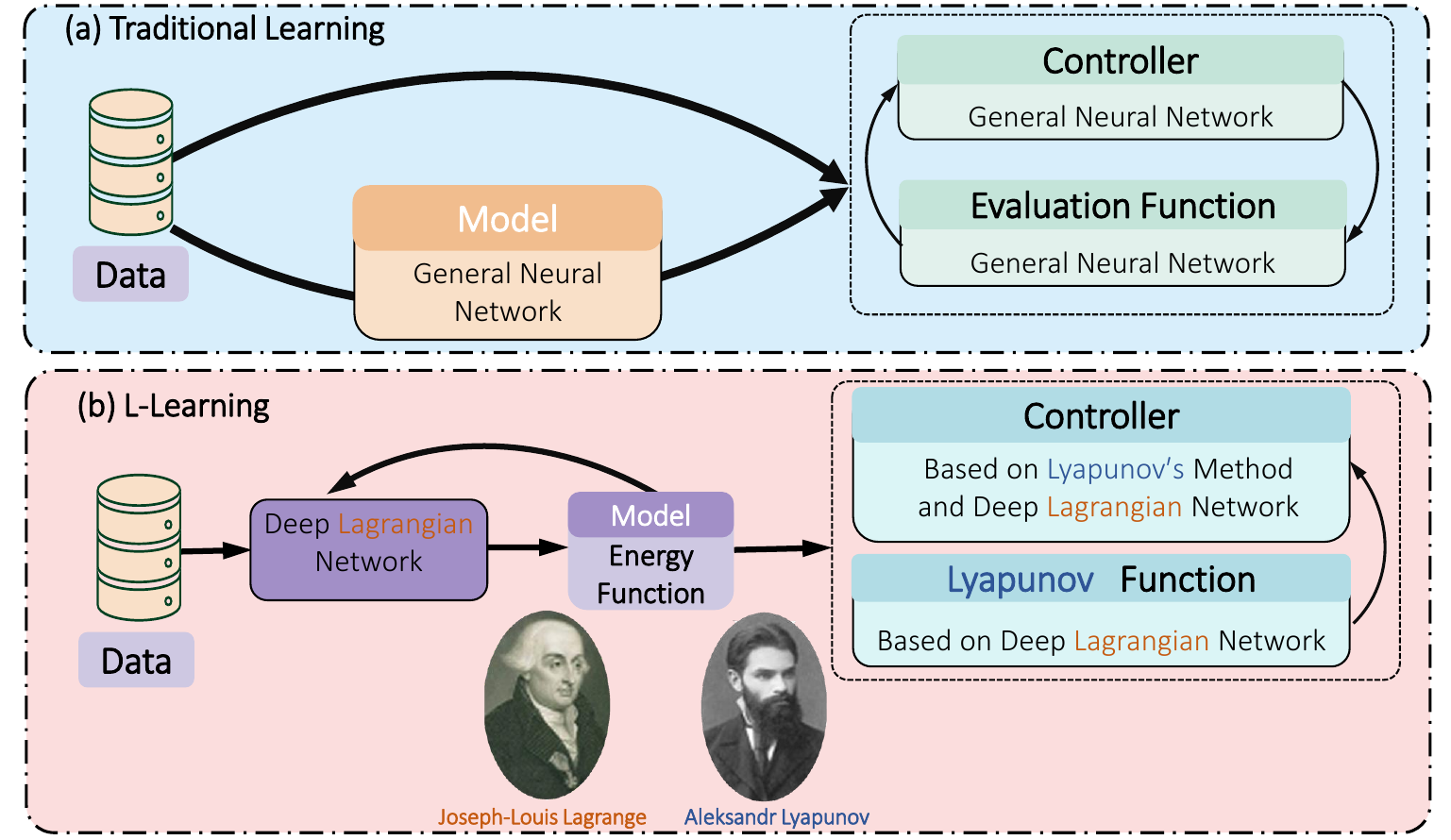}
	\caption{L-learning { vs.} traditional learning-based control\cite{quan2026llearning}}
	\label{fig:l_learning}
\end{figure}

As shown in Figure \ref{fig:l_learning}, unlike traditional learning methods that require separate learning of models, controllers, and value functions, the core idea of L-learning\cite{quan2026llearning} is to learn the system's Lagrangian, which is the core energy representation of the system. Through the learned Lagrangian and the automatic differentiation mechanism, the system dynamics matrix is no longer an independently identified parameter but second-order derivative information naturally induced by the Lagrangian via the Euler-Lagrange equation. This design means that { if the Lagrangian learning satisfies appropriate structural constraints, it can enhance physical consistency. L-learning demonstrates the feasibility of the stabilization learning paradigm as a path toward reconciling safety and efficiency.} Meanwhile, the controller of L-learning is no longer a ``black-box policy" but is constructed based on a Lyapunov function derived from the dynamics generated by the energy function, endowing the resulting controller with inherent stability guarantees. Therefore, L-learning does not optimize the model and policy separately but focuses entirely on optimizing the Lagrangian energy function. As the estimation accuracy of the Lagrangian improves, the system's dynamics model approximates true physical characteristics, and the controller's stability margin is correspondingly enhanced. In high-fidelity simulations for a 2-DOF manipulator and a quadrotor UAV system, L-learning was utilized for trajectory tracking control experiments and compared with algorithms like PID (Proportional-Integral-Derivative), SAC (Soft Actor-Critic), and TD3 (Twin Delayed Deep Deterministic Policy Gradient). Experimental results in \cite{quan2026llearning} demonstrate that L-learning exhibits superior tracking accuracy and learning efficiency. Compared to SAC and TD3, L-learning achieved controllers with better tracking performance in significantly shorter training times. L-learning indicates that the new stabilization learning paradigm, bridging control theory and machine learning, is an effective path to resolving the contradiction between safety and efficiency in robotic control.

\paragraph{(2) Deep Utilization of Full Information}
This path aims to break through the reliance of traditional data-driven methods on ``optimal samples" and fully utilize positive/negative sample data along with system structural prior information to improve data utilization.
\begin{itemize}[leftmargin=*, itemsep=2pt]
\item \textbf{Synergistic Learning of Positive and Negative Samples}: Introducing negative sample data (i.e., sub-optimal data leading to control failure or system instability) as a ``reverse supervision signal" for stability constraints. By penalizing the Lyapunov function values corresponding to negative samples, the system's ability to avoid unstable regions is reinforced. Simultaneously, optimal samples assist in refining the boundaries of the stability domain, enhancing the robustness and generalization of the learning policy.
\item \textbf{Resampling Guided by Model Priors}: Based on the structural priors of identifiable systems (such as mechanical systems described by Euler-Lagrange equations or nonlinear systems driven by visual perception), training data is resampled and weighted. The focus is placed on leveraging data from key dynamic regions of the system (e.g., nonlinear strongly coupled regions, mode switching regions), reducing the interference of redundant data on the learning process. For example, in hybrid system control, data at mode-switching moments can be intensively sampled based on the structural prior of mixed Lyapunov functions to improve learning accuracy during transitions.
\item \textbf{Multi-Source Information Fusion}: Integrating multi-modal sensor data (such as vision, force, and inertial measurements) and transforming them into a unified stability constraint via the Lyapunov function. This approach avoids the limitations of single-modal data and further enhances data utilization efficiency.
\end{itemize}

\subsubsection{Robust Learning: Strengthening Disturbance Rejection and Stability Guarantees}

The core goal of robust learning is to enhance the resistance of stabilization learning systems against uncertain factors such as external disturbances, internal parameter drift, and unmodeled dynamics, ensuring stable operation under complex conditions. The key technical path is to fuse time-domain and frequency-domain analysis methods from traditional control to build a dual robustness guarantee mechanism of ``time-domain convergence + frequency-domain disturbance rejection".

\paragraph{(1) Time-Domain Robust Constraints}
Centered on Lyapunov stability theory and combined with engineering requirements like input-output constraints and state boundary constraints, this path constructs highly robust stability criteria. For nonlinear systems, methods such as adaptive Lyapunov functions and reaching law design in sliding mode control are employed to ensure the system maintains asymptotic stability or input-output stability under parameter perturbations and external disturbances. For uncertain systems, robust adaptive laws are introduced to dynamically compensate for uncertainties through online learning, ensuring the negative definiteness of the Lyapunov function's derivative and achieving a synergy between stability and disturbance rejection.

\paragraph{(2) Frequency-Domain Robust Optimization}
Drawing on frequency-domain analysis methods from traditional control (e.g., Bode plots, Nyquist criterion, $\mathcal{H}_\infty$ control theory), frequency-domain characteristic design is applied to stabilization learning systems:
\begin{itemize}[leftmargin=*, itemsep=2pt]
\item \textbf{Bandwidth Optimization}: Optimizing the closed-loop bandwidth of the learning system based on the dynamic characteristics of the controlled plant and the frequency distribution of disturbances, balancing the system's response speed to useful signals with its attenuation capability against disturbance signals.
\item \textbf{High-Frequency Disturbance Rejection Design}: Suppressing the impact of high-frequency unmodeled dynamics (e.g., sensor noise, high-frequency vibration disturbances) on system stability by introducing Lyapunov function terms with low-pass filtering characteristics or regularization terms in the control policy.
\item \textbf{Robustness Index Constraints}: Transforming frequency-domain robustness indices, such as the $\mathcal{H}_\infty$ norm and peak gain, into optimization constraints for stabilization learning, ensuring the system meets disturbance rejection requirements across the frequency spectrum.
\end{itemize}
Through the deep fusion of time-domain and frequency-domain methods, stabilization learning systems can simultaneously guarantee the stability of state convergence and strong resistance to uncertainties, achieving robust performance on par with traditional control. For example, in UAV flight control, time-domain constraints guarantee the stable convergence of UAV attitude under gust disturbances, while frequency-domain constraints suppress the effects of high-frequency propeller vibrations and sensor noise, ensuring flight smoothness and safety.

\paragraph{(3) Fusion Innovation for Efficient and Robust Stabilization Learning}
The realization of efficient and robust stabilization learning is not a simple superposition of efficient learning and robust learning, but a deep integration at the levels of Lyapunov function modeling, constraint fusion, and optimization algorithm design to achieve collaborative optimization of ``efficiency-robustness-stability":
\begin{itemize}[leftmargin=*, itemsep=2pt]
\item At the \textbf{modeling} level, robustness constraints are embedded into the Lyapunov function design, endowing the Lyapunov function with the dual attributes of ``parameter efficiency optimization" and ``disturbance rejection capability reinforcement."
\item At the \textbf{optimization} level, optimization algorithms such as the Alternating Direction Method of Multipliers (ADMM) and Robust Stochastic Gradient Descent are employed to simultaneously solve the parameter optimization problem of efficient learning and the constraint satisfaction problem of robust learning, avoiding conflicts between the two.
\item At the \textbf{engineering implementation} level, considering hardware platform characteristics (e.g., computational resources, response speed requirements), lightweight designs are applied to efficient and robust learning algorithms, ensuring the unity of theoretical performance and engineering practicality.
\end{itemize}

\section{Conclusion and Thoughts on the First Principles of Artificial Intelligence}

This article has systematically constructed \textbf{Stabilization Learning}—a cross-disciplinary innovative paradigm bridging control theory and machine learning. It clarifies its core positioning of \textbf{``Stability First, Optimality Advanced"}, establishes a unified mathematical framework based on a six-tuple, and extends two types of expanded models: constrained learning with barrier spaces and tracking problems with targets. Furthermore, it accomplishes the formal translation of 11 typical problems, including multi-agent cooperative tracking, visual servo control, and game decision-making. It also elucidates its core boundaries and intrinsic mathematical connections with paradigms like certificate learning and reinforcement learning, providing a solution that combines theoretical rigor with engineering practicality for the intelligent control of complex systems characterized by high dimensionality, nonlinearity, and strong uncertainty. Based on the proposed theoretical system and multi-scenario verification presented in this article, we further distill the \textbf{First Principles of Artificial Intelligence Based on Stabilization Learning}, aiming to provide a foundational theoretical support for the development of Artificial General Intelligence (AGI).

The core hypothesis of these first principles is:

\begin{theorembox}{First Principles of Artificial Intelligence Based on Stabilization Learning}{first-principles}
Agents in open dynamic environments take system steady-state convergence as the core premise and \textbf{stability metric function minimization} as the unified criterion. Through real-time sensory feedback and adaptive policy regulation, they continuously reduce uncertainty or disturbance-induced deviations, in an entropy-like or energy-based sense, making the system state stably converge to the desired target set, while possessing the capability of continuous evolution for target expansion and environmental adaptation.
\end{theorembox}

This principle is proposed as a theoretical organizing hypothesis rather than an empirical summary of specific algorithms. It comprises three mutually supporting core pillars: 
First is the \textbf{Steady-State Priority Criterion}, which is the prerequisite for the existence and operation of agents. It breaks the logical shackles of the traditional optimality-first paradigm. 
Second is the \textbf{Closed-Loop Feedback Criterion of Stability Metric Function Minimization}, which is the unified pathway to realizing intelligent capabilities. Most intelligent decision-making and learning processes can be transformed into the monotonic decrease of positive definite stability metric functions, providing a generalized perspective for a broad set of AI tasks. 
Third is the \textbf{Adaptive Evolution Criterion}, which is the core mechanism for intelligent generalization and continuous evolution. Through the addition, merging, and adaptive weight allocation of stability metric functions, smooth expansion of capabilities is achieved, providing a theoretical pathway to alleviate the core pain points of traditional paradigms, such as catastrophic forgetting and weak generalization.

These first principles share a deep connection with and draw inspiration from the underlying operational laws of the physical world. Their theoretical foundation is rooted in two fundamental laws of nature: On one hand, the Principle of Least Action describes the energy-optimal path of conservative system evolution through the Euler-Lagrange equation, forming the primordial basis for the dynamic modeling of agents and the construction of stability metric functions. On the other hand, the law of entropy increase revealed by the Second Law of Thermodynamics clarifies that the core of stable operation in open systems is the active suppression of disorder. The process of minimizing the stability metric function achieved through closed-loop feedback control is precisely an active negative entropy process oriented towards control. The two achieve essential unity under the stability metric function framework. The stability metric function origin of the models and the entropy reduction goals of control jointly constitute the physical cornerstone of the first principles, endowing them with solid theoretical rationality.

Inspired by these first principles, core corollaries regarding various artificial intelligence capabilities can be formulated: { Most AI tasks, including perception and recognition, planning and decision-making, behavioral control, and generative tasks, may be reformulated from a stabilization convergence perspective.} Imitation learning can be interpreted as a stabilizing tracking problem of reference target sets; multi-task and continual learning can be viewed as the incremental combination and adaptive weight allocation of stability metric functions; perceptual robustness can be interpreted as the input-state stabilization problem of disturbance suppression; and the optimality pursued by traditional paradigms is merely a specific constrained optimization solution within the steady-state feasible region.

In summary, stabilization learning is not only an innovative paradigm for solving complex system control problems but also constructs a foundational theoretical system for artificial intelligence that conforms to physical laws, aligns with human cognition, and offers a unified perspective for diverse scenarios through the distillation of first principles. It breaks the disciplinary barriers between control and learning, { offering a path toward addressing key challenges in traditional AI paradigms, including module fragmentation, theoretical disunity, and lack of stability guarantees.} In the future, based on these first principles, we will further perfect unified transformation methods for cross-domain tasks, build efficient and robust integrated learning algorithms, and drive artificial intelligence towards an essential leap from ``optimal fitting in a single scenario" to ``stable adaptation across diverse open environments."

\section*{Acknowledgments}

The author would like to thank Yiqun Liang for proofreading the entire manuscript, arranging the layout, and providing relevant materials for the transformation of stabilization learning problems. Special thanks are extended to Chenyu Wang and Yixiang Guo for their supplements and careful review on the correlation between stabilization learning and reinforcement learning, as well as for contributing relevant materials to the transformation of stabilization learning problems; Chenyu Wang also provided materials for the problem of relative degree changes with state variations. Gratitude is also expressed to Ze Lu, Zhaolong Shen, Wenqi Song and Shuli Lv for providing supporting materials for the transformation of stabilization learning problems.

\bibliographystyle{ieeetr}
\bibliography{references}

\end{document}